\documentclass[a4paper,fleqn]{cas-dc}
\usepackage{float}
\usepackage[numbers]{natbib}
\usepackage{subfigure} 
\def\tsc#1{\csdef{#1}{\textsc{\lowercase{#1}}\xspace}}
\tsc{WGM}
\tsc{QE}
\tsc{EP}
\tsc{PMS}
\tsc{BEC}
\tsc{DE}

\begin{document}
\let\printorcid\relax
\let\WriteBookmarks\relax
\def\floatpagepagefraction{1}
\def\textpagefraction{.001}
\shorttitle{Leveraging social media news}
\shortauthors{Ao Chen et~al.}

\title [mode = title]{A Comparison for Anti-noise Robustness of Deep Learning Classification Methods on a Tiny Object Image Dataset: from Convolutional Neural Network to Visual Transformer and Performer}

\author[a]{Ao Chen}[type=editor,
						style=chinese,
                        auid=000,bioid=1,]
\author[a]{Chen Li}[style=chinese]
\cormark[1]
\ead{lichen201096@hotmail.com}
\author[a]{Haoyuan Chen}[style=chinese]

\author[a]{Hechen Yang}[style=chinese]

\author[a]{Peng Zhao}[style=chinese]

\author[a]{Weiming Hu}[style=chinese]

\author[a]{Wanli Liu}[style=chinese]

\author[a]{Shuojia Zou}[style=chinese]

\author[b]{Marcin Grzegorzek}[style=german]

\address[a]{Microscopic Image and Medical Image Analysis Group, MBIE College, Northeastern University, 110169, Shenyang, PR China}

\address[b]{Institute of Medical Informatics, University of Luebeck, Luebeck, Germany}

\cortext[cor1]{Corresponding author}

\begin{abstract}
Image classification has achieved unprecedented advance with the the rapid development of deep learning. However, the classification of tiny object images is still not well investigated. In this paper, we first briefly review the development of Convolutional Neural Network and Visual Transformer in deep learning, and introduce the sources and development of conventional noises and adversarial attacks. Then we use various models of Convolutional Neural Network and Visual Transformer to conduct a series of experiments on the image dataset of tiny objects (sperms and impurities), and compare various evaluation metrics in the experimental results to obtain a model with stable performance. Finally, we discuss the problems in the classification of tiny objects and make a prospect for the classification of tiny objects in the future.
\end{abstract}


\maketitle

\section{Introduction}
\subsection{Purposes of Tiny Object Analysis}
In computer vision and image processing, the classification of tiny objects is a fundamental task. There are mainly two definitions of tiny objects. One is an object with a relatively tiny physical size in real world. Another is mentioned in MS-COCO~\cite{2014Microsoft} metric evaluation, where objects occupying areas less than and equal to 32 × 32 pixels come under "tiny objects" category and this size threshold is generally accepted within the community for datasets related to common objects~\cite{tong2020recent}.\par 
Although many technologies are emerged to achieve the classification of medium and large objects in images, the classification of tiny objects is still extremely challenging. the challenges in
object classification include but not limited to the following aspects: indistinguishable features, low resolution, complicated backgrounds, limited context information~\cite{2021A}, tiny object rotation and scale change, precise object positioning, dense and occluded objects, accelerated detection, et al~\cite{zou2019object}.
\par However, the classification of tiny objects is very important. Research on tiny objects is often carried out in scientific research and medicine. Tiny object classification has been widely used in academia and real world applications, such as robot vision, autonomous driving, intelligent transportation, drone scene analysis, military reconnaissance, surveillance, and medical physiology research~\cite{tong2020recent}. such as the study of sperm. For human beings, one of six human couples is known to have fertility problems and more than 30 \% are clearly related to male infertility~\cite{barratt2007semen}.  Among many problems in the practical applications of tiny objects, the most serious one is the noise interference, so it is very necessary to find a better method to resist noise robustness.
\begin{figure*}
\centerline{\includegraphics[scale=.6]{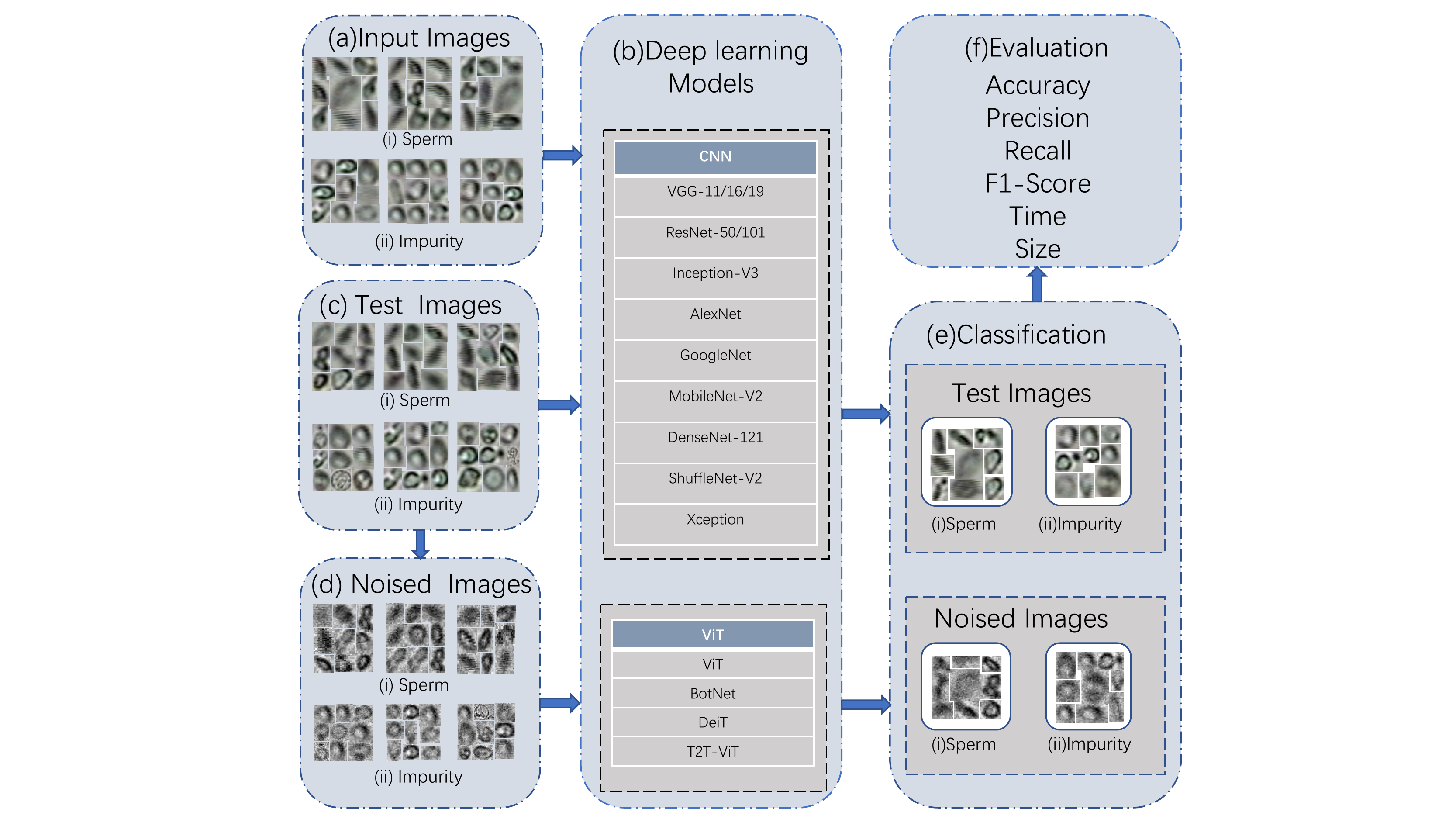}}
	\caption{\centering{A flowchart of comparison experiment on anti-noise robustness of tiny objects in deep learning}}
	\label{FIG:1}
\end{figure*}
\subsection{General Techniques of Image Classification}
In recent years, deep learning has a fast outstanding development trend in the field of computer vision, and has brought great benefits to the analysis of tiny objects. Among them, the most commonly used are \emph{Convolutional Neural Networks} (CNNs)~\cite{2018What}, such as AlexNet~\cite{krizhevsky2012imagenet}, VGG~\cite{simonyan2014very}, ResNet ~\cite{he2016deep}, GoogleNet~\cite{szegedy2015going}, DenseNet~\cite{huang2017densely}. In addition to CNNs, \emph{Visual Transformers} (VTs) are a kind of novel deep learning approaches for computer vision tasks, which is successfully inspired by the attention-based model in natural language processing~\cite{devlin2018bert}~\cite{vaswani2017attention}. More and more people are interested in the architecture of attention mechanism in the network~\cite{hu2018squeeze}. Recently, some researchers have proposed a hybrid architecture to transplant transformer components to solve visual tasks~\cite{carion2020end}~\cite{shen2020global}, including Vision Transformer (ViT)~\cite{dosovitskiy2020image}, BotNet~\cite{srinivas2021bottleneck}, Distilled Vision Transformer (DeiT)~\cite{touvron2020training} using a transformer, and Tokens-to-Token ViT (T2T-ViT)~\cite{yuan2021tokens} using a performer. The above methods have achieved good results in their respective fields, but the ability to analysis tiny objects is not yet known.
\subsection{Motivation of This Paper}
Robustness of prediction model based on deep learning refers to its ability to tolerate network input interference / noise. The analysis of image classification shows that the small change of input will lead to the sharp change of output~\cite{xie2019feature}~\cite{su2017One}, so the anti-noise robustness of deep learning model is very worthy of attention. In the microscopic image analysis, there are many noise interference problems in the classification of tiny objects. In order to find out the deep learning model with strong robustness to noise, a series of experiments are carried out in this paper. The flow chart of the experiment in this paper is shown in Fig.~\ref{FIG:1}. Here,  the sperm and impurity images are first acquired from  databases and used as training samples. Then, in the preprocessing step, the data set is allocated into a training set, a verification set and a test set according to a ratio of 6:2:2, and the test set is processed with noise as another additional noise test set. After that, the trained model is obtained through the training set and the verification set. Transfer learning is used to classify the original test set and the test set after noise processing. Finally, through the classification results, the accuracy, precision, recall, F1 score, model size, running time and other evaluation metrics are calculated.

\subsection{Structure of This Paper}
This paper is organized as follows: Section 2 illustrates the work about the difficulties in the application of tiny object analysis, the development of deep learning in computer vision, conventional noises for images and adversarial attacks for images. In Section 3, we conduct a series of comparative experiments, and make a simple analysis according to the evaluation index, and discuss the reasons why the classification results of tiny objects are excellent in the model. In Section 4, we summarize the paper and present prospective work.
\section{Related Works}
This section summarizes the deep learning methods used to analyze tiny objects in images and images after noise processing. In addition, this section is compiled from four aspects, mainly introduces the technical problems focused on the analysis of tiny objects, the conventional noises and adversarial attacks in images, and the development of deep learning in the field of computer vision.
\subsection{Technical Difficulties of Tiny Object Analysis}
In the real world, due to the limitation of camera resolution and other reasons, the research object can be very tiny. Therefore, there are a series of analytical technical problems~\cite{yang2019scrdet}. In remote sensing image classification~\cite{cheng2016survey}, tiny objects such as vehicles and ships are densely distributed, resulting in cluttered location of tiny objects and easy superposition~\cite{tang2017arbitrary}; air objects such as aircraft and satellites  can appear in various orientations~\cite{liu2017learning}. In tiny pedestrian classification~\cite{alahi2016social}, for example, in applications such as maritime rescue and driving assistance, the surrounding environment is complex, resulting in tiny people in the image display fuzzy, low resolution and easy to be overlapped by other objects~\cite{enzweiler2008monocular}. In microbial classification, due to the majority of microbial sampling from the outdoor, there are a lot of impurities, resulting in noise image problems~\cite{kosov2018environmental}. 
\subsection{Conventional Noises for Images}
Noise is introduced in image acquisition or transmission. Different factors may lead to noise in the image. The main sources of noise in the digital image are~\cite{verma2013comparative} :
\begin{enumerate}[(1)]
\itemsep=0pt
\item The imaging sensor is affected by the environment during the image acquisition process, and the sensor performance is affected;
\item Laboratory light and temperature are insufficient;
\item The image transmission channel is disturbed;
\item Scanner screen dust and other debris.
\end{enumerate}\par  
Depending on the type of noise interference, conventional noises for images can be classified as "salt and pepper noise", "gaussian noise", "uniform noise", "speckle noise", "poisson noise", "periodic noise", "rayleigh noise" and "exponential noise"~\cite{gonzalez2008Digital}.\par

Salt and pepper noise are also called impulse noise, spike noise, random noise or independent noise.  In salt and pepper noise, because of sparse light and dark interference, pixels in the image are very different in color or intensity, different from the surrounding pixels, resulting in the image contains black and white points, so called salt and pepper noise~\cite{gonzalez2008Digital}.  Salt and pepper noise in the images may be caused by the sharp interference of the signal, and only affect a small number of pixels in the digital image~\cite{mythili2011Efficient}.\par 

Gaussian noise is also commonly known as electronic noise because it is produced in amplifiers or detectors.  Looking for the source, Gaussian noise is caused by natural sources, such as thermal vibrations of atoms and discrete characteristics of radiation from warm objects~\cite{das2015comparative}. Usually, Gaussian noise in images interfere with the gray value in digital images~\cite{boyat2015review}.\par

Uniform noise is to quantify the pixels of the perceived image to multiple discrete levels, so that the noise distribution is approximately uniform~\cite{mythili2011Efficient}. The appearance of uniform noise is inherent in the amplitude quantization process, and is usually due to the existence of converting the simulated data into digital data~\cite{boyat2015review}.\par

Speckle noise can be multiplied by random value by pixel value, so it is also called multiplicative noise~\cite{gonzalez2008Digital}, and speckle noise arises from coherent processing of backscattered signals from multiple distribution points~\cite{verma2013comparative}. The speckle noise image is similar to Gaussian noise, and its probability density function follows gamma distribution~\cite{boyat2015review}.\par

Poisson noise has a root-mean-square value proportional to the square root of image intensity. Noises at different pixels are independent of each other and follow Poisson distribution~\cite{mythili2011Efficient}. Poisson noise may be caused by small number of photons perceived by sensors, which is insufficient to provide detectable statistical information~\cite{patidar2010image}. In practical applications, poisson noise and other sensor-based noise destroy images in different proportions~\cite{verma2013comparative}.\par

Periodic noise is generally generated by electronic interference, especially in the power signal during image acquisition. Periodic noise has special characteristics, such as spatial dependence and multiples of sine waves at specific frequencies~\cite{boyat2015review}. In practical applications, an image is affected by periodic noise, which causes many bars to appear on the image~\cite{mythili2011Efficient}.\par

Rayleigh noise is generated in radar ranging image~\cite{verma2013comparative}. Meanwhile, rayleigh density is beneficial to characterize noise in depth imaging~\cite{gonzalez2008Digital}. Moreover, Exponential noise refers to a type of noise whose probability density function obeys the exponential distribution~\cite{gonzalez2008Digital}.\par

\subsection{Adversarial Attacks for Images}
Although deep learning is excellent in many tasks in the field of computer vision, there is still a question whether the deep model can still produce satisfactory results for an abnormal input. The concept of adversarial samples is first proposed by Szegedy et al~\cite{szegedy2013intriguing}. They proved that despite the high accuracy, modern deep networks are very vulnerable to attacks from adversarial samples. These adversarial samples have only a slight disturbance, so that the human visual system cannot detect this disturbance (the image looks almost the same). Such attacks will lead to neural networks completely change its image classification. In addition, the same image perturbation can deceive many network classifiers.
In this paper, we apply four popular adversarial attack methods to evaluate the robustness of networks:\begin{enumerate}[(1)]
\itemsep=0pt
\item \emph{Fast gradient method} (FGM);
\item  \emph{Fast sign gradient method} (FSGM);
\item  \emph{Iterative fast sign gradient method} (I-FSGM);
\item DeepFool.
\end{enumerate}\par  
Deepfool~\cite{moosavi2016deepfool} is a classical adversarial attack method based on gradient iteration method, which generates the smallest disturbance and has high attack accuracy. It is the first time to define the sample robustness and model robustness, and it can accurately calculate the depth classifier disturbance to large-scale datasets, so as to reliably quantify the robustness of the classifier. 

FGM~\cite{miyato2016adversarial} and FGSM~\cite{goodfellow2014explaining} are adversarial attack methods based on gradient generation of adversarial samples. The idea of the two is that the direction of disturbance is along the direction of gradient increase, which means that the loss increases the most. The difference between them is that the gradient of FGM is normalized by $L_2$ norm, and the gradient of FGSM is normalized by max through sign function.\par 

I-FGSM~\cite{kurakin2016adversarial} finds the disturbance of each pixel in an iterative manner, rather than changing all pixels at once. I-FGSM is an extension of FGSM. It has been iterative for many small steps, and the pixel values of the results are clip after each step to ensure that the results are in the $\epsilon$ neighborhood of the original image. This iterative method is possible to find a adversarial sample when each pixel changes less than $\epsilon$, and if not, the worst effect is the same as the original FGSM. (Clip : In the iterative update process, as the number of iterations increases, part of the pixel values of the sample may overflow, such as beyond the range of 0 to 1, which needs to be replaced by 0 or 1 before generating a valid image. The process ensures that each pixel of the new sample is not distorted in an area of each pixel of the original sample.)\par

\subsection{Development of Deep Learning in Computer Vision}
During the study of \emph{Artificial Neural Networks} (ANNs), the concept of deep learning ienghs proposed~\cite{hinton2006reducing}. In the past few decades, ANNs have been widely used in various fields, but the network has the problems of long calculation time and easy to fall into local optimum and over-fitting~\cite{liu2017survey}. Until 2006, Hinton propose a new method that marked the formal formation of deep learning~\cite{hinton2006fast}, where the main idea of the new method is to train a simple two-layer unsupervised model, freeze all parameters, paste a new layer on the upper layer, and only train the parameters of the new layer~\cite{pouyanfar2018survey}. After decades of development, deep learning has become one of the most popular and effective tools. The rise of deep learning is mainly due to the following two reasons. First, the development technology of big data analysis shows that deep learning can partially solve the over-fitting problem in the previous network training data. Secondly, non-random initial values can be assigned to the network in the pre-training process. Because of the above reasons, it can obtain better local minimum and faster convergence speed after the training process~\cite{liu2017survey}.\par

With the rapid development of deep learning, many deep learning networks have emerged. In deep learning, CNNs are very popular and widely used~\cite{lecun1995convolutional}, VTs are a kind of novel deep learning approaches for computer vision tasks. In this paper, we mainly explore the classification effect of typical CNNs and VTs for tiny object and their robustness to anti-noise.\par

\subsubsection{Convolutional Neural Networks}
In typical CNNs, Alexnet is proposed by Alex Krizhevsky et al, and the whole network structure is composed of five convolution layers and three fully connected layers. The main new technology is to introduce ReLU as activation function, add Dropout layer to prevent overfitting, use overlapping maximum pooling to avoid the fuzzy effect of average pooling, and propose LPN layer to increase generalization ability~\cite{krizhevsky2012imagenet}.\par

VGGNet is a deep convolutional neural network developed by Oxford University Computer Vision and Google DeepMind researchers. The entire network uses the same size of 3 $\times$ 3 convolution kernel and 2 $\times$ 2 pooling layer, and forms the receptive fields of 5 $\times$ 5 and 7 $\times$ 7 by stacking two to three 3 $\times$ 3 convolution layers, which verifies that the performance can be improved by deepening the network structure~\cite{simonyan2014very}. This article uses VGG-11, VGG-16, VGG-19.\par

GoogleNet is a deep neural network model based on Inception module. The main innovation of this network is the use of inception module, which is convenient to add and modify, and the use of global average pooling layer instead of connection layer to reduce parameters~\cite{szegedy2015going}.\par 

Based on GoogleNet, Inception-V3 is proposed by Christian Szegedy et al. The main highlight of the model is to propose general network structure design guidelines, use volume integral solution to improve efficiency and reduce dimensionality through efficient feature maps~\cite{szegedy2016rethinking}. \par

ResNet is proposed by Kaiming He and other four Chinese of Microsoft Research Institute. This network model is first introduced into the residual network. Meanwhile, the residual network mainly has two residual modules, one is that two convolutional networks of 3 $\times$ 3 are connected in series as a residual module, and the other is that three convolutional networks of 1 $\times$ 1, 3 $\times$ 3 and 1 $\times$ 1 are connected in series as a residual module. 
The residual network not only improves the shortcomings of the past that the deeper the network, the more difficult it is to train, but also speeds up the convergence of the model. This article uses the widely used ResNet50 and ResNet101~\cite{he2016deep}.\par

Inspired by ResNet, DenseNet is proposed in 2016. The idea of the DenseNet network is to design a dense convolutional network where the input of each layer is the union of the outputs of all the previous layers, and the learned feature maps in this layer are also directly transmitted to all subsequent layers as input. Meanwhile, this dense method allows each layer to use all the features learned before, without repeated learning. In addition, the innovation of this network is to alleviate the problem of gradient disappearance, enhance feature propagation, promote feature reuse, and greatly reduce the number of parameters~\cite{huang2017densely}.\par

Although the above various CNNs improve the network performance, they also bring about the storage problem of the model and the speed problem of the model's prediction. Therefore, the lightweight model design is proposed. The idea is to reduce the network parameters without loss of accuracy. This article mainly introduces ShuffleNet v2~\cite{ma2018shufflenet}, Mobilenet v2~\cite{sandler2018mobilenetv2} and Xception~\cite{chollet2017xception}. \par
The main innovation of ShuffleNet v2 is to propose that \emph{Floating Point Operations Per Second} (FLOPs) is an indirect indicator of speed and propose the following four lightweight principles~\cite{ma2018shufflenet}: 
\begin{enumerate}[(1)]
\itemsep=0pt 
\item When the value of the number of input channels and the number of output channels is close to 1:1, the \emph{memory access cost} (MAC) time can be reduced; 
\item Too much group convolution increases MAC time; 
\item The split of the network reduces the parallelism; 
\item Element-wise operation cannot be ignored.
\end{enumerate}\par

The main innovation of MoblieNet V2 is to propose a new layer structure: the inverted residual with linear bottleneck. Moreover, the network application of this structure successfully solves the low dimensional data collapse caused by ReLU and makes feature reuse effectively slow down the degradation~\cite{sandler2018mobilenetv2}. \par
Xception is another improvement Google proposed for Inception-V3 after Inception. The main innovation is to use depth-wise separable convolution to replace the convolution operation in the original Inception-V3 to improve network efficiency~\cite{chollet2017xception}. Under strict definition, Xception is not a lightweight model, but it draws on depth-wise separable convolution, and depth-wise separable convolution is the key point of the above several lightweight models, Therefore, this article introduces Xception together.\par

\subsubsection{Visual Transformers}
In recent VTs, ViT is proposed by Dosovitskiy et al~\cite{dosovitskiy2020image}. The main working process of ViT is as follows. Firstly, the original image is divided into blocks and flattened into sequences. Then, the sequences are input into the encoder part of the transformer model. Finally, a full connection layer is connected to complete the ViT image classification task.~\cite{han2020survey}. Meanwhile, the main highlight of ViT is to show that it does not rely on convolutional neural networks and can also achieve good results in image classification~\cite{dosovitskiy2020image}. \par
BotNet is proposed by Srinivas et al, where the bottleneck in the fourth block in ResNet is replaced by a Multi-Head Self-Attention module~\cite{srinivas2021bottleneck}. \par
\emph{Distilled Vision Transformer} (DeiT) is proposed by Hugo Touvron et al. The innovation of DeiT is to introduce a distillation token, and then continuously interact with class token and patch token in self-attention layers~\cite{touvron2020training}. \par
\emph{Tokens-to-Token ViT} (T2T-ViT) improves the original ViT, and is proposed by Li Yuan et al. Through a method similar to the convolution windowing in CNNs, the adjacent tokens are locally aggregated, which help to model local features. At the same time, a deep and narrow structure is designed to reduce the calculation amount and improve the performance.~\cite{yuan2021tokens}.\par

\section{Materials and Methods}
\subsection{Dataset}
\subsubsection{Dataset organization}
\begin{figure}[h]
	\centering
		\includegraphics[scale=.4]{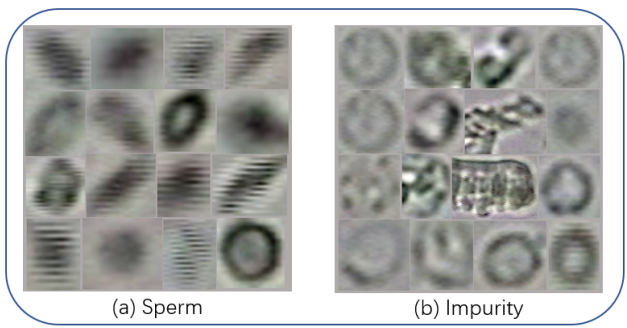}
	\caption{\centering{Some examples of sperm and impurity images}}
	\label{FIG:2}
\end{figure}

During the period from $2017$ to $2020$, data are collected from the sperm image database of the Dongfang Jinghua Hospital, Shenyang, P.R. China. This dataset is produced through the laboratory, which contains more than $278000$ annotated tiny objects. Moreover, the dataset includes three subsets of A, B and C. Subset-A and Subset-B are mainly used for tiny object detection and tracking, Subset-C is used for tiny object classification. In this paper, we use Subset-C as the database, which contains $125880$ independent sperm and impurity ($121401$ sperm images and $4479$ impurity images). Some examples of the dataset used in the experiment are shown in Fig.~\ref{FIG:2}.
\subsubsection{Data setting}
First, based on Subset-C, 4479 impurity images and 5058 sperm images are randomly selected, by a non-repeated and proportional algorithm, to balance our data setting. Then, the selected images are randomly grouped into training, validation, and test dataset according to a ratio of 6 : 2 : 2 (as shown in Table~\ref{tbl1}).
\begin{table}[width=.9\linewidth,cols=4,pos=h]
\caption{The arrangements of training, validation, and test dataset.}\label{tbl1}
\begin{tabular*}{\tblwidth}{@{} LLLL@{} }
\toprule
Dataset/Class & Sperm & Impurity & Total\\
\midrule
Train & 3035 & 2688 & 5723 \\
Validation & 1012 & 896 & 1908 \\
Test & 1011  & 895 & 1906 \\
Total & 5058 & 4479 & 9537 \\
\bottomrule
\end{tabular*}
\end{table}

\subsection{Conventional Noises Addition}
In order to research the anti-noise robustness of deep learning classifiers on tiny object image dataset, we add different conventional noises to the test dataset in Table~\ref{tbl2}, and compare the results through experiments. Meanwhile, the types and parameters of conventional noises are shown in Table~\ref{tbl2}.
\begin{table}[width=.9\linewidth,cols=4,pos=H]
\caption{Types and parameters of conventional noises introduced (/ represent default values, A and B represent sine amplitude and angle in periodic noise, and A and B represent mean value and variance in other noises.) }
\label{tbl2}
\begin{tabular*}{\tblwidth}{@{} LLLL@{} }
\toprule
Class & A & B \\
\midrule
\multirow{4}{*}{Gaussian} & 0 & 0.01  \\
 & 0 & 0.05\\
 & 0.2 & 0.01\\
 & 0.2 & 0.05\\
\multirow{2}{*}{Speckle} & 0 & 0.1  \\ 
 & 0  & 0.05 \\
Uniform & 0 & 1 \\
Rayleigh & 0 & 1 \\
Exponential & 1 & / \\
Poisson & / & / \\
\multirow{2}{*}{Salt \& Pepper} & 0.05 & / \\
& 0.1 & / \\
\multirow{2}{*}{Periodic} & 50 & 50 \\
 & 50 & 40 \\
\bottomrule
\end{tabular*}
\end{table}
Meanwhile,  we use adversarial attacks and take adversarial samples as test dataset. In the test, the $\epsilon$ is set as : $\epsilon$ = 0.001, 0.002, 0.004, 0.008, 0.016, 0.032, 0.064, 0.128, 0.256.
\subsection{Hyper Parameters for Deep learning Training}
In this section, we briefly show the parameters setting of model training in the form of a table. Table~\ref{tbl3} shows the parameters of the deep learning model training, such as learning rate, batch number, epoch number and optimizer. \par
\begin{table}[width=.9\linewidth,cols=4,pos=H]
\caption{The deep learning models train uniform parameters.}
\label{tbl3}
\begin{tabular*}{\tblwidth}{@{} LLLL@{} }
\toprule
Parameter & Value/Type & Parameter & Value/Type\\
\midrule
Batch &16  & Epochs &  100\\
Learning rate &0.0002 & Resize & (224,224) \\
Workers & 0 & Factor & 0.1  \\
Optimizer & Adam & Patience &10 \\
\bottomrule
\end{tabular*}
\end{table}

\subsection{Evaluation Metrics}
To select appropriate evaluation metrics is the key to evaluate the effect of deep learning model for tiny object classification. For classification standards, precision, recall, specificity, F1-score and accuracy are the most commonly used metrics~\cite{xie2015beyond}. Next, we briefly introduce the definition of the evaluation metrics in Table~\ref{tbl4}, Precision, which is for prediction results, means the probability of actually positive samples in all predicted positive samples. Recall is for the original sample, which means the probability of being predicted as a positive sample in the actual positive sample. F1-score is a balance point index that considers both the precision rate and the recall rate, so that the two reach the highest at the same time. The definition of accuracy is to predict the correct results as a percentage of the total sample. Specificity is defined as the proportion of the number of positive samples to all negative samples~\cite{powers2020evaluation}.
We introduce the mathematical expressions of precision, recall, F1-score, accuracy, sensitivity and specificity in the form of a table (shown in Table~\ref{tbl4}).
\begin{table}[width=0.9\linewidth,cols=4,pos=H]
\caption{Evaluation Metrics for Image Classification.}\label{tbl4}
\renewcommand\arraystretch{1.8}
\begin{tabular*}{\tblwidth}{@{} LL@{} }
\toprule
\makecell[c]{Assessments} & \makecell[c]{Formula} \\
\midrule
\makecell[c]{Precision ($P$)} & \makecell[c]{$\rm \frac{TP}{TP + FP}$} \\
\makecell[c]{Recall ($R$)} & \makecell[c]{$\rm \frac{TP}{TP + FN}$} \\
\makecell[c]{Specificity} & \makecell[c]{$\rm \frac{TN}{TN + FP}$} \\
\makecell[c]{F1-score} & \makecell[c]{$ 2 \times \frac{P \times R}{P + R}$} \\
\makecell[c]{Accuracy} & \makecell[c]{$\rm \frac{TP+TN}{TP + TN + FP + FN}$} \\
\bottomrule
\end{tabular*}
\end{table}
Meanwhile, we introduce the definition of TP, TN, FP and FN in Table~\ref{tbl4}. TP : true positive, indicating that the prediction is positive, the prediction is right, so the actual is also a positive example. TN : true negative, indicating that prediction is negative, prediction is right, so the actual is also negative. FP : false positive, indicating that the prediction is positive, wrong prediction, so the actual negative. FN : false negative, indicating that the prediction is negative, the prediction is wrong, so the actual is positive~\cite{sammut2011encyclopedia}.

\section{Comparison of Classification Experiment}
\subsection{Experimental Device}
Our classification comparison experiment is conducted on a local computer.   Based on the Win10 Professional operating system, the computer runs 16 GB memory, i7-8700 CPU and 8 GB NVIDIA GeForce GTX 1080 GPU.   Python 3.8.8, Pytorch 1.7.1 and Torchvision 8.0.2 are installed on the software configuration. In addition, our code runs in the integrated development environment Pycharm Community Edition.

\subsection{Experimental Results and Analysis}
\subsubsection{Classification Performance on Training and Validation Sets with Different Deep Learning Models}
\begin{table*}
\caption{Performance on Train and Validation dataset with Different Deep Learning Models. (In [\%].)}
\label{tbl5}
\begin{tabular}{@{}cllllllccc@{}}
\toprule
\multicolumn{1}{l}{Model}      & Class    & Precision & Recall & Specificity & Sensitivity & F1-Score & \multicolumn{1}{l}{Max Accuracy} & \multicolumn{1}{l}{Time(s)} & \multicolumn{1}{l}{Params Size(MB)} \\
\midrule
\multirow{2}{*}{AlexNet}       & Sperm    & 98.40     & 95.80  & 98.20       & 95.80       & 97.10    & \multirow{2}{*}{96.91}               & \multirow{2}{*}{1542}       & \multirow{2}{*}{55.64}              \\
                               & Impurity & 95.30     & 98.20  & 95.80       & 98.20       & 96.70    &                                      &                             &                                     \\
\multirow{2}{*}{VGG-11}        & Sperm    & 98.80     & 96.70  & 98.70       & 96.70       & 97.70    & \multirow{2}{*}{97.64}               & \multirow{2}{*}{5679}       & \multirow{2}{*}{247.2}              \\
                               & Impurity & 96.40     & 98.70  & 96.70       & 98.70       & 97.50    &                                      &                             &                                     \\
\multirow{2}{*}{VGG-16}        & Sperm    & 98.50     & 95.50  & 98.30       & 95.50       & 97.00    & \multirow{2}{*}{96.80}               & \multirow{2}{*}{9160}       & \multirow{2}{*}{268.16}             \\
                               & Impurity & 95.00     & 98.30  & 95.50       & 98.30       & 96.60    &                                      &                             &                                     \\
\multirow{2}{*}{VGG-19}        & Sperm    & 98.50     & 95.50  & 98.30       & 95.50       & 97.00    & \multirow{2}{*}{97.11}               & \multirow{2}{*}{11575}      & \multirow{2}{*}{532.45}             \\
                               & Impurity & 95.00     & 98.30  & 95.50       & 98.30       & 96.60    &                                      &                             &                                     \\
\multirow{2}{*}{ResNet-50}     & Sperm    & 99.30     & 96.20  & 99.20       & 96.20       & 97.70    & \multirow{2}{*}{97.64}               & \multirow{2}{*}{7163}       & \multirow{2}{*}{89.69}              \\
                               & Impurity & 95.90     & 99.20  & 96.20       & 99.20       & 97.50    &                                      &                             &                                     \\
\multirow{2}{*}{ResNet-101}    & Sperm    & 99.20     & 96.40  & 99.10       & 96.40       & 97.80    & \multirow{2}{*}{97.69}               & \multirow{2}{*}{10933}      & \multirow{2}{*}{162.14}             \\
                               & Impurity & 96.10     & 99.10  & 96.40       & 99.10       & 97.60    &                                      &                             &                                     \\
\multirow{2}{*}{GoogleNet}     & Sperm    & 98.80     & 93.90  & 98.70       & 93.90       & 96.30    & \multirow{2}{*}{96.12}               & \multirow{2}{*}{3749}       & \multirow{2}{*}{21.37}              \\
                               & Impurity & 93.40     & 98.70  & 93.90       & 98.70       & 96.00    &                                      &                             &                                     \\
\multirow{2}{*}{DenseNet-121}  & Sperm    & 99.30     & 97.00  & 99.20       & 97.00       & 98.10    & \multirow{2}{*}{98.06}               & \multirow{2}{*}{7800}       & \multirow{2}{*}{26.55}              \\
                               & Impurity & 96.70     & 99.20  & 97.00       & 99.20       & 97.90    &                                      &                             &                                     \\
\multirow{2}{*}{Inception-V3}  & Sperm    & 99.30     & 97.70  & 99.20       & 97.70       & 98.50    & \multirow{2}{*}{98.42}               & \multirow{2}{*}{6559}       & \multirow{2}{*}{83.12}              \\
                               & Impurity & 97.50     & 99.20  & 97.70       & 99.20       & 98.30    &                                      &                             &                                     \\
\multirow{2}{*}{MobileNet-V2}  & Sperm    & 99.20     & 95.90  & 99.10       & 95.90       & 97.50    & \multirow{2}{*}{97.43}               & \multirow{2}{*}{3830}       & \multirow{2}{*}{8.49}               \\
                               & Impurity & 95.60     & 99.10  & 95.90       & 99.10       & 97.30    &                                      &                             &                                     \\
\multirow{2}{*}{ShuffleNet-V2} & Sperm    & 99.40     & 97.00  & 99.30       & 97.00       & 98.20    & \multirow{2}{*}{98.11}               & \multirow{2}{*}{2667}       & \multirow{2}{*}{4.79}               \\
                               & Impurity & 96.70     & 99.30  & 97.00       & 99.30       & 98.00    &                                      &                             &                                     \\
\multirow{2}{*}{Xception}      & Sperm    & 99.20     & 97.30  & 99.10       & 97.30       & 98.20    & \multirow{2}{*}{98.16}               & \multirow{2}{*}{9609}       & \multirow{2}{*}{79.39}              \\
                               & Impurity & 97.00     & 99.10  & 97.30       & 99.10       & 98.00    &                                      &                             &                                     \\
\multirow{2}{*}{ViT}           & Sperm    & 93.00     & 94.20  & 92.00       & 94.20       & 93.60    & \multirow{2}{*}{93.13}               & \multirow{2}{*}{10143}      & \multirow{2}{*}{195.21}             \\
                               & Impurity & 93.30     & 92.00  & 94.20       & 92.00       & 92.60    &                                      &                             &                                     \\
\multirow{2}{*}{BotNet}        & Sperm    & 98.10     & 91.20  & 98.00       & 91.20       & 94.50    & \multirow{2}{*}{94.39}               & \multirow{2}{*}{7011}       & \multirow{2}{*}{71.72}              \\
                               & Impurity & 90.80     & 98.00  & 91.20       & 98.00       & 94.30    &                                      &                             &                                     \\
\multirow{2}{*}{DeiT-Base}     & Sperm    & 96.80     & 91.10  & 96.70       & 91.10       & 93.90    & \multirow{2}{*}{93.71}               & \multirow{2}{*}{17262}      & \multirow{2}{*}{329.65}             \\
                               & Impurity & 90.60     & 96.70  & 91.10       & 96.70       & 93.60    &                                      &                             &                                     \\
\multirow{2}{*}{DeiT-Tiny}     & Sperm    & 97.60     & 95.00  & 97.30       & 95.00       & 96.30    & \multirow{2}{*}{96.07}               & \multirow{2}{*}{3713}       & \multirow{2}{*}{21.67}              \\
                               & Impurity & 94.50     & 97.30  & 95.00       & 97.30       & 95.90    &                                      &                             &                                     \\
\multirow{2}{*}{T2T-ViT-t-19}  & Sperm    & 90.70     & 88.50  & 89.70       & 88.50       & 89.60    & \multirow{2}{*}{89.10}               & \multirow{2}{*}{16376}      & \multirow{2}{*}{147.39}             \\
                               & Impurity & 87.40     & 89.70  & 88.50       & 89.70       & 88.50    &                                      &                             &                                     \\
\multirow{2}{*}{T2T-ViT-t-24}  & Sperm    & 88.50     & 84.40  & 87.60       & 84.40       & 86.40    & \multirow{2}{*}{85.90}               & \multirow{2}{*}{21270}      & \multirow{2}{*}{242.19}             \\
                               & Impurity & 83.20     & 87.60  & 84.40       & 87.60       & 85.30    &                                      &                             &                                     \\
\multirow{2}{*}{T2T-ViT-7}     & Sperm    & 97.60     & 94.10  & 97.40       & 94.10       & 95.80    & \multirow{2}{*}{95.65}               & \multirow{2}{*}{4184}       & \multirow{2}{*}{15.25}              \\
                               & Impurity & 93.60     & 97.40  & 94.10       & 97.40       & 95.50    &                                      &                             &                                     \\
\multirow{2}{*}{T2T-ViT-10}    & Sperm    & 95.10     & 93.60  & 94.50       & 93.60       & 94.30    & \multirow{2}{*}{94.02}               & \multirow{2}{*}{4841}       & \multirow{2}{*}{21.28}              \\
                               & Impurity & 92.90     & 94.50  & 93.60       & 94.50       & 93.70    &                                      &                             &                                     \\
\multirow{2}{*}{T2T-ViT-12}    & Sperm    & 97.20     & 93.20  & 97.00       & 93.20       & 95.20    & \multirow{2}{*}{94.96}               & \multirow{2}{*}{5350}       & \multirow{2}{*}{25.29}              \\
                               & Impurity & 92.60     & 97.00  & 93.20       & 97.00       & 94.70    &                                      &                             &                                     \\
\multirow{2}{*}{T2T-ViT-14}    & Sperm    & 94.90     & 94.70  & 94.20       & 94.70       & 94.80    & \multirow{2}{*}{94.44}               & \multirow{2}{*}{8462}       & \multirow{2}{*}{80.42}              \\
                               & Impurity & 94.00     & 94.20  & 94.70       & 94.20       & 94.10    &                                      &                             &                                     \\
\multirow{2}{*}{T2T-ViT-19}    & Sperm    & 97.20     & 88.70  & 97.10       & 88.70       & 92.80    & \multirow{2}{*}{92.66}               & \multirow{2}{*}{13089}      & \multirow{2}{*}{147.39}             \\
                               & Impurity & 88.40     & 97.10  & 88.70       & 97.10       & 92.50    &                                      &                             &                                     \\
\multirow{2}{*}{T2T-ViT-24}    & Sperm    & 96.30     & 92.20  & 96.00       & 92.20       & 94.20    & \multirow{2}{*}{93.97}               & \multirow{2}{*}{17869}      & \multirow{2}{*}{242.19}             \\
                               & Impurity & 91.60     & 96.00  & 92.20       & 96.00       & 93.70    &                                      &                             &       \\                             
\bottomrule
\end{tabular}
\end{table*}
We select several training and loss curves for each CNN and VT model series in Fig~\ref{FIG:3}, and combine the evaluation metrics in Table~\ref{tbl5} to briefly understand the overall network performance used for classification training and verification of tiny objects and the performance of each epoch network for classification of tiny objects. \par
It can be seen from the Table~\ref{tbl5} that for the classification training and verification of tiny objects, the evaluation metrics of CNNs are generally higher than those of VTs. The curve of VGG-16 is the most stable, the verification effect of Inception-V3 is the highest, the running time of AlexNet is the least, and the size of ShuffleNet-V2 is the smallest. The accuracy of T2T-t-ViT-24 using transformer is the lowest, and the running time is the longest.\par  
When we focus on Fig~\ref{FIG:3}, we can find that under VGG, the accuracy of the training dataset increases rapidly from epoch 0 to 10 to 95 \%, and converges to 92 \% after 60 epochs. At the same time, the accuracy of the verification dataset reaches 95 \%. For the loss curve of VGG-16 training dataset and verification dataset, the development trends of the two are roughly similar. In the first ten epochs, the loss curve dropped rapidly to 0.2, and remained stable to 0.1 after 40 epochs. For the study of Inception accuracy and loss broken line diagram, it can be seen that the accuracy of the training dataset has been showing an upward trend. The accuracy of the verification dataset fluctuates greatly in the first 40 epochs, and the loss curves of the training dataset and the verification dataset are also roughly the same. The first 10 epochs decrease rapidly, and it decreases to 0.1 in the first 40 epochs, and then remains relatively stable. \par
However, for networks with poor performance such as T2T-t-ViT, BotNet and DenseNet-121, when we pay attention to the accuracy and loss curves of training and verification dataset of BotNet and DenseNet-121, we can find hat their curves fluctuate greatly, and the accuracy is basically stable at 98 \% and 90 \% respectively after the 80th epoch. For T2T-t-ViT, the accuracy curves of training dataset and verification dataset fluctuate greatly from 0 to 30th epoch, and it is not stable to 85 \% until the 50th epoch. At the same time, the loss curves of training dataset and verification dataset also fluctuate greatly from 0 to 30th epoch, and it is not stable to 0.35 after the 50th epoch. \par
In addition, we can see from Table~\ref{tbl5} that T2T-ViT-t-24 has the longest running time, with a total time of 21270 s, and AlexNet has the shortest running time, with a total time of 1542 s. For model size, ShuffleNet-V2 and MobileNet have the smallest sizes , Respectively 4.79 MB and 8.49 MB, VGG-19 has the largest size 532.45 MB.
\begin{figure*}
\centering
\subfigure[AlexNet]{
\begin{minipage}[t]{0.22\linewidth}
\includegraphics[width=4.5cm]{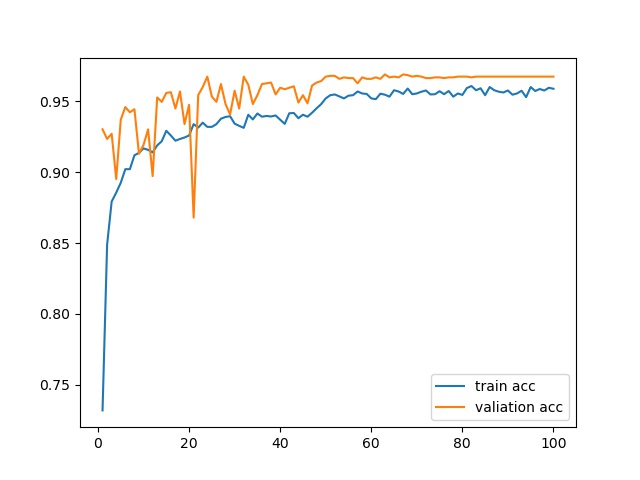}
\includegraphics[width=4.5cm]{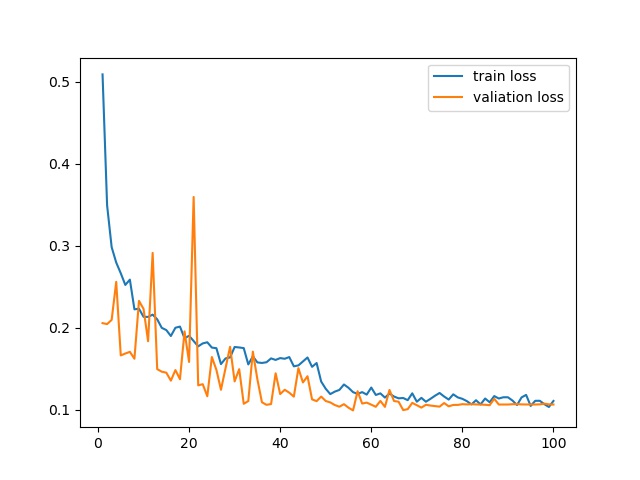}
\end{minipage}
}
\subfigure[VGG-16]{
\begin{minipage}[t]{0.22\linewidth}
\includegraphics[width=4.5cm]{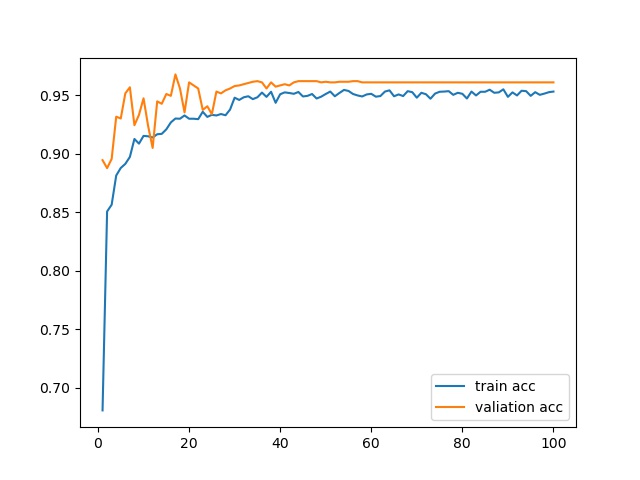}
\includegraphics[width=4.5cm]{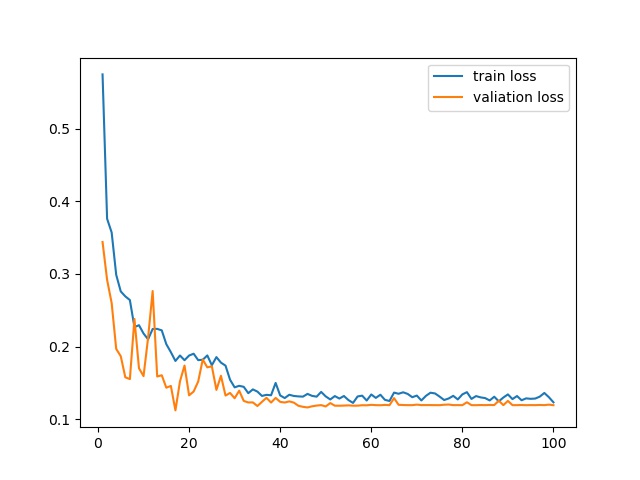}
\end{minipage}
}
\subfigure[VGG-19]{
\begin{minipage}[t]{0.22\linewidth}
\includegraphics[width=4.5cm]{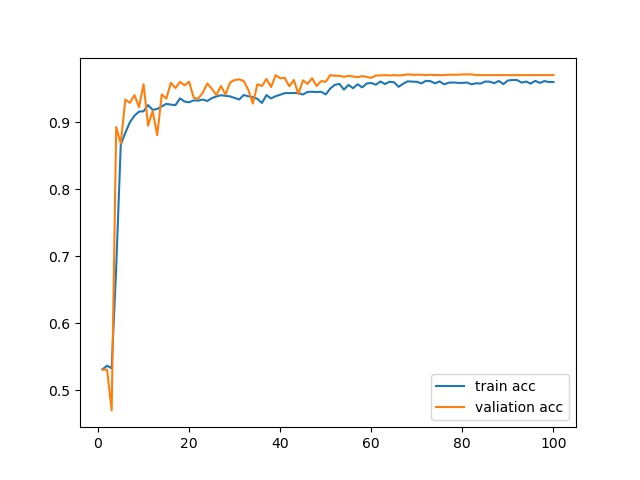}
\includegraphics[width=4.5cm]{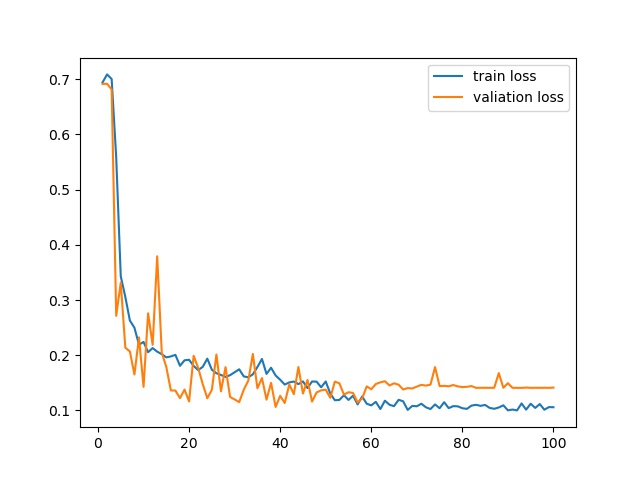}
\end{minipage}
}
\subfigure[ResNet-50]{
\begin{minipage}[t]{0.22\linewidth}
\includegraphics[width=4.5cm]{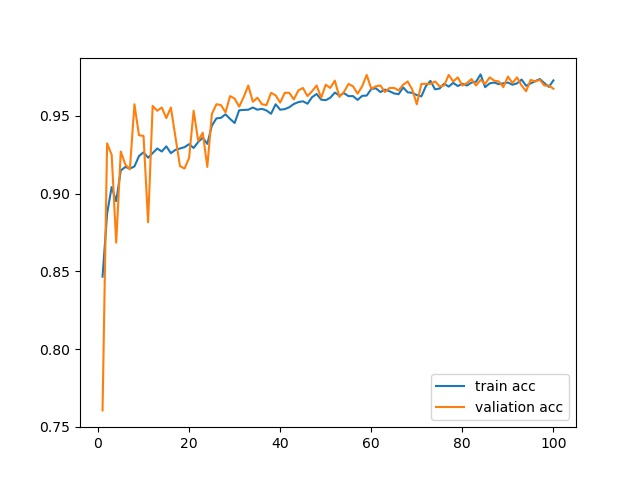}
\includegraphics[width=4.5cm]{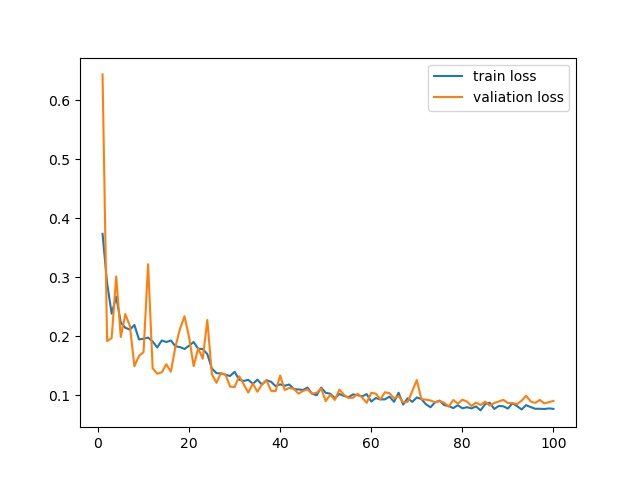}
\end{minipage}
}
\quad
\subfigure[GoogleNet]{
\begin{minipage}[t]{0.22\linewidth}
\includegraphics[width=4.5cm]{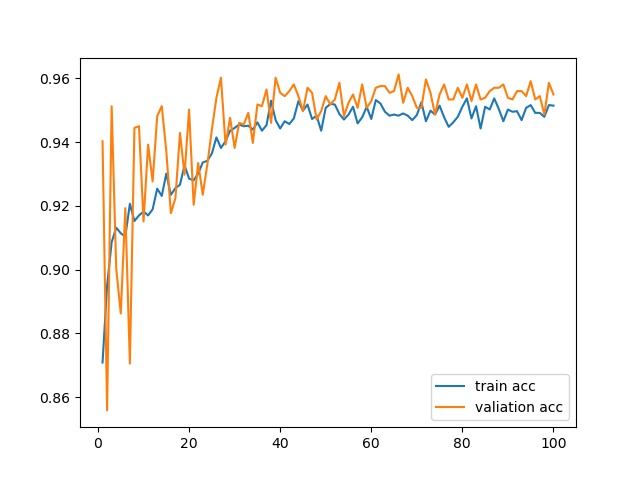}
\includegraphics[width=4.5cm]{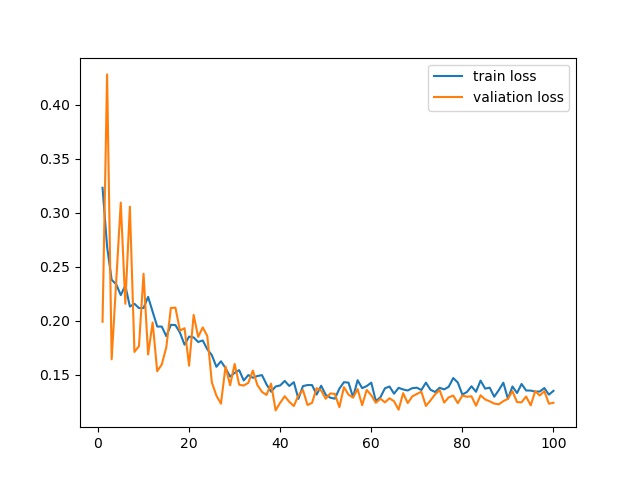}
\end{minipage}
}
\subfigure[DensenNet-121]{
\begin{minipage}[t]{0.22\linewidth}
\includegraphics[width=4.5cm]{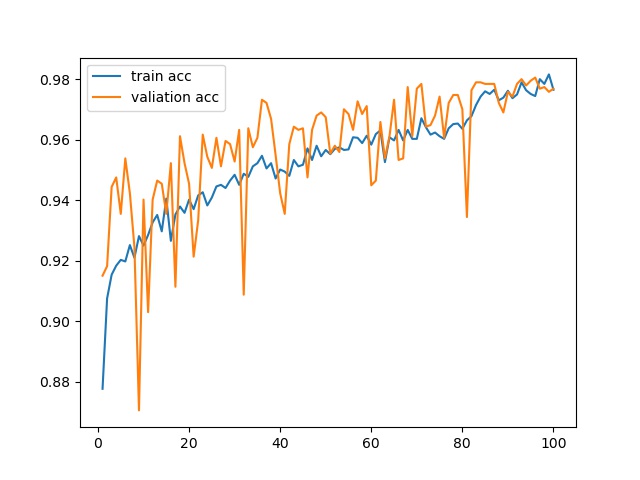}
\includegraphics[width=4.5cm]{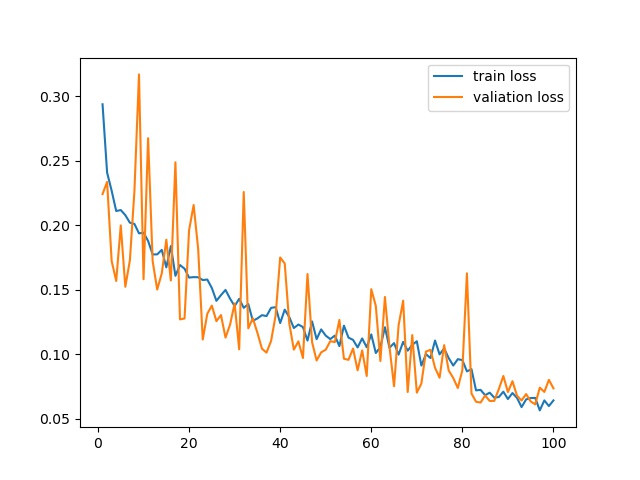}
\end{minipage}
}
\subfigure[Inception-V3]{
\begin{minipage}[t]{0.22\linewidth}
\includegraphics[width=4.5cm]{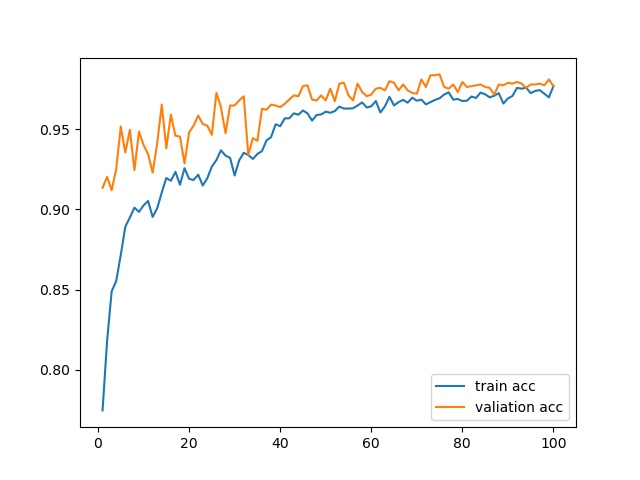}
\includegraphics[width=4.5cm]{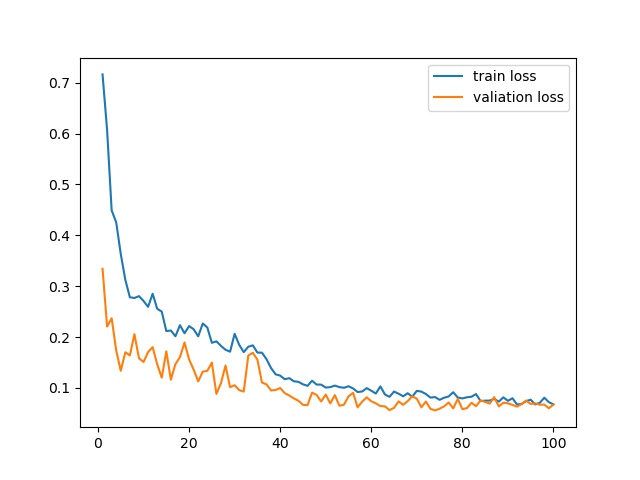}
\end{minipage}
}
\subfigure[MobileNet-V2]{
\begin{minipage}[t]{0.22\linewidth}
\includegraphics[width=4.5cm]{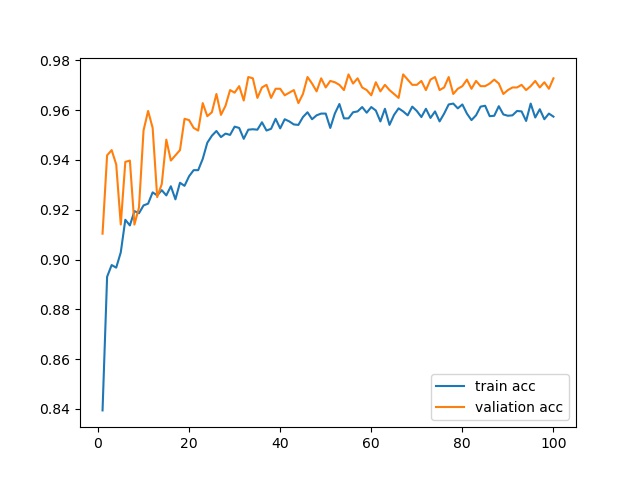}
\includegraphics[width=4.5cm]{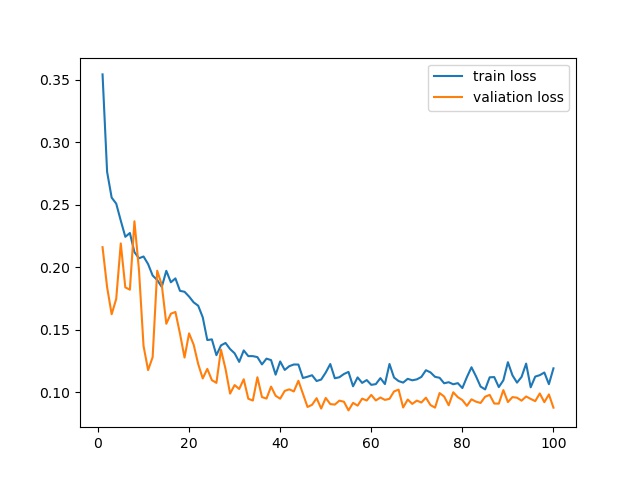}
\end{minipage}
}
\quad
\subfigure[ShuffleNet-V2]{
\begin{minipage}[t]{0.22\linewidth}
\includegraphics[width=4.5cm]{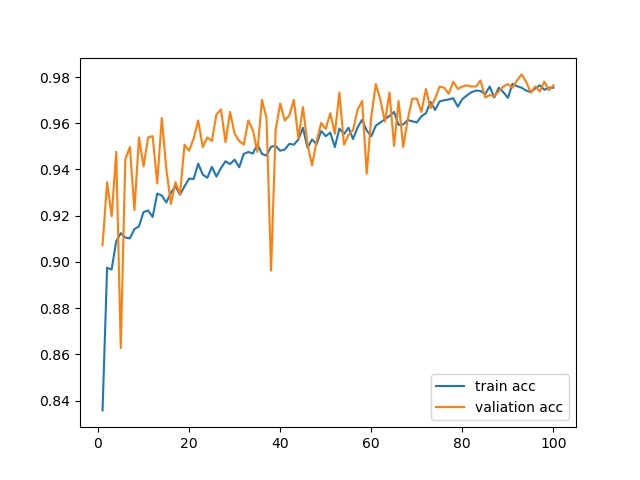}
\includegraphics[width=4.5cm]{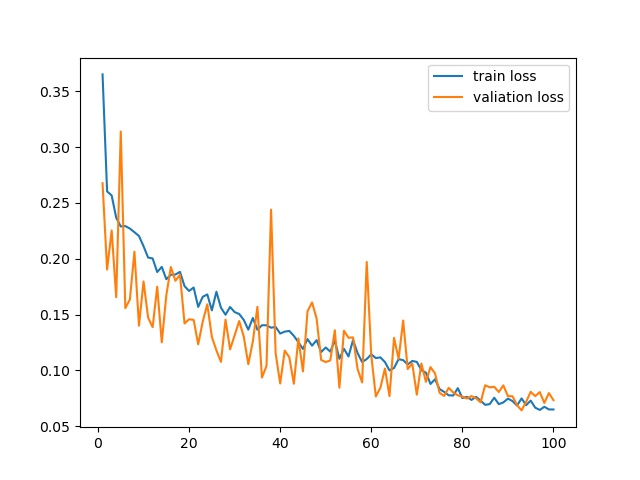}
\end{minipage}
}
\subfigure[Xception]{
\begin{minipage}[t]{0.22\linewidth}
\includegraphics[width=4.5cm]{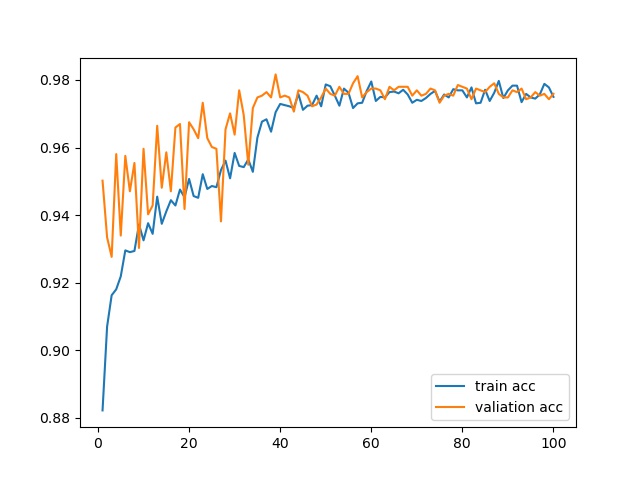}
\includegraphics[width=4.5cm]{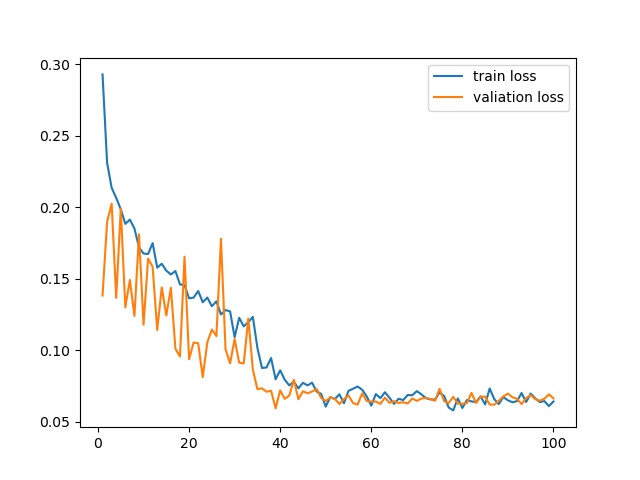}
\end{minipage}
}
\subfigure[ViT]{
\begin{minipage}[t]{0.22\linewidth}
\includegraphics[width=4.5cm]{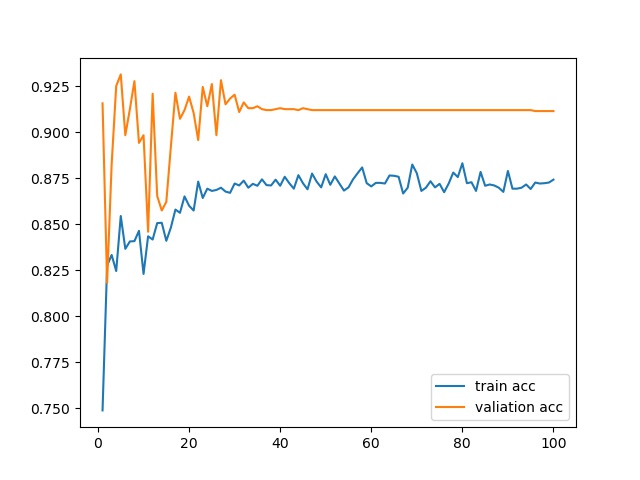}
\includegraphics[width=4.5cm]{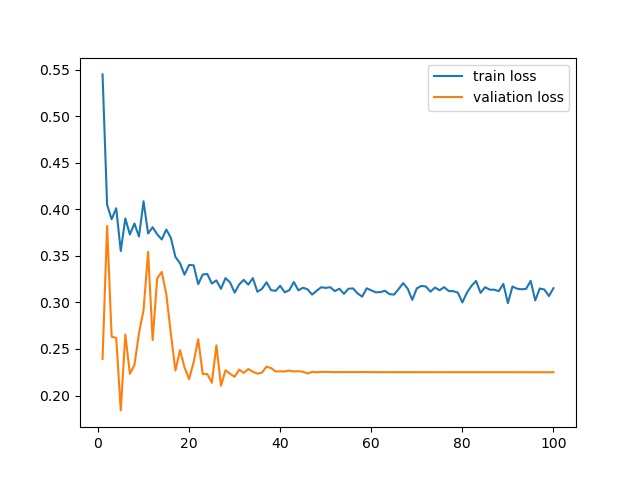}
\end{minipage}
}
\subfigure[BotNet]{
\begin{minipage}[t]{0.22\linewidth}
\includegraphics[width=4.5cm]{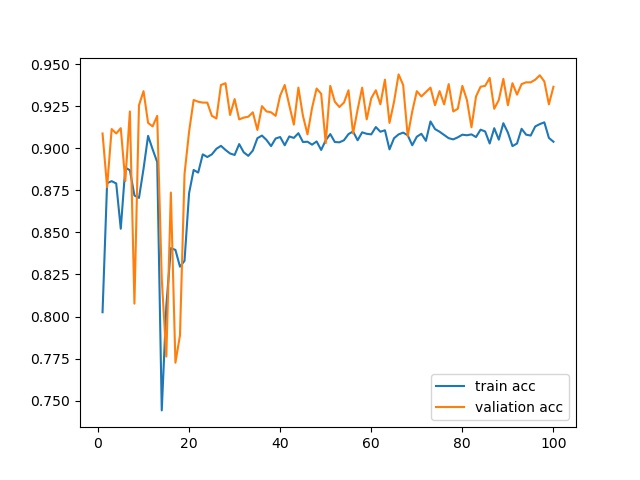}
\includegraphics[width=4.5cm]{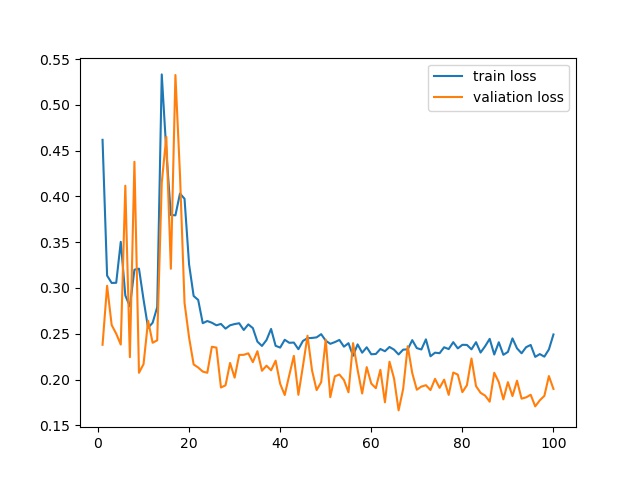}
\end{minipage}
}
\caption{Accuracy and loss curves of selected deep learning models. Here, (a) exhibits the training and validation accuracy and loss curves of AlexNet model. Similarly, (b) presents the accuracy and loss curves of VGG-16 model, (c) is generated on VGG-19,
(d) is on ResNet-50, (e) is on GoogleNet, (f) is on DensenNet-121, (g) is on Inception-V3, (h) is on MobileNet-V2, (i) is on ShuffleNet-V2, (j) is on Xception, (k) is on ViT, (l) is on BotNet, (m) is on DeiT-Tiny, (n) is on DeiT-Base, (o) is on T2T-t-ViT-24, (p) is on T2T-ViT-24.}
\label{FIG:3}
\end{figure*}
\addtocounter{figure}{-1}
\begin{figure*}
\centering
\addtocounter{subfigure}{12}
\subfigure[DeiT-Tiny]{
\begin{minipage}[t]{0.22\linewidth}
\includegraphics[width=4.5cm]{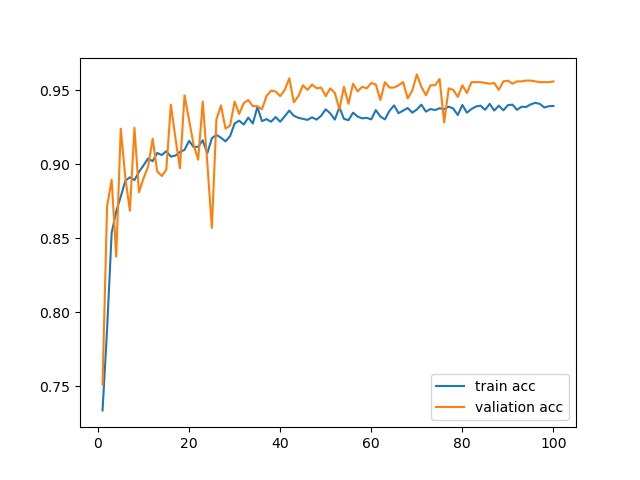}
\includegraphics[width=4.5cm]{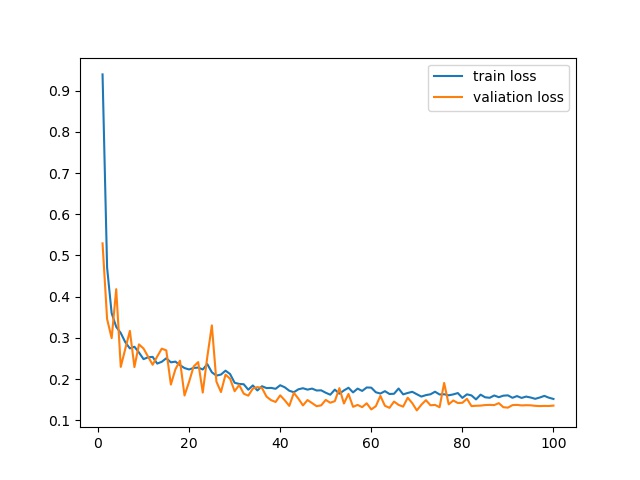}
\end{minipage}
}
\subfigure[DeiT-Base]{
\begin{minipage}[t]{0.22\linewidth}
\includegraphics[width=4.5cm]{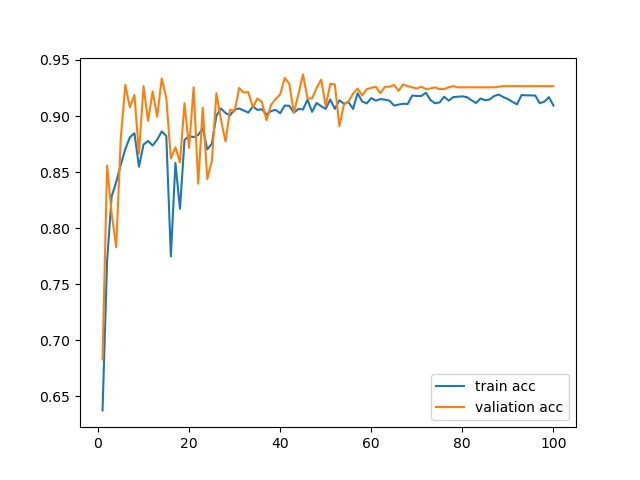}
\includegraphics[width=4.5cm]{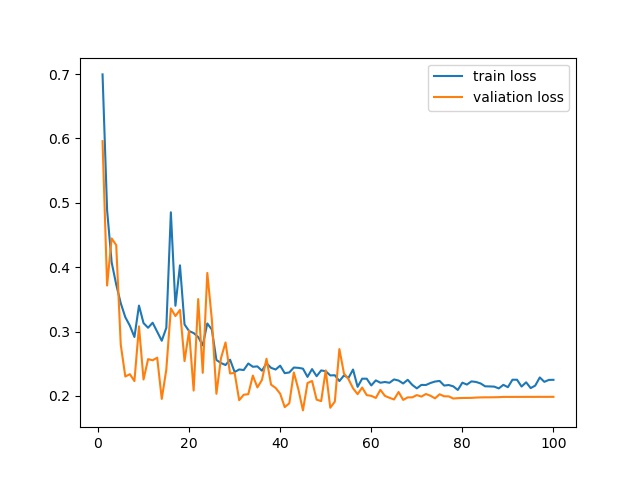}
\end{minipage}
}
\subfigure[T2T-ViT-t-24]{
\begin{minipage}[t]{0.22\linewidth}
\includegraphics[width=4.5cm]{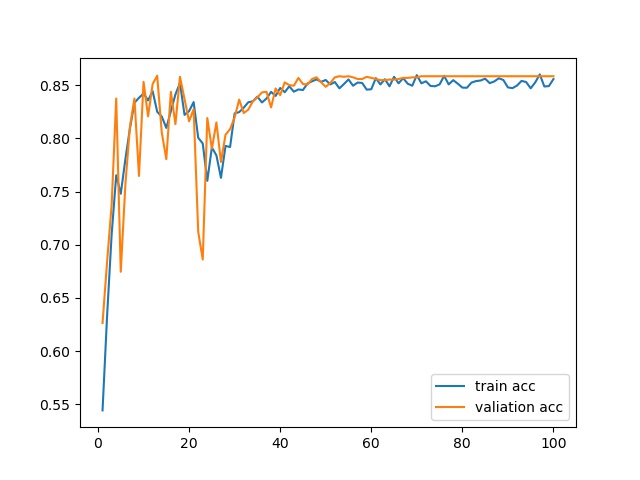}
\includegraphics[width=4.5cm]{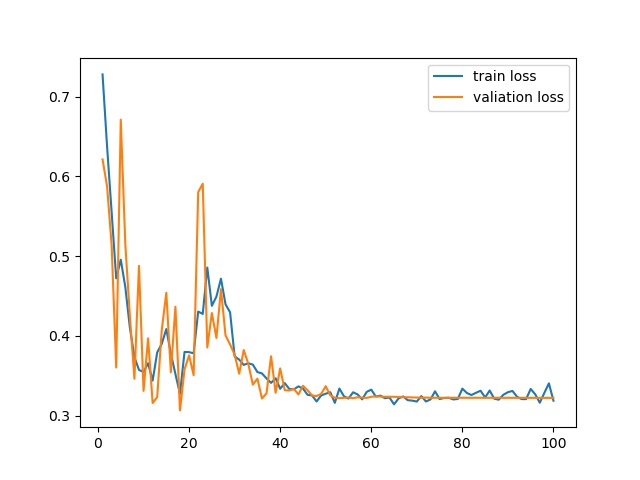}
\end{minipage}
}
\subfigure[T2T-ViT-24]{
\begin{minipage}[t]{0.22\linewidth}
\includegraphics[width=4.5cm]{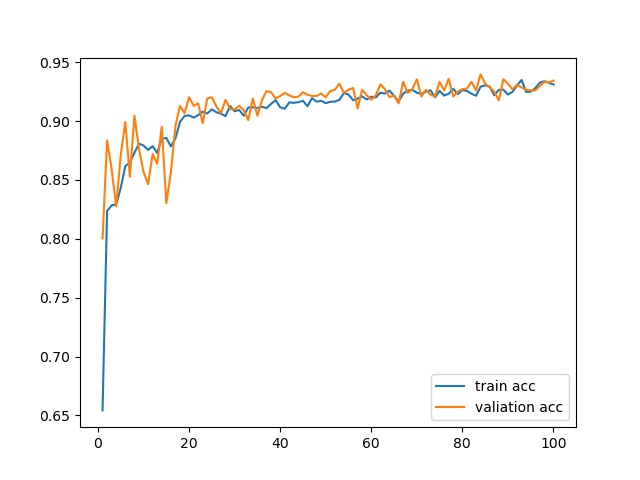}
\includegraphics[width=4.5cm]{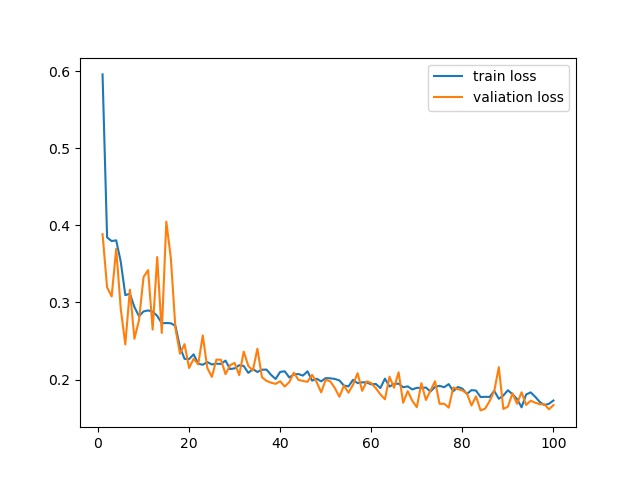}
\end{minipage}
}
\caption{Accuracy and loss curves of selected deep learning models. Here, (a) exhibits the training and validation accuracy and loss curves of AlexNet model. Similarly, (b) presents the accuracy and loss curves of VGG-16 model, (c) is generated on VGG-19,
(d) is on ResNet-50, (e) is on GoogleNet, (f) is on DensenNet-121, (g) is on Inception-V3, (h) is on MobileNet-V2, (i) is on ShuffleNet-V2, (j) is on Xception, (k) is on ViT, (l) is on BotNet, (m) is on DeiT-Tiny, (n) is on DeiT-Base, (o) is on T2T-ViT-t-24, (p) is on T2T-ViT-24.}
\label{FIG:3}
\end{figure*}

\subsubsection{Classification Performance on Test Set with Different Deep Learning Models}
In order to evaluate the performance of CNNs and VTs and evaluate the robustness of CNNs and VTs against conventional noise in the classification of tiny objects, in this section, we calculate the precision, recall, specificity, F1-score, accuracy and running time of each model in the original test dataset. Table~\ref{tbl6} shows the performance metrics of different deep learning models for identifying tiny objects (sperm and impurity) on the original test dataset. It can be seen from Table~\ref{tbl6} that DensenNet-121 and Inception-V3 obtained the highest precision and specificity value of 99.4 \% and 99.3 \% for sperm recognition, followed by the precision and specificity value of 99.2 \% and 99.1 \% for MobileNet-V2 and Xception, and the highest recall value of 97.8 \% for Xception. For impurity identification, DensenNet-121 and Inception-V3 have the highest recall value of 99.3 \%, followed by MobileNet-V2 and Xception recall value of 99.1 \%, and Xception get the highest specificity value of 97.8 \%. In addition, for sperm recognition, the lowest precision and recall value of T2T-ViT-t-24 is 89.6 \% and 82.5 \%, respectively, and the lowest specificity of ViT is 88.8 \%. For impurity identification, the lowest precision and specificity value of T2T-ViT-t-24 is 81.8 \% and 82.5 \%, respectively, and the lowest recall value of ViT is 88.8 \%.\par
Meanwhile, in all the test models for tiny objects (sperm and impurity) classification effect, the accuracy of Xception is the highest, and its value is 98.43 \%. The performance of DenseNet-121, Inception-V3 and ShuffleNet-V2 is located in the second echelon, with accuracy of 98.06 \%, 98.32 \% and 98.16 \% respectively. The performance of VGG-11, VGG-19, ResNet-50, ResNet-101 and MobileNet-V2 is located in the third echelon, with accuracy of 97.22 \%, 97.69 \%, 97.80 \%, 97.69 \% and 97.50 \%, respectively. Followed by VGG-16 and GoogleNet of the fourth echelon, with accuracy of 96.69 \% and 96.22 \% respectively. The members of the fifth echelon are BotNet, DeiT-Base, T2T-ViT-7 and T2T-ViT-12, the accuracy is 95.17 \%, 95.86 \%, 95.65 \% and 95.33 \% respectively. DeiT-Tiny and T2T-ViT-10 / 14 / 19 / 24 of the sixth echelon also receive considerable accuracy of about 94 \%. Subsequent echelons are ViT, AlexNet and T2T-ViT-t-19, with accuracy of 91.45 \%, 92.76 \% and 90.82 \%.  T2T-ViT-t-24 has the lowest accuracy of 85.62 \%. \par
In addition, T2T-ViT-t-24 has the longest testing time of 36 s, AlexNet has the shortest testing time of 6 s, and the good performance models Xception, Inception-V3, DenseNet-121 and ShuffleNet-V2 testing time are 15 s, 28 s, 34 s and 16 s, respectively.\par
\begin{figure*}
\centering
\subfigure[AlexNet]{
\begin{minipage}[t]{0.2\linewidth}
\includegraphics[width=4cm]{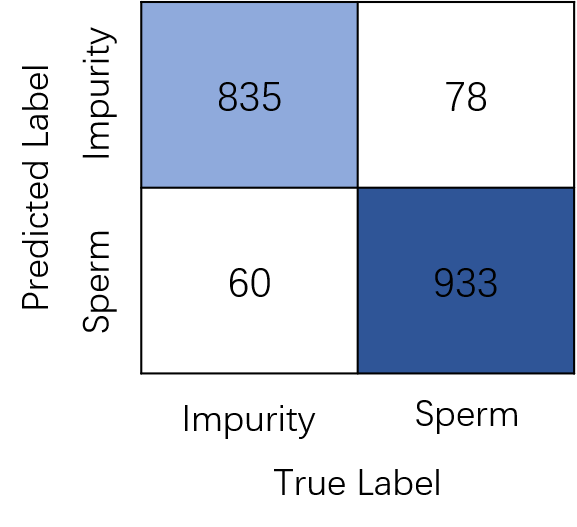}
\end{minipage}
}
\quad
\subfigure[VGG-16]{
\begin{minipage}[t]{0.2\linewidth}
\includegraphics[width=4cm]{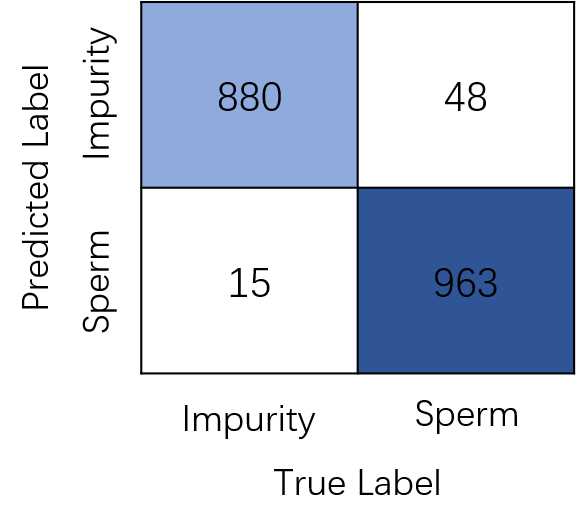}
\end{minipage}
}
\quad
\subfigure[VGG-19]{
\begin{minipage}[t]{0.2\linewidth}
\includegraphics[width=4cm]{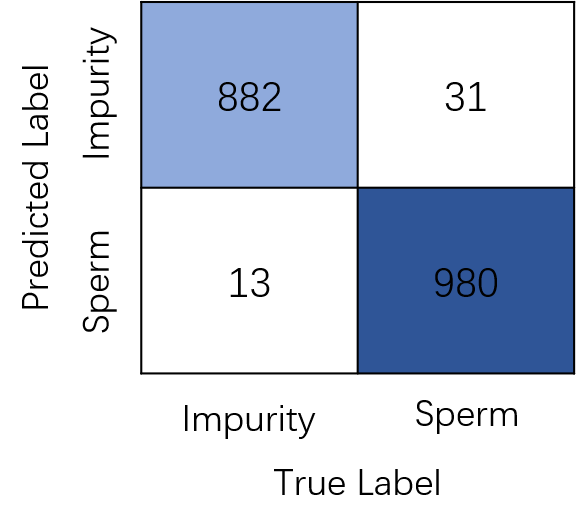}
\end{minipage}
}
\quad
\subfigure[ResNet-50]{
\begin{minipage}[t]{0.2\linewidth}
\includegraphics[width=4cm]{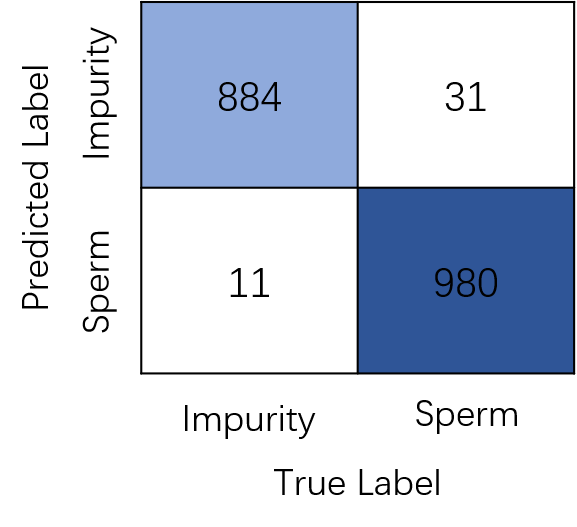}
\end{minipage}
}
\quad
\subfigure[ResNet-101]{
\begin{minipage}[t]{0.2\linewidth}
\includegraphics[width=4cm]{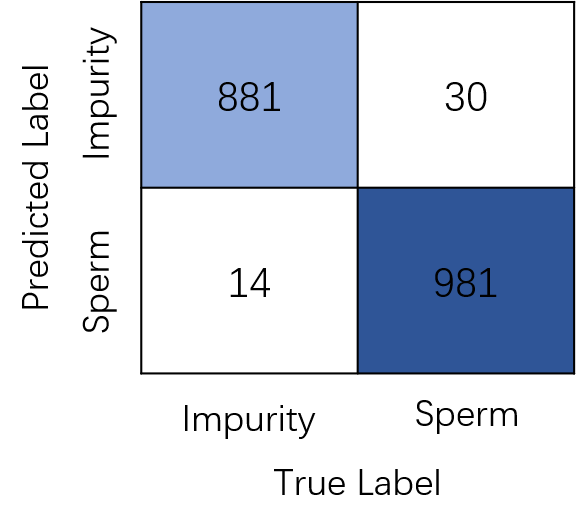}
\end{minipage}
}
\quad
\subfigure[GoogleNet]{
\begin{minipage}[t]{0.2\linewidth}
\includegraphics[width=4cm]{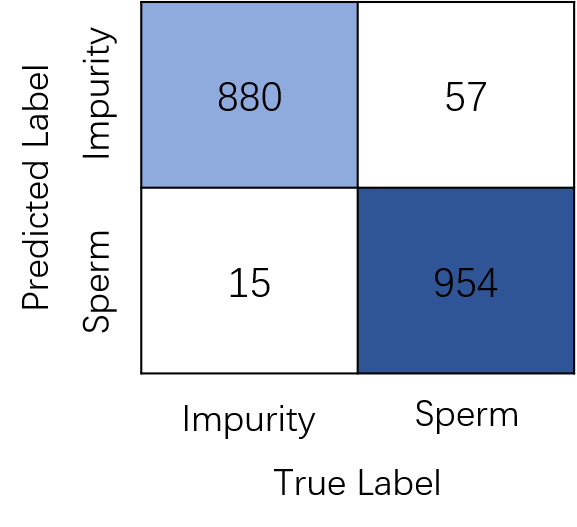}
\end{minipage}
}
\quad
\subfigure[DenseNet-121]{
\begin{minipage}[t]{0.2\linewidth}
\includegraphics[width=4cm]{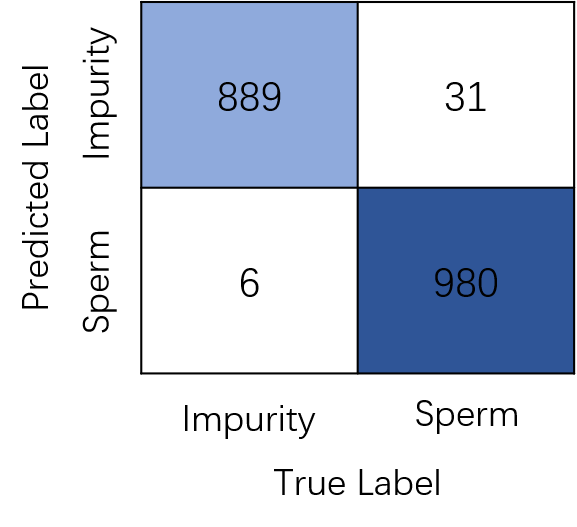}
\end{minipage}
}
\quad
\subfigure[Inception-V3]{
\begin{minipage}[t]{0.2\linewidth}
\includegraphics[width=4cm]{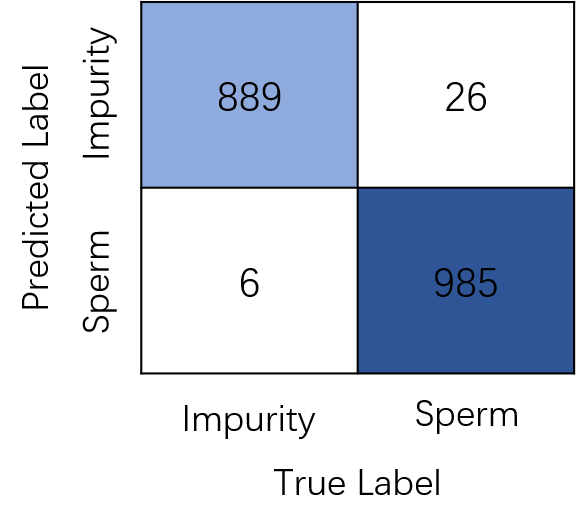}
\end{minipage}
}
\quad
\subfigure[MobileNet-V2]{
\begin{minipage}[t]{0.2\linewidth}
\includegraphics[width=4cm]{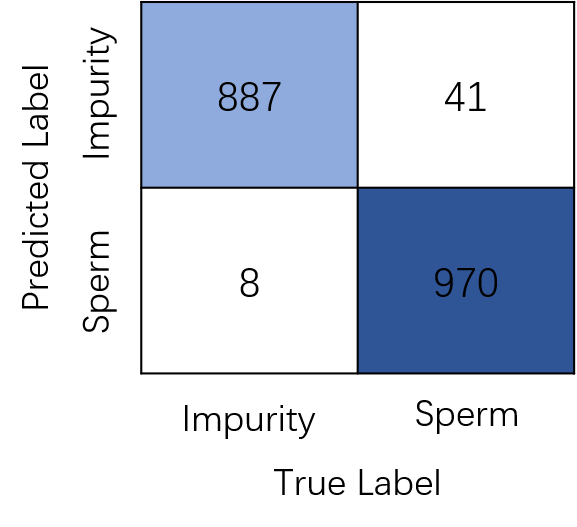}
\end{minipage}
}
\quad
\subfigure[ShuffleNet-V2]{
\begin{minipage}[t]{0.2\linewidth}
\includegraphics[width=4cm]{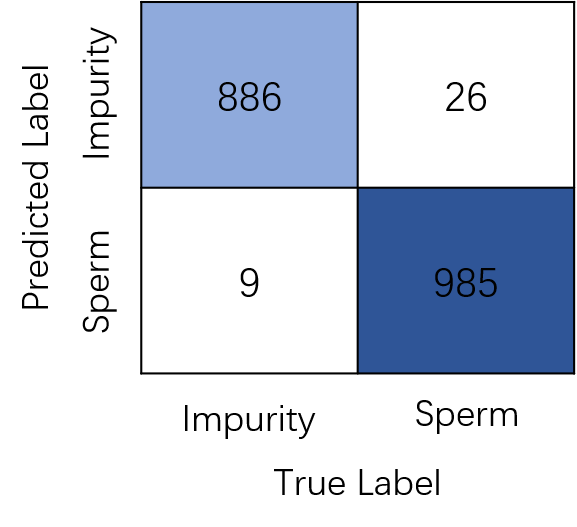}
\end{minipage}
}
\quad
\subfigure[Xception]{
\begin{minipage}[t]{0.2\linewidth}
\includegraphics[width=4cm]{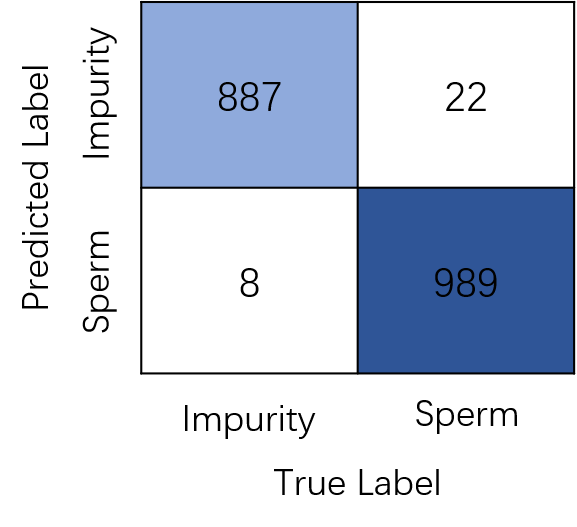}
\end{minipage}
}
\quad
\subfigure[ViT]{
\begin{minipage}[t]{0.2\linewidth}
\includegraphics[width=4cm]{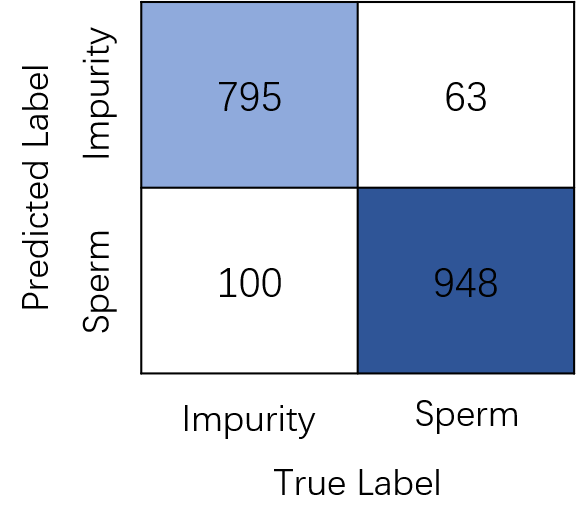}
\end{minipage}
}
\quad
\subfigure[BotNet]{
\begin{minipage}[t]{0.2\linewidth}
\includegraphics[width=4cm]{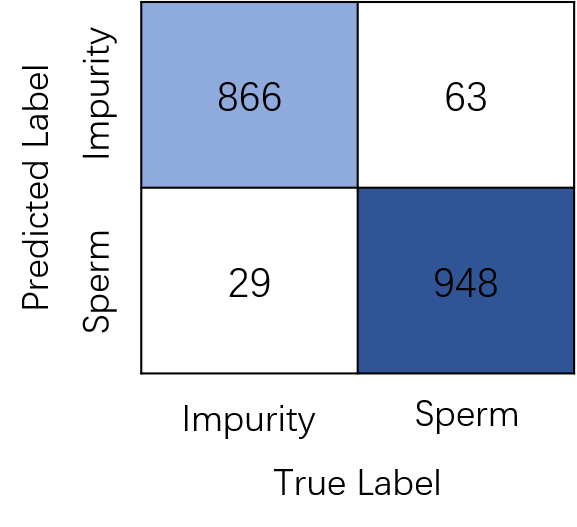}
\end{minipage}
}
\quad
\subfigure[DeiT-Base]{
\begin{minipage}[t]{0.2\linewidth}
\includegraphics[width=4cm]{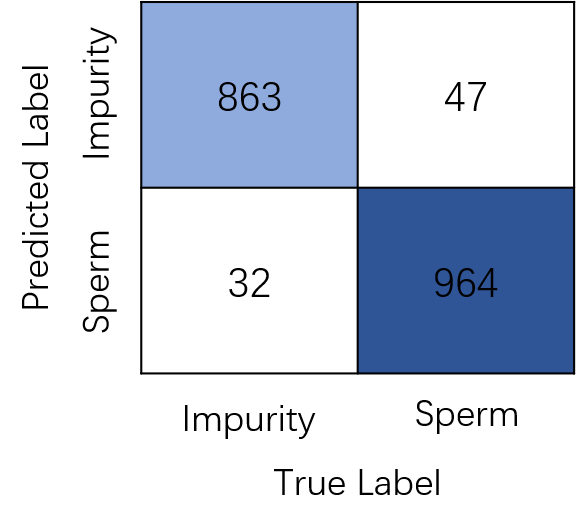}
\end{minipage}
}
\quad
\subfigure[DeiT-Tiny]{
\begin{minipage}[t]{0.2\linewidth}
\includegraphics[width=4cm]{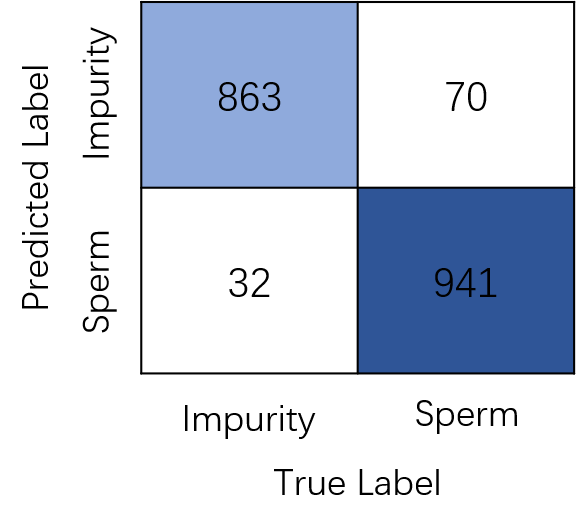}
\end{minipage}
}
\quad
\subfigure[T2T-ViT-t-19]{
\begin{minipage}[t]{0.2\linewidth}
\includegraphics[width=4cm]{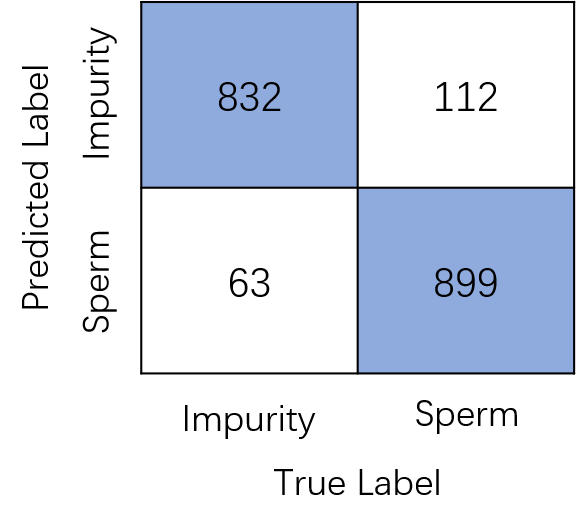}
\end{minipage}
}
\quad
\subfigure[T2T-ViT-t-24]{
\begin{minipage}[t]{0.2\linewidth}
\includegraphics[width=4cm]{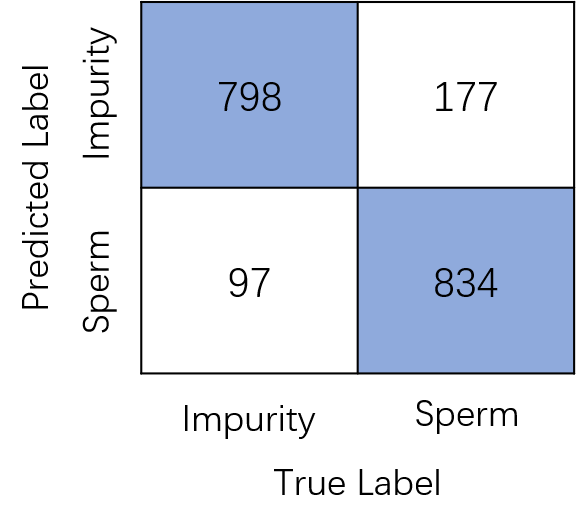}
\end{minipage}
}
\quad
\subfigure[T2T-ViT-7]{
\begin{minipage}[t]{0.2\linewidth}
\includegraphics[width=4cm]{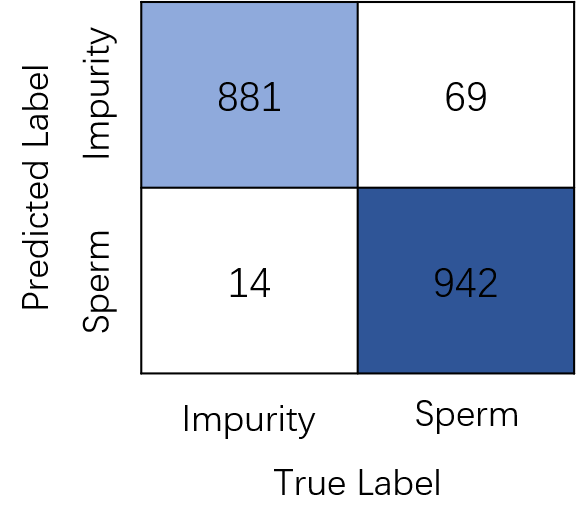}
\end{minipage}
}
\quad
\subfigure[T2T-ViT-19]{
\begin{minipage}[t]{0.2\linewidth}
\includegraphics[width=4cm]{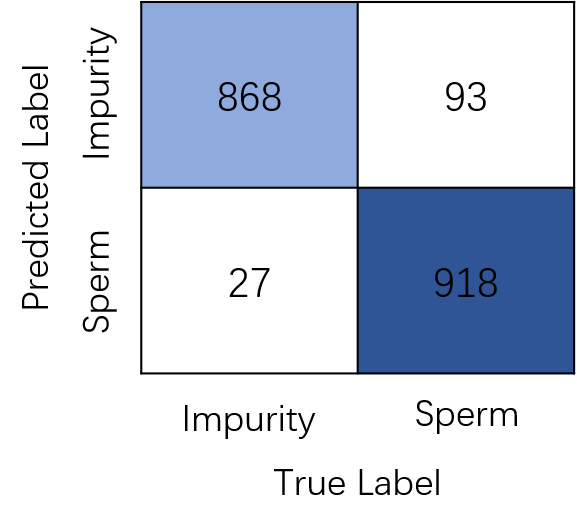}
\end{minipage}
}
\quad
\subfigure[T2T-ViT-24]{
\begin{minipage}[t]{0.2\linewidth}
\includegraphics[width=4cm]{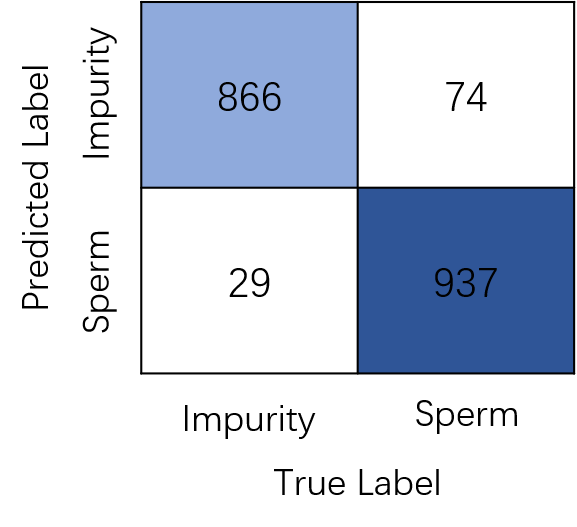}
\end{minipage}
}

\caption{The resulted confusion matrix of different deep learning models. Here, (a) is the confusion matrix of AlexNet generated on test dataset. Similarly, (b) is generated on VGG-16 model, (c) is generated on VGG-19, (d) is on ResNet-50, (e) is on ResNet-101 (f) is on GoogleNet, (g) is on DensenNet-121, (h) is on Inception-V3, (i) is on MobileNet-V2, (j) is on ShuffleNet-V2, (k) is on Xception, (l) is on ViT, (m) is on BotNet, (n) is on DeiT-Base, (o) is on DeiT-Tiny, (p) is on T2T-ViT-t-19, (q) is on T2T-ViT-t-24, (r) is on T2T-ViT-7, (s) is on T2T-ViT-19, (t) is on T2T-ViT-24.}
\label{FIG:4}
\end{figure*}

Fig.~\ref{FIG:4} depicts many confusion matrices generated by the test dataset to more intuitively show the performance of CNN and VT models for tiny object classification. Table~\ref{tbl6} shows that the highest accuracy model is Xception and the lowest accuracy is T2T-ViT-t-24. Therefore, when we focus on the confusion matrix of Xception, we can know that Xception accurately identifies 989 images as sperm class, and 22 images are wrongly divided into impurity class. At the same time, 887 images are accurately identified as impurity class, and only 8 images are wrongly identified as sperm class. Meanwhile, the lowest accuracy model is T2T-ViT-t-24. We observe its confusion matrix and find that 834 sperm images are successfully identified, but 177 images are also wrongly identified. At the same time, 798 impurity images are successfully identified, and 97 images are wrongly identified. \par
Furthermore, the precision, specificity and accuracy of Inception-V3 and DenseNet-121 are also good, so we analyze the confusion matrix of these two models. We focus on the confusion matrix of Inception-V3, 985 sperm images are correctly classified into sperm class, 26 sperm images are mistakenly classified into impurity class, 889 impurity images are correctly classified into impurity class, and 6 impurity images are mistakenly classified into sperm class. In addition, from the confusion matrix of DenseNet-121, we can know that 980 sperm images are correctly classified into sperm class, 31 sperm images are incorrectly classified into impurity class, and 889 impurity images are correctly classified into impurity class, and 6 impurity images are mistakenly classified as sperms.\par
Through the comprehensive analysis of the above various evaluation metrics, we can see that for various CNN and VT models tested in this paper, In the classification of tiny objects (sperm and impurity), the evaluation metrics obtained by DenseNet-121, Xception and Inception-V3 are the top. Therefore, it is obvious that Xception, DenseNet-121 and Inception-V3 models are more suitable for the classification of tiny objects (sperm and impurity).
\begin{table*}
\caption{Performance of different deep learning models on original test set. (In [\%].)}
\label{tbl6}
\begin{tabular}{@{}clllllccll@{}}
\toprule
\multicolumn{1}{l}{Model}      & Class    & Precision & Recall & Specificity & F1-Score & \multicolumn{1}{l}{Accuracy} & \multicolumn{1}{l}{Time(s)} &  &  \\
\midrule
\multirow{2}{*}{AlexNet}       & Sperm    & 94.00     & 92.30  & 93.30       & 93.10    & \multirow{2}{*}{92.76}           & \multirow{2}{*}{6}          &  &  \\
                               & Impurity & 91.50     & 93.30  & 92.30       & 92.40    &                                  &                             &  &  \\
\multirow{2}{*}{VGG-11}        & Sperm    & 98.30     & 96.40  & 98.10       & 97.30    & \multirow{2}{*}{97.22}           & \multirow{2}{*}{13}         &  &  \\
                               & Impurity & 96.10     & 98.10  & 96.40       & 97.10    &                                  &                             &  &  \\
\multirow{2}{*}{VGG-16}        & Sperm    & 98.50     & 95.30  & 98.30       & 96.90    & \multirow{2}{*}{96.69}           & \multirow{2}{*}{19}         &  &  \\
                               & Impurity & 94.80     & 98.30  & 95.30       & 96.50    &                                  &                             &  &  \\
\multirow{2}{*}{VGG-19}        & Sperm    & 98.70     & 96.90  & 98.50       & 97.80    & \multirow{2}{*}{97.69}           & \multirow{2}{*}{31}         &  &  \\
                               & Impurity & 96.60     & 98.50  & 96.90       & 97.50    &                                  &                             &  &  \\
\multirow{2}{*}{ResNet-50}     & Sperm    & 98.90     & 96.90  & 98.80       & 97.90    & \multirow{2}{*}{97.80}           & \multirow{2}{*}{17}         &  &  \\
                               & Impurity & 96.60     & 98.80  & 96.90       & 97.70    &                                  &                             &  &  \\
\multirow{2}{*}{ResNet-101}    & Sperm    & 98.60     & 97.00  & 98.40       & 97.80    & \multirow{2}{*}{97.69}           & \multirow{2}{*}{28}         &  &  \\
                               & Impurity & 96.70     & 98.40  & 97.00       & 97.50    &                                  &                             &  &  \\
\multirow{2}{*}{GoogleNet}     & Sperm    & 98.50     & 94.40  & 98.30       & 96.40    & \multirow{2}{*}{96.22}           & \multirow{2}{*}{17}         &  &  \\
                               & Impurity & 93.90     & 98.30  & 94.40       & 96.00    &                                  &                             &  &  \\
\multirow{2}{*}{DenseNet-121}  & Sperm    & 99.40     & 96.90  & 99.30       & 98.10    & \multirow{2}{*}{98.06}           & \multirow{2}{*}{34}         &  &  \\
                               & Impurity & 96.60     & 99.30  & 96.90       & 97.90    &                                  &                             &  &  \\
\multirow{2}{*}{Inception-V3}  & Sperm    & 99.40     & 97.40  & 99.30       & 98.40    & \multirow{2}{*}{98.32}           & \multirow{2}{*}{28}         &  &  \\
                               & Impurity & 97.20     & 99.30  & 97.40       & 98.20    &                                  &                             &  &  \\
\multirow{2}{*}{MobileNet-V2}  & Sperm    & 99.20     & 95.90  & 99.10       & 97.50    & \multirow{2}{*}{97.43}           & \multirow{2}{*}{14}         &  &  \\
                               & Impurity & 95.60     & 99.10  & 95.90       & 97.30    &                                  &                             &  &  \\
\multirow{2}{*}{ShuffleNet-V2} & Sperm    & 99.10     & 97.40  & 99.00       & 98.20    & \multirow{2}{*}{98.16}           & \multirow{2}{*}{16}         &  &  \\
                               & Impurity & 97.10     & 99.00  & 97.40       & 98.00    &                                  &                             &  &  \\
\multirow{2}{*}{Xception}      & Sperm    & 99.20     & 97.80  & 99.10       & 98.50    & \multirow{2}{*}{98.43}           & \multirow{2}{*}{15}         &  &  \\
                               & Impurity & 97.60     & 99.10  & 97.80       & 98.30    &                                  &                             &  &  \\
\multirow{2}{*}{ViT}           & Sperm    & 90.50     & 93.80  & 88.80       & 92.10    & \multirow{2}{*}{91.45}           & \multirow{2}{*}{16}         &  &  \\
                               & Impurity & 92.70     & 88.80  & 93.80       & 90.70    &                                  &                             &  &  \\
\multirow{2}{*}{BotNet}        & Sperm    & 97.00     & 93.80  & 96.80       & 95.40    & \multirow{2}{*}{95.17}           & \multirow{2}{*}{18}         &  &  \\
                               & Impurity & 93.20     & 96.80  & 93.80       & 95.00    &                                  &                             &  &  \\
\multirow{2}{*}{DeiT-Base}     & Sperm    & 96.80     & 95.40  & 96.40       & 96.10    & \multirow{2}{*}{95.86}           & \multirow{2}{*}{15}         &  &  \\
                               & Impurity & 94.80     & 96.40  & 95.40       & 95.60    &                                  &                             &  &  \\
\multirow{2}{*}{DeiT-Tiny}     & Sperm    & 96.70     & 93.10  & 96.40       & 94.90    & \multirow{2}{*}{94.65}           & \multirow{2}{*}{34}         &  &  \\
                               & Impurity & 92.50     & 96.40  & 93.10       & 94.40    &                                  &                             &  &  \\
\multirow{2}{*}{T2T-ViT-t-19}  & Sperm    & 93.50     & 88.90  & 93.00       & 91.10    & \multirow{2}{*}{90.82}           & \multirow{2}{*}{31}         &  &  \\
                               & Impurity & 88.10     & 93.00  & 88.90       & 90.50    &                                  &                             &  &  \\
\multirow{2}{*}{T2T-ViT-t-24}  & Sperm    & 89.60     & 82.50  & 89.20       & 85.90    & \multirow{2}{*}{85.62}           & \multirow{2}{*}{36}         &  &  \\
                               & Impurity & 81.80     & 89.20  & 82.50       & 85.30    &                                  &                             &  &  \\
\multirow{2}{*}{T2T-ViT-7}     & Sperm    & 98.50     & 93.20  & 98.40       & 95.80    & \multirow{2}{*}{95.65}           & \multirow{2}{*}{14}         &  &  \\
                               & Impurity & 92.70     & 98.40  & 93.20       & 95.50    &                                  &                             &  &  \\
\multirow{2}{*}{T2T-ViT-10}    & Sperm    & 96.50     & 93.00  & 96.20       & 94.70    & \multirow{2}{*}{94.49}           & \multirow{2}{*}{16}         &  &  \\
                               & Impurity & 92.40     & 96.20  & 93.00       & 94.30    &                                  &                             &  &  \\
\multirow{2}{*}{T2T-ViT-12}    & Sperm    & 97.30     & 93.80  & 97.10       & 95.50    & \multirow{2}{*}{95.33}           & \multirow{2}{*}{19}         &  &  \\
                               & Impurity & 93.20     & 97.10  & 93.80       & 95.10    &                                  &                             &  &  \\
\multirow{2}{*}{T2T-ViT-14}    & Sperm    & 95.80     & 93.80  & 95.30       & 94.80    & \multirow{2}{*}{94.49}           & \multirow{2}{*}{20}         &  &  \\
                               & Impurity & 93.10     & 95.30  & 93.80       & 94.20    &                                  &                             &  &  \\
\multirow{2}{*}{T2T-ViT-19}    & Sperm    & 97.10     & 90.80  & 97.00       & 93.80    & \multirow{2}{*}{93.70}           & \multirow{2}{*}{25}         &  &  \\
                               & Impurity & 90.30     & 97.00  & 90.80       & 93.50    &                                  &                             &  &  \\
\multirow{2}{*}{T2T-ViT-24}    & Sperm    & 97.00     & 92.70  & 96.80       & 94.80    & \multirow{2}{*}{94.60}           & \multirow{2}{*}{30}         &  &  \\
                               & Impurity & 92.10     & 96.80  & 92.70       & 94.40    &                                  &                             &  & \\
\bottomrule
\end{tabular}
\end{table*}

\subsubsection{Classification Performance on Conventional Noises in Test Set of Different Deep Learning Models}
In this section, we mainly test a variety of different conventional noise datasets under CNN and VT models to obtain evaluation metrics, including accuracy, specificity, F1-score and accuracy. In order to intuitively compare the robustness of each model to conventional noises, the evaluation metrics obtained from the original test dataset under each model are made into a histogram (as shown in Fig.~\ref{FIG:8}).  At the same time, each model in this paper tests the conventional noise test datasets, and the obtained evaluation metrics are made into histograms (as shown in Fig.~\ref{FIG:9} \~{} Fig.~\ref{FIG:16}).
\paragraph{Evaluation of different deep learning models under Ga\-ussian noise}~{}

First, we compare the histogram (as shown in Fig.~\ref{FIG:8}) of the original test set with the histogram (as shown in Fig.~\ref{FIG:9} (a)) of Gaussian noise with the mean value of 0 and the variance of 0.01, which can be seen from the two figures: from the columnar trend of evaluation metrics in the overall histogram, DenseNet-121, Inception-V3, ShuffluNet-V2, Xception and ViT have little changes. When we focus on these five models, the change of each evaluation metric of DenseNet-121 are about 4.6 \%, that of Inception-V3 are about 4.25 \%, that of ShuffluNet-V2 are about 4.8 \%, that of Xception are about 4.4 \%, and that of ViT are about 1.6 \%.\par

Then, we compare the histogram (as shown in Fig.~\ref{FIG:8}) of the original test set with the histogram (as shown in Fig.~\ref{FIG:9} (b)) of Gaussian noise with the mean value of 0 and the variance of 0.05, which can be seen from the two figures: from the columnar trend of evaluation metrics in the overall histogram, the Inception-V3, ViT and BotNet have little changes. When we focus on these three models, the change of each evaluation metric of Inception-V3 are about 20 \%, that of ViT are about 3 \%, that of BotNet are about 10 \%.\par
Next, we compare the histogram (as shown in Fig.~\ref{FIG:8}) of the original test set with the histogram (as shown in Fig.~\ref{FIG:9} (c))  of Gaussian noise with the mean value of 0.2 and the variance of 0.01, which can be seen from the two figures: under the influence of this noise, the evaluation metrics in each model change greatly. Compared with the performance of all models, the change of VGG-11, VGG-16 and T2T-ViT-t-19 are slightly flat. However, if we focus on these three models, we can find that the sperm recall , F1-score and impurity specificity of VGG-11 and VGG-16 have changed significantly, with a range of more than 30 \%. At the same time, the impurity F1-score and precision have also changed by about 20 \%. In addition, many evaluation metrics obtained from T2T-ViT-t-19 also change greatly. The change of sperm recall and F1-score reached more than 25 \% and 35 \%, and the change of impurity precision, specificity and F1-score reached about 25 \%, 35 \% and 15 \%, respectively. The overall accuracy change about 20 \%. \par

Finally, we compare the histogram (as shown in Fig.~\ref{FIG:8}) of the original test set with the histogram (as shown in Fig.~\ref{FIG:9} (d)) of Gaussian noise with the mean value of 0.2 and the variance of 0.05, which can be seen from the two figures: under the influence of this noise, the evaluation metrics in each model change greatly. Compared with the performance of all models, the change of VGG-16 and T2T-ViT-t-19 are slightly flat. However, when we focus on these two models, we can find that the sperm recall ,F1-score and impurity specificity obtained by VGG-16 have changed more than 50 \%, and the impurity precision, F1-score and accuracy change reach more than 30 \%. In addition, T2T-ViT-t-19 has no change in sperm specificity and impurity recall, and the sperm recall change by 8 \%. The other evaluation metrics reach more than 30 \%, and even the change in sperm recall, F1-score and impurity specificity reached more than 50 \%. \par
Therefore, under the influence of  Gaussian noise with a mean value of 0 and a variance of 0.01 or 0.05, in the CNN and VT models used in this paper, ViT has strong anti-noise robustness for the classification of tiny object (sperm and impurity) image dataset. However, under the influence of   Gaussian noise with a mean value of 0.2 and a variance of 0.01 or 0.05, all models in this paper have poor anti-noise robustness for the classification of tiny object (sperm and impurity) image dataset.\par

\paragraph{Evaluation of different deep learning models under sp\-eckle noise}~{}

First, we compare the histogram (as shown in Fig.~\ref{FIG:8}) of the original test set with the histogram (as shown in Fig.~\ref{FIG:10} (a)) of speckle noise with the mean value of 0 and the variance of 0.1, which can be seen from the two figures: from the columnar trend of evaluation metrics in the overall histogram, the change of AlexNet, VGG-16, ViT and BotNet are relatively small. When we focus on these four models, we can find that the sperm recall, impurity precision and impurity specificity under AlexNet change greatly, reaching about 18 \%, 15 \% and 18 \% respectively, and other evaluation metrics change about 10 \%. Under the VGG-16, the sperm recall, impurity precision and impurity specificity change greatly to about 20 \%, the sperm F1-score, impurity F1-score and accuracy change about 10 \%, and the other evaluation metrics change little. Under ViT, the change of sperm recall, impurity accuracy and impurity specificity reach about 6 \%, and the change of other evaluation metrics is only about 2 \%. Under the BotNet, sperm precision, sperm specificity and impurity recall change about 10 \%, sperm F1-score, impurity F1-score and accuracy change about 7 \%, and the remaining evaluation metrics change about 4 \%.\par
Then, we compare the histogram (as shown in Fig.~\ref{FIG:8}) of the original test set with the histogram (as shown in Fig.~\ref{FIG:10} (b)) of speckle noise with the mean value of 0 and the variance of 0.05, which can be seen from the two figures: from the columnar trend of evaluation metrics in the overall histogram, the change of VGG-16, ViT and BotNet are relatively small. Under the VGG-16, the sperm recall, impurity precision and impurity recall change about 8 \%, the sperm F1-score, impurity F1-score and accuracy change about 5 \%, and the other evaluation metrics change about 1.5 \%. Under ViT, the sperm recall, impurity precision and impurity recall change by about 4 \%, the sperm F1-score, impurity F1-score and accuracy change by about 1 \%, the sperm precision change by about 2 \%, and the other evaluation metrics change by about 3 \%. Under BotNet, the change of sperm specificity and impurity recall are about 9 \%, the change of sperm F1-score and impurity F1-score is about 4 \% and 5 \%, respectively, the change of sperm accuracy is about 7 \%, the change of accuracy is about 4 \%, and the change of other evaluation metrics are little.\par 
Therefore, under the influence of speckle noise with a mean value of 0 and a variance of 0.1 or 0.05, in the CNN and VT models used in this paper, ViT has strong anti-noise robustness for the classification of tiny object (sperm and impurity) image dataset.
\begin{figure*}
\centering
\subfigure[AlexNet]{
\begin{minipage}[t]{0.23\linewidth}
\includegraphics[width=4cm]{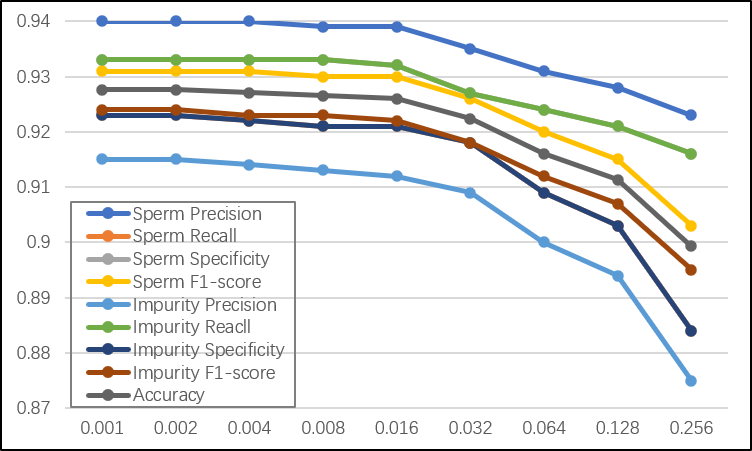}
\end{minipage}
}
\subfigure[VGG-11]{
\begin{minipage}[t]{0.23\linewidth}
\includegraphics[width=4cm]{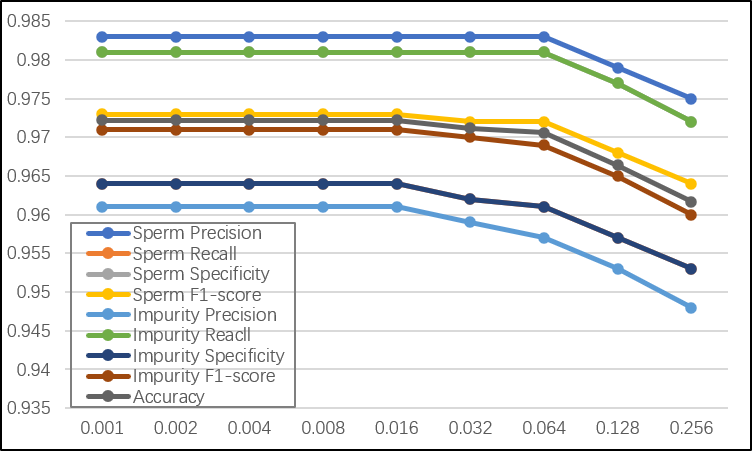}
\end{minipage}
}
\subfigure[VGG-16]{
\begin{minipage}[t]{0.23\linewidth}
\includegraphics[width=4cm]{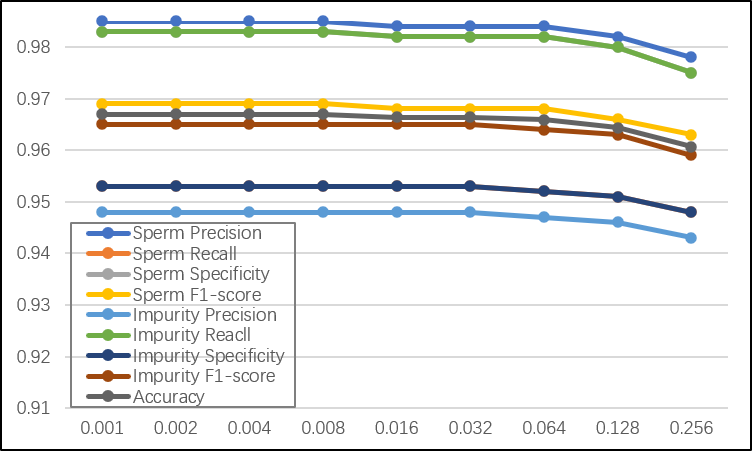}
\end{minipage}
}
\subfigure[VGG-19]{
\begin{minipage}[t]{0.23\linewidth}
\includegraphics[width=4cm]{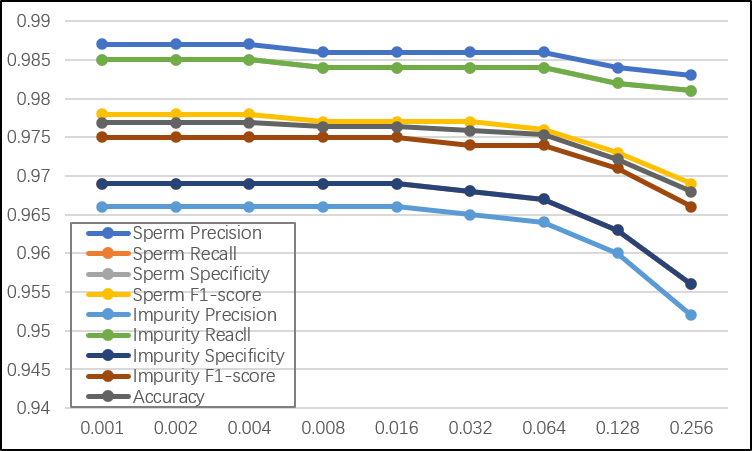}
\end{minipage}
}
\quad
\subfigure[ResNet-50]{
\begin{minipage}[t]{0.23\linewidth}
\includegraphics[width=4cm]{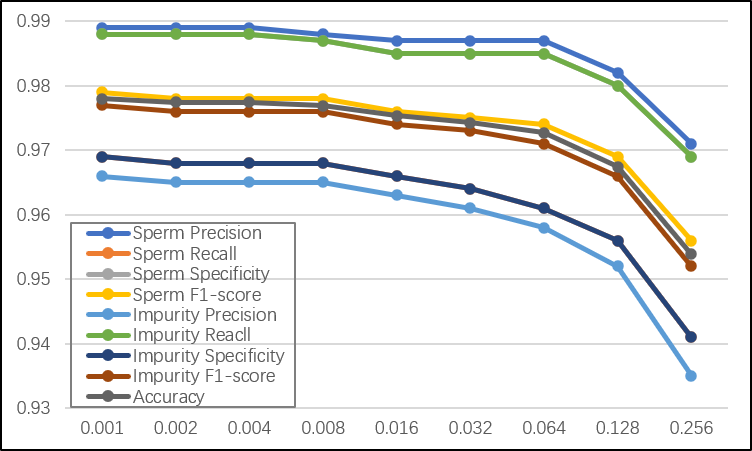}
\end{minipage}
}
\subfigure[ResNet-101]{
\begin{minipage}[t]{0.23\linewidth}
\includegraphics[width=4cm]{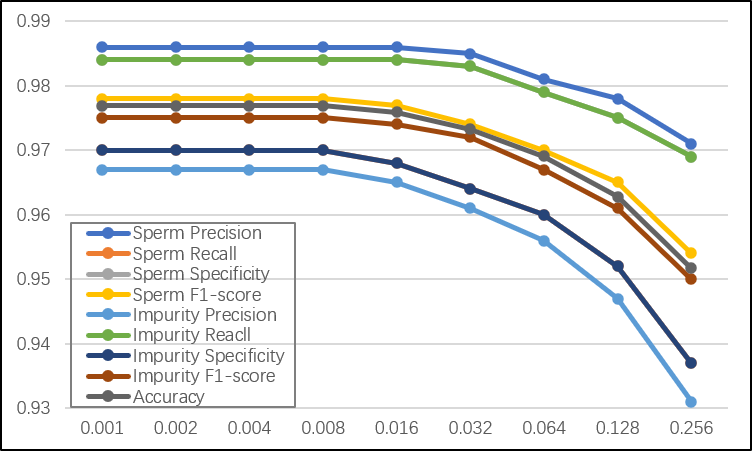}
\end{minipage}
}
\subfigure[GoogleNet]{
\begin{minipage}[t]{0.23\linewidth}
\includegraphics[width=4cm]{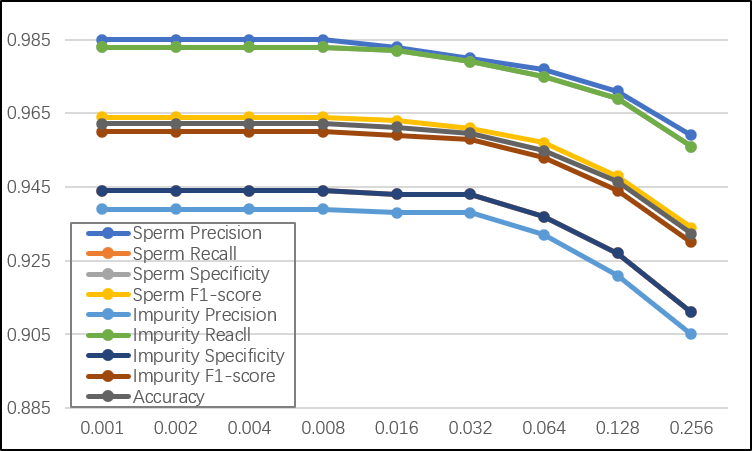}
\end{minipage}
}
\subfigure[DensenNet-121]{
\begin{minipage}[t]{0.23\linewidth}
\includegraphics[width=4cm]{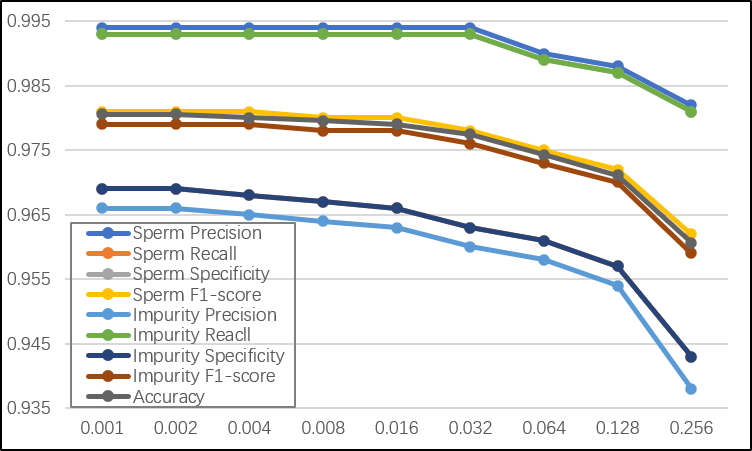}
\end{minipage}
}
\quad
\subfigure[Inception-V3]{
\begin{minipage}[t]{0.23\linewidth}
\includegraphics[width=4cm]{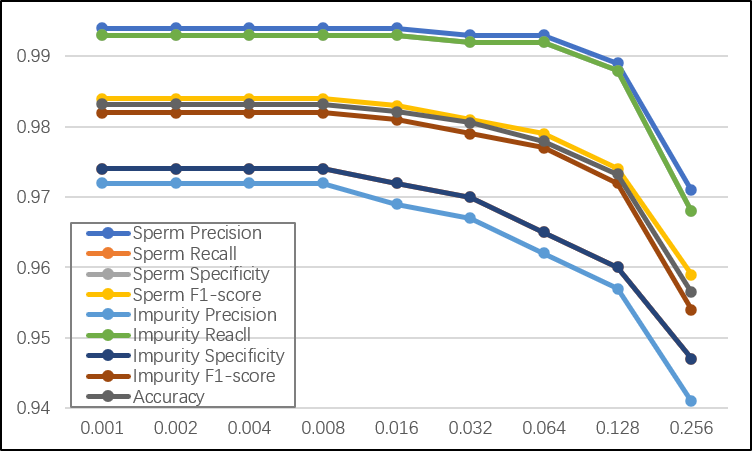}
\end{minipage}
}
\subfigure[MobileNet-V2]{
\begin{minipage}[t]{0.23\linewidth}
\includegraphics[width=4cm]{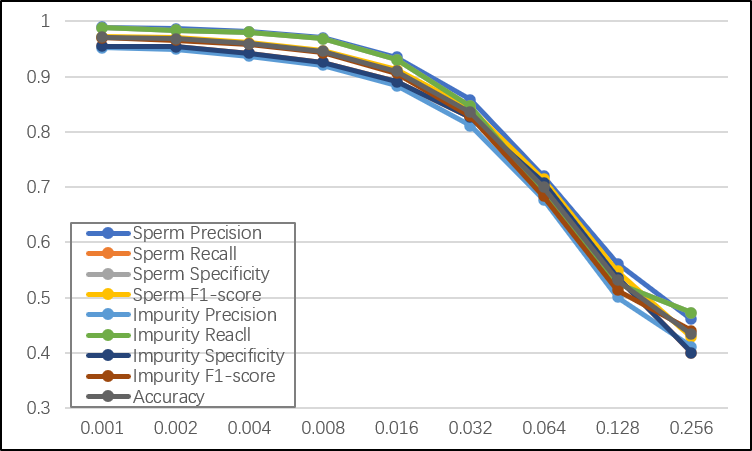}
\end{minipage}
}
\subfigure[ShuffleNet-V2]{
\begin{minipage}[t]{0.23\linewidth}
\includegraphics[width=4cm]{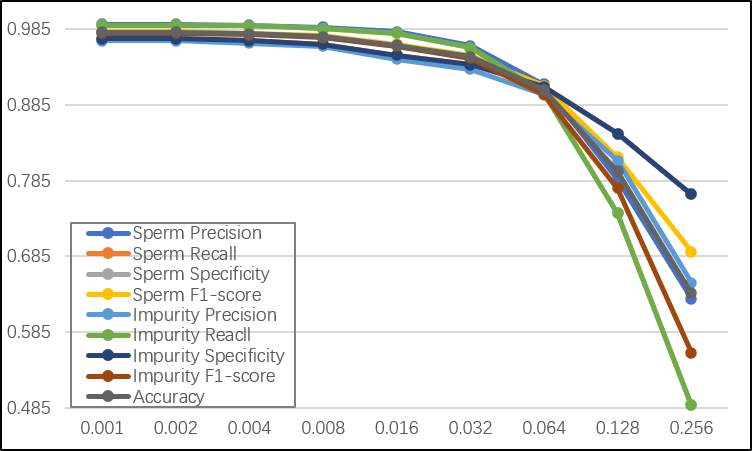}
\end{minipage}
}
\subfigure[Xception]{
\begin{minipage}[t]{0.23\linewidth}
\includegraphics[width=4cm]{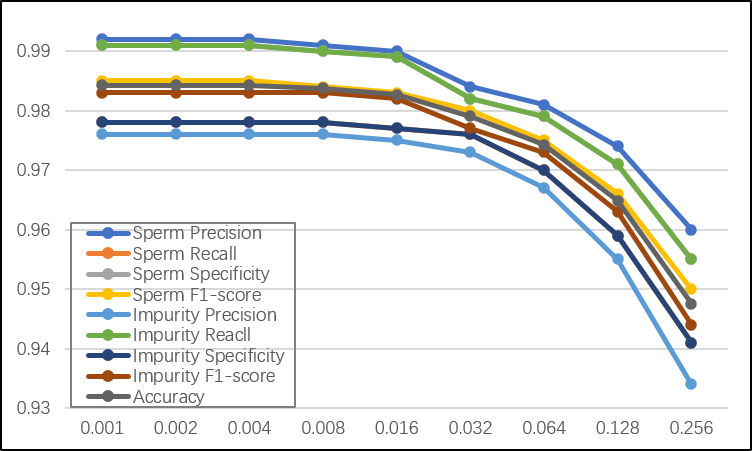}
\end{minipage}
}
\quad
\subfigure[ViT]{
\begin{minipage}[t]{0.23\linewidth}
\includegraphics[width=4cm]{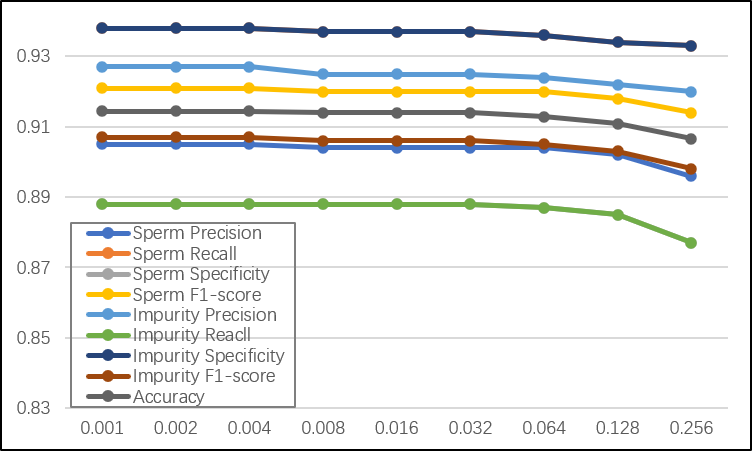}
\end{minipage}
}
\subfigure[BotNet]{
\begin{minipage}[t]{0.23\linewidth}
\includegraphics[width=4cm]{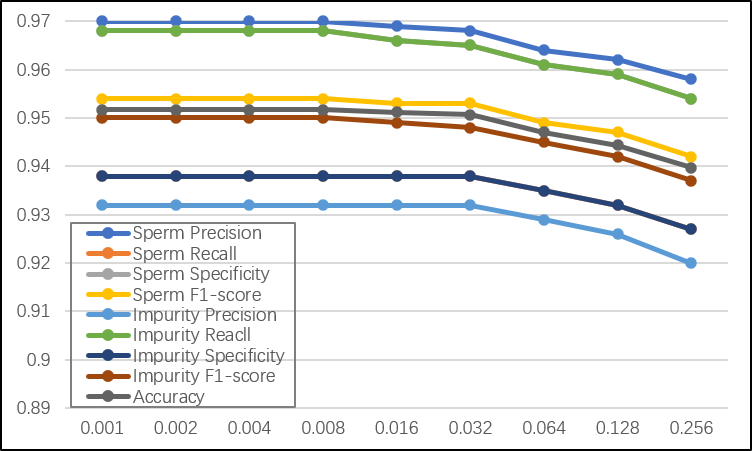}
\end{minipage}
}
\subfigure[DeiT-Base]{
\begin{minipage}[t]{0.23\linewidth}
\includegraphics[width=4cm]{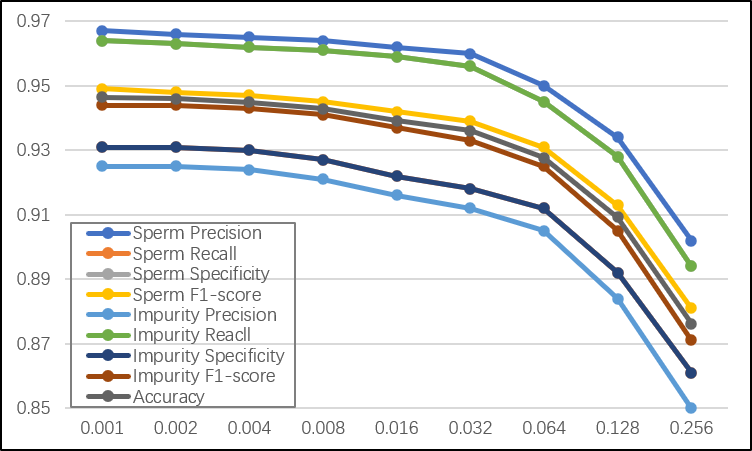}
\end{minipage}
}
\subfigure[DeiT-Tiny]{
\begin{minipage}[t]{0.23\linewidth}
\includegraphics[width=4cm]{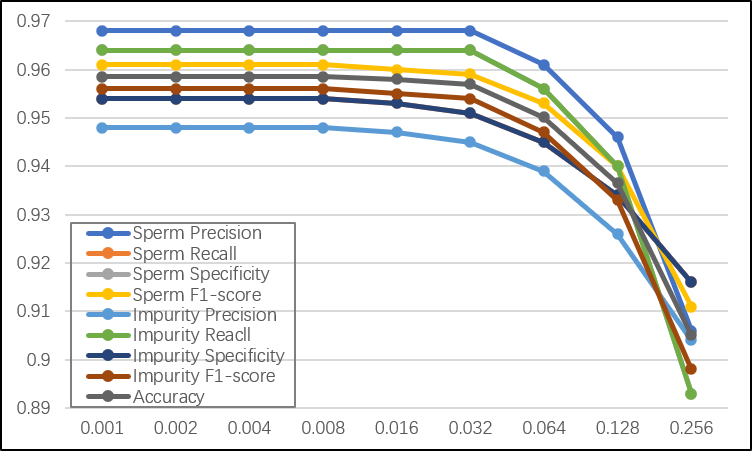}
\end{minipage}
}
\quad
\subfigure[T2T-ViT-t-19]{
\begin{minipage}[t]{0.23\linewidth}
\includegraphics[width=4cm]{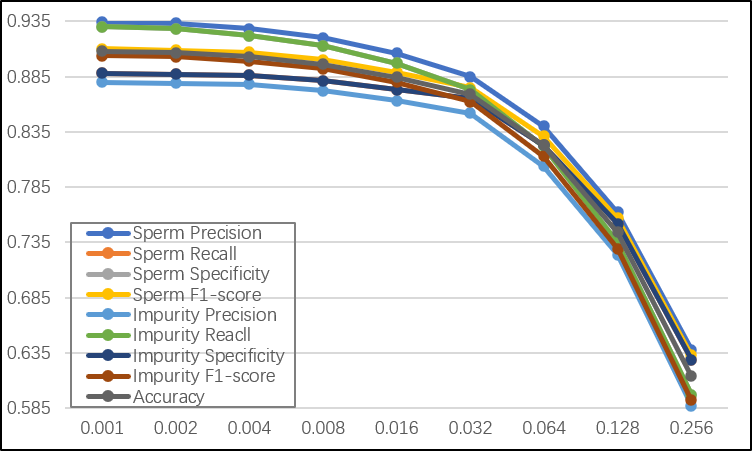}
\end{minipage}
}
\subfigure[T2T-ViT-t-24]{
\begin{minipage}[t]{0.23\linewidth}
\includegraphics[width=4cm]{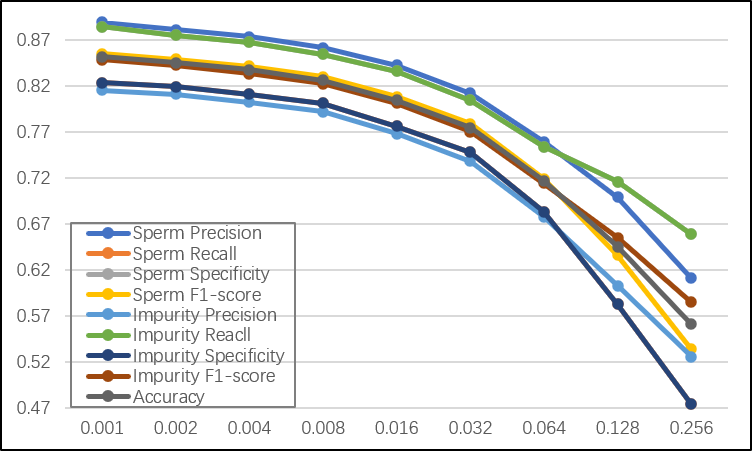}
\end{minipage}
}
\subfigure[T2T-ViT-7]{
\begin{minipage}[t]{0.23\linewidth}
\includegraphics[width=4cm]{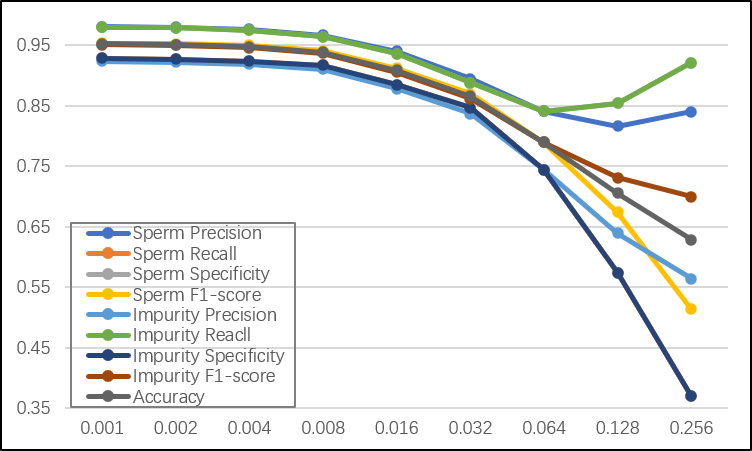}
\end{minipage}
}
\subfigure[T2T-ViT-10]{
\begin{minipage}[t]{0.23\linewidth}
\includegraphics[width=4cm]{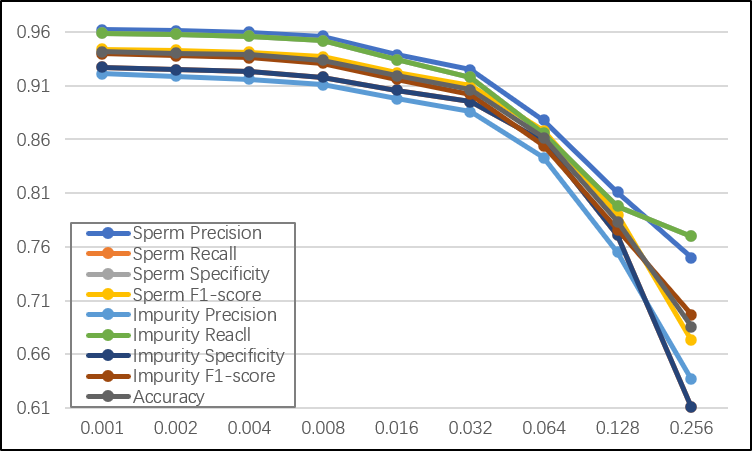}
\end{minipage}
}
\quad
\subfigure[T2T-ViT-12]{
\begin{minipage}[t]{0.23\linewidth}
\includegraphics[width=4cm]{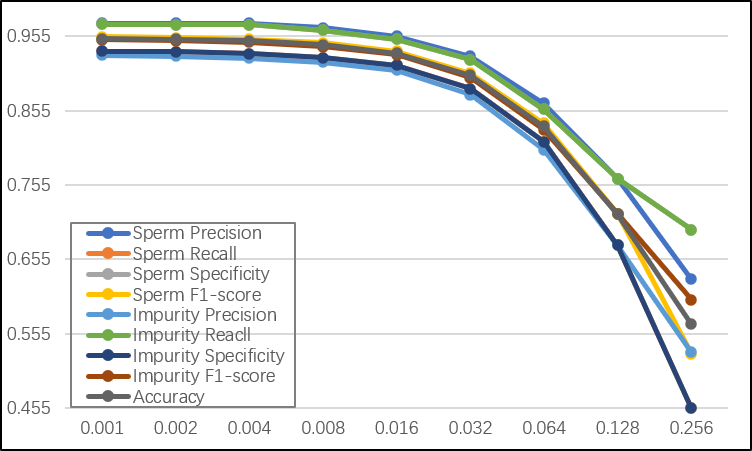}
\end{minipage}
}
\subfigure[T2T-ViT-14]{
\begin{minipage}[t]{0.23\linewidth}
\includegraphics[width=4cm]{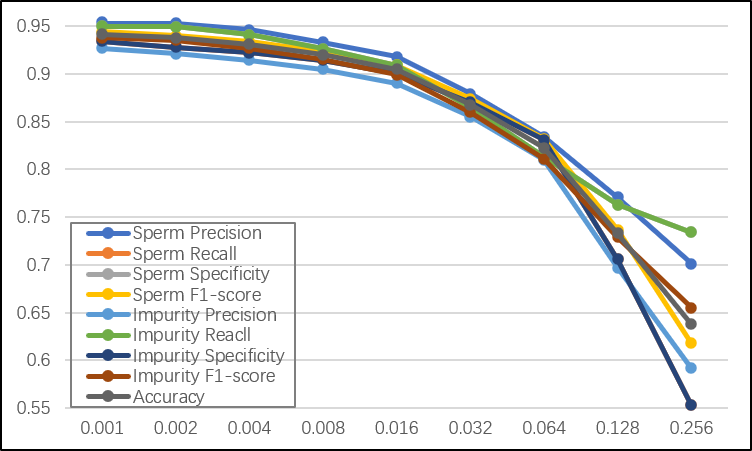}
\end{minipage}
}
\subfigure[T2T-ViT-19]{
\begin{minipage}[t]{0.23\linewidth}
\includegraphics[width=4cm]{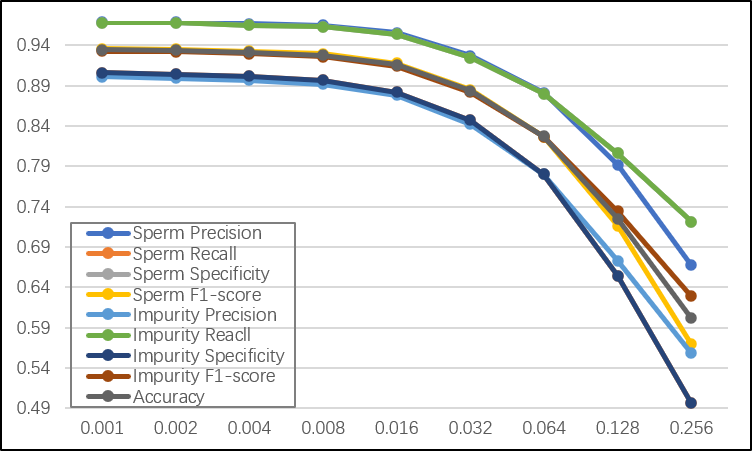}
\end{minipage}
}
\subfigure[T2T-ViT-24]{
\begin{minipage}[t]{0.23\linewidth}
\includegraphics[width=4cm]{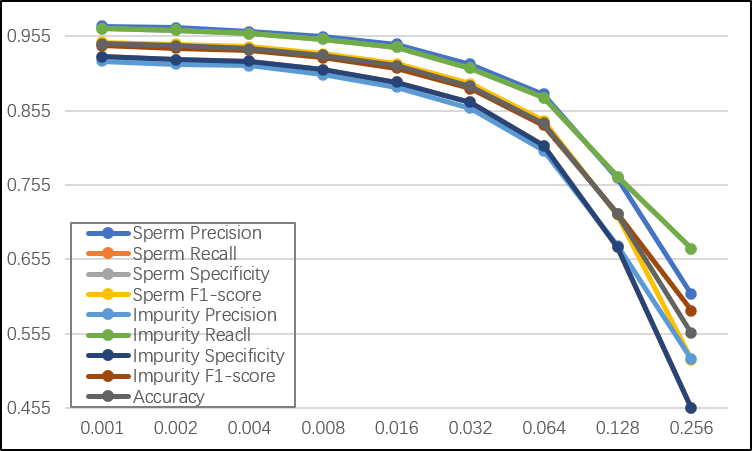}
\end{minipage}
}
\caption{Evaluation metrics curves of deep learning models under FGM. Here, (a) exhibits the evaluation  metrics curves of AlexNet model. Similarly, (b) presents the evaluation metrics curves of VGG-11 model, (c) is generated on VGG-16, (d) is on VGG-19, (e) is on ResNet-50,  (f) is on ResNet-101, (g) is on GoogleNet, (h) is on DensenNet-121, (i) is on Inception-V3, (j) is on MobileNet-V2, (k) is on ShuffleNet-V2, (l) is on Xception, (m) is on ViT, (n) is on BotNet, (o) is on DeiT-Base, (p) is on DeiT-Tiny, (q) is on T2T-ViT-t-19, (r) is on T2T-ViT-t-24, (s) is on T2T-ViT-7, (t) is on T2T-ViT-10, (u) is on T2T-ViT-12, (v) is on T2T-ViT-14, (w) is on T2T-ViT-19, (x) is on T2T-ViT-24.}
\label{FIG:5}
\end{figure*}

\paragraph{Evaluation of different deep learning models under salt and pepper noise}~{}

First, we compare the histogram (as shown in Fig.~\ref{FIG:8}) of the original test set with the histogram (as shown in Fig.~\ref{FIG:11} (a)) of salt and pepper noise with the mean value of 0.1, which can be seen from the two figures: from the columnar trend of evaluation metrics in the overall histogram, the change of DenseNet-121, ViT, BotNet and T2T-ViT-t-19 are relatively small. When we focus on these four models, we can find that under DenseNet-121, the change of sperm recall, sperm accuracy and impurity specificity are about 14 \%, the change of sperm F1-score, impurity F1-score and accuracy are about 10 \%, and the change of other evaluation metrics are about 6 \%. Under ViT, sperm recall, impurity precision and impurity specificity change about 5 \%, sperm F1-score, impurity F1-score and accuracy change about 2.5 \%, and other evaluation metrics change little. Under BotNet, the change of sperm specificity and impurity recall are about 8 \%, sperm F1-score, impurity F1-score and accuracy are about 5 \%, sperm precision is about 7 \%, and other evaluation metrics are about 2 \%. Under T2T-ViT-t-19, the sperm recall, impurity precision and impurity specificity change about 9 \%, the sperm F1-score, impurity F1-score and accuracy change about 6 \%, and the other evaluation metrics change about 4 \%.\par
Then, we compare the histogram (as shown in Fig.~\ref{FIG:8}) of the original test set with the histogram (as shown in Fig.~\ref{FIG:11} (b)) of salt and pepper noise with the mean value of 0.05, which can be seen from the two figures: from the columnar trend of evaluation metrics in the overall histogram, the change of DenseNet-121, ViT, T2T-ViT-t-19 and T2T-ViT-7/10/12/14 are relatively small. When we focus on these seven models, we can find that under DenseNet-121, the change of sperm specificity and impurity recall are about 6 \%, sperm precision, sperm F1-score, impurity F1-score and accuracy are about 5 \%, and the other evaluation metrics are about 3 \%. Under ViT, the sperm recall, impurity precision and impurity specificity change about 3.6 \%, the sperm precision, impurity F1-score and accuracy change about 1 \%, and the other evaluation metrics change about 1.5 \%. Under T2T-ViT-t-19, sperm specificity and impurity recall change about 9 \%, sperm precision change about 7 \%, sperm F1-score, impurity F1-score and accuracy change about 4.5 \%, and the other evaluation metrics change little.\par 
Meanwhile, under T2T-ViT-7, sperm precision, sperm specificity and impurity recall change about 4 \%, sperm F1-score, impurity F1-score and accuracy change about 3 \%, and the remaining evaluation metrics change about 2 \%. At same time, the change of each evaluation metric in T2T-ViT-14 are larger than those in T2T-ViT-7, specifically sperm precision, sperm specificity and impurity recall change about 4.5 \%, sperm F1-score, impurity F1-score and accuracy change about 3.5 \%, and the remaining evaluation metrics change about 2.5 \%. Under T2T-ViT-12, sperm precision, sperm specificity and impurity recall change about 5\%, sperm F1-score, impurity F1-score and accuracy change about 3 \%, and the remaining evaluation metrics change little. The change of each evaluation metric in T2T-ViT-10 are larger than those in T2T-ViT-12, specifically sperm precision, sperm specificity and impurity recall change about 6.5\%, sperm F1-score, impurity F1-score and accuracy change about 4 \%, and the remaining evaluation metrics change little. \par
Therefore,under the influence of salt and pepper noise with a mean value of 0.1 or 0.05, in the CNN and VT models used in this paper, ViT, T2T-ViT-t-19 and T2T-ViT-7/12 have strong anti-noise robustness for the classification of tiny object (sperm and impurity) image dataset.

\paragraph{Evaluation of different deep learning models under periodic noise}~{}

First, we compare the histogram (as shown in Fig.~\ref{FIG:8}) of the original test set with the histogram (as shown in Fig.~\ref{FIG:12} (a)) of periodic noise with the sin amplitude and angle are 50, which can be seen from the two figures: under the influence of this periodic noise, the evaluation metrics in each model change greatly. Compared with the performance of all models, the change of VGG-19, ResNet-101 and ViT are slightly flat. However, when we focus on these three models, we can find that under VGG-19, sperm precision, sperm specificity and impurity recall change about 20 \%, sperm F1-score, impurity F1-score and accuracy change about 10 \%, and the other evaluation metrics change little. Under ResNet-101, the change of sperm precision is about 15 \%, the change of sperm specificity and impurity recall are about 20 \%, the change of sperm F1-score, impurity F1-score and accuracy are about 10 \%, and the other evaluation metrics change about 4.5 \%. Under the ViT, the change of sperm precision is about 10 \%, the change of sperm specificity and impurity recall are about 17 \%, the impurity F1-score is about 8 \%, and the other evaluation metrics change about 5 \%.\par

Then, we compare the histogram (as shown in Fig.~\ref{FIG:8}) of the original test set with the histogram (as shown in Fig.~\ref{FIG:12} (b)) of periodic noise with the sin amplitude is 50 and angle is 40, which can be seen from the two figures: under the influence of this periodic noise, we can find that each evaluation metric of all models change dramatically, and ResNet-101 has the smallest change. However, when we focus on ResNet-101, we can find that under ResNet-101 model, sperm recall and impurity specificity change about 1 \%, impurity precision change about 3.5 \%, and the remaining evaluation metrics are more than 14.5 \%\par
Therefore, under the influence of periodic noise with sin amplitude value of 50 and angle value of 50 or 40, in the CNN and VT models used in this paper, all models in this paper have poor anti-noise robustness for the classification of tiny object (sperm and impurity) image dataset.

\paragraph{Evaluation of different deep learning models under uniform, rayleigh or exponential noise}~{} 

We compare the histogram (as shown in Fig.~\ref{FIG:8}) of the original test set with the histogram (as shown in Fig.~\ref{FIG:13} (a)) of uniform noise with a mean value of 0 and a variance of 1,  rayleigh noise with a mean value of 0 (as shown in Fig.~\ref{FIG:14} (a)) and a variance of 1 and exponential noise with a mean value of 1 (as shown in Fig.~\ref{FIG:15} (a)), respectively. It is obvious that each evaluation metric change dramatically in all models, so under the influence of uniform noise with a mean value of 0 and a variance of 1 , rayleigh noise with a mean value of 0 and a variance of 1 or exponential noise with a mean value of 1 tested in CNN and VT models, all models in this paper have poor anti-noise robustness for the classification of tiny object (sperm and impurity) image dataset.

\paragraph{Evaluation of different deep learning models under poission noise}~{} 

We compare the histogram (as shown in Fig.~\ref{FIG:8}) of the original test set with the histogram (as shown in Fig.~\ref{FIG:16} (a)) of poission noise with default values, which can be seen from the two figures: from the columnar trend of evaluation metrics in the overall histogram, we can find that except T2T-ViT-24 and T2T-ViT-7 / 10 / 12 / 14 / 19 / 24, the remaining model change are relatively small. When we focus on the better performance of VGG-19, DenseNet-121, Inception-V3, ShflleNet-V2, Xception, ViT and BotNet, we can find that under VGG-19, sperm recall, impurity precision and impurity specificity change about 1.6 \%, sperm F1-score, impurity F1-score and accuracy change about 0.7 \%, other evaluation metrics change little. Under DenseNet-121, sperm precision, sperm recall and impurity specificity change about 1.5 \%, other evaluation metrics change about 1.2 \%. Under Inception-V3, all evaluation metrics change about 1.2 \%. Under ShflleNet-V2, sperm recall, impurity precision and impurity specificity change about 1.3 \%, impurity F1-score changes about 1.2 \%, other evaluation metrics change about 1 \%. Under Xception, sperm recall, impurity precision and impurity specificity change about 1.2 \%, other evaluation metrics change about 1 \%. Under ViT, sperm recall, impurity precision and impurity specificity change about 1.3 \%, sperm specificity and impurity recall change 0.7 \%, other evaluation metrics change about 0.3 \%. Under BotNet, sperm precision changes 3.8 \%, sperm specificity and impurity recall change 4.7 \%, sperm recall, impurity precision and impurity specificity change 2.3 \%, sperm F1-score, impurity F1-score and accuracy change about 1 \%.\par
Therefore, under the influence of posission noise with default values tested in the CNN and VT models, VGG-19, DenseNet-121, Inception-V3, ShflleNet-V2, Xception, ViT and BotNet have strong anti-noise robustness for the classification of tiny object (sperm and impurity) image dataset.

\subsubsection{Classification Performance of Adversarial Attacks on Test Set with Different Deep Learning Models}
In this paper, we apply adversarial attacks to the test training of different deep learning models. In this section, we obtain the curves and histograms of each evaluation metric under FGM, FGSM and I-FGSM. Because of the application of DeepFool, each evaluation metric change to 0, so the evaluation metric curves and histograms of DeepFool are not shown. Meanwhile,we can known that under DeepFool adversarial attack, as $\epsilon$ increases in multiplication from 0.01 to 0.256, in the CNN and VT models used in this paper, all models have poor anti-noise robustness for the classification of tiny object (sperm and impurity) image dataset.

\paragraph{Evaluation of different deep learning models under FGM}~{} 

Fig.~\ref{FIG:5} shows the variation of each evaluation metric in CNN and VT models with the increase of $\epsilon$ under the FGM adversarial attack. First of all, we observe the evaluation metrics curves of each model in Fig.~\ref{FIG:5}, and it can be known that the change trend of evaluation metrics of AlexNet, VGG-11, VGG-16, VGG-19, ResNet-50, ResNet-101, GoogleNet, DensenNet-121, Inception-V3, Xception, ViT, BotNet, DeiT-Base and DeiT-Tiny are relatively smooth.\par

Then, when we focus on these models, it can be clearly seen from these histograms (as shown in Fig.~\ref{FIG:17} (a) \~{} (i)) in Appendix: when the parameter $\epsilon$ increases from 0.01 to 0.256 by multiplication, we can find that under AlexNet, sperm recall, impurity precision and impurity specificity cha\-nge about 4 \%, sperm F1-score, impurity F1-score and accuracy change about 3 \%, and other evaluation metrics change 1.7 \%. Under VGG-11, all evaluation metrics change about 1 \%. Under VGG-16, sperm precision, sperm specificity and impurity recall changes 0.7 \%, 0.8 \% and 0.8 \%, respectively, sperm recall, impurity precision and impurity specificity change range are 0.5 \%, other evaluation metrics change 0.6 \%. Under VGG-19, sperm precision, sperm specificity and impurity recall change 0.4 \%, sperm F1-score, impurity F1-score and accuracy change 0.9 \%, other evaluation metrics change about 1.3 \%. Under ResNet-50, sperm precision, sperm specificity and impurity recall change about 2 \%, sperm F1-score, impurity F1-score and accuracy change about 2.5 \%, other evaluation metrics change about 3 \%. Under ResNet-101, sperm precision, sperm specificity and impurity recall change 1.5 \%, sperm F1-score, impurity F1-score and accuracy change about 2.5 \%, other evaluation metrics change about 3.5 \%. Under GoogleNet, all evaluation metrics change about 3 \%. Under DensenNet-121, sperm precision, sperm specificity and impurity recall change 1.2 \%, sperm F1-score, impurity F1-score and accuracy change about 2 \%, other evaluation metrics change about 2.6 \%. Under Inception-V3, impurity precision and impurity F1-score change about 3 \%, other evaluation metrics change about 2.5 \%. Under Xception, sperm precision change 3.2 \%, impurity precision and impurity F1-score changes 4.2 \% and 3.9 \%, respectively, other evaluation metrics change about 3.5 \%.\par 

Meanwhile, under ViT, sperm recall and impurity specificity change 0.5 \%, sperm F1-score and impurity precision change 0.7 \%, other evaluation metrics change about 1 \%. Under BotNet, sperm specificity, impurity recall and impurity F1-score change about 1 \%, other evaluation metrics change about 1.1 \%. Under DeiT-Base, sperm precision, sperm specificity and impurity recall changes 6.6 \%, 7 \% and 7 \%,respectively, sperm F1-score, impurity F1-score and accuracy changes 8 \%, 8.5 \% and 8.24 \%,respectively, sperm recall impurity precision and impurity specificity changes 9.3 \%, 9.8 \% and 9.3 \%,respectively. Under DeiT-Tiny, sperm recall and impurity specificity change 1.5 \%, impurity precision change 2.1 \%, other evaluation metrics change about 5 \%.\par

Therefore, it can be known that under FGM adversarial attack, as $\epsilon$ increases in multiplication from 0.01 to 0.256, in the CNN and VT models used in this paper, VGG-11/16/19, ViT and BotNet have strong anti-noise robustness for the classification of tiny object (sperm and impurity) image dataset.

\begin{figure*}
\centering
\subfigure[AlexNet]{
\begin{minipage}[t]{0.23\linewidth}
\includegraphics[width=4cm]{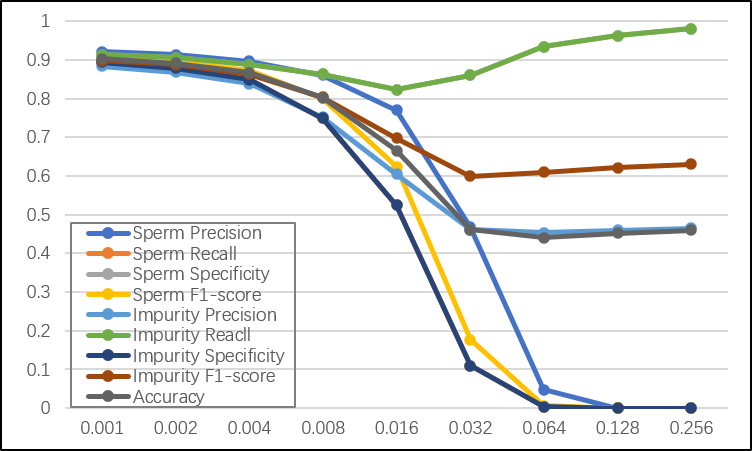}
\end{minipage}
}
\subfigure[VGG-11]{
\begin{minipage}[t]{0.23\linewidth}
\includegraphics[width=4cm]{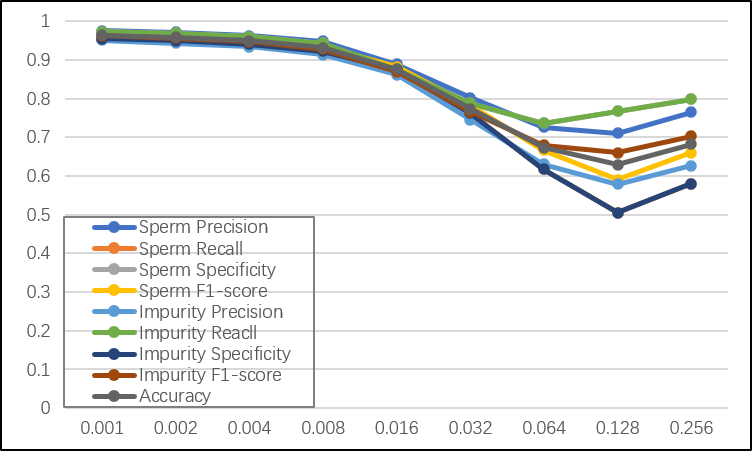}
\end{minipage}
}
\subfigure[VGG-16]{
\begin{minipage}[t]{0.23\linewidth}
\includegraphics[width=4cm]{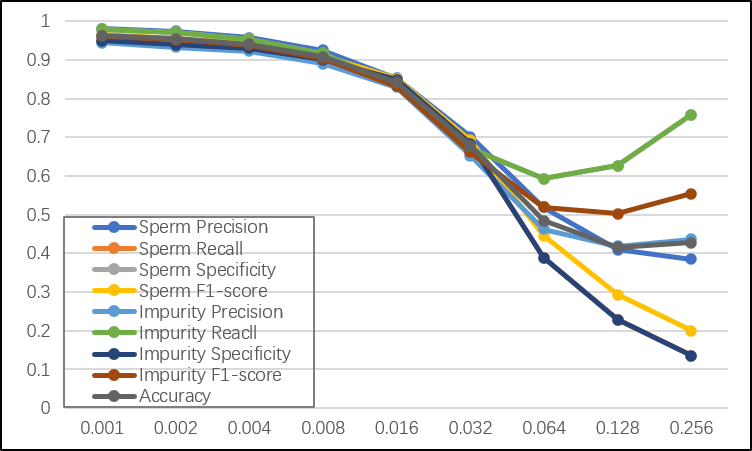}
\end{minipage}
}
\subfigure[VGG-19]{
\begin{minipage}[t]{0.23\linewidth}
\includegraphics[width=4cm]{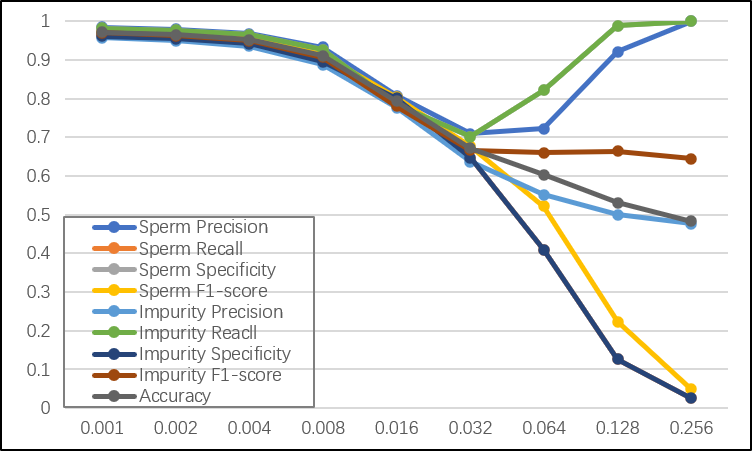}
\end{minipage}
}
\quad
\subfigure[ResNet-50]{
\begin{minipage}[t]{0.23\linewidth}
\includegraphics[width=4cm]{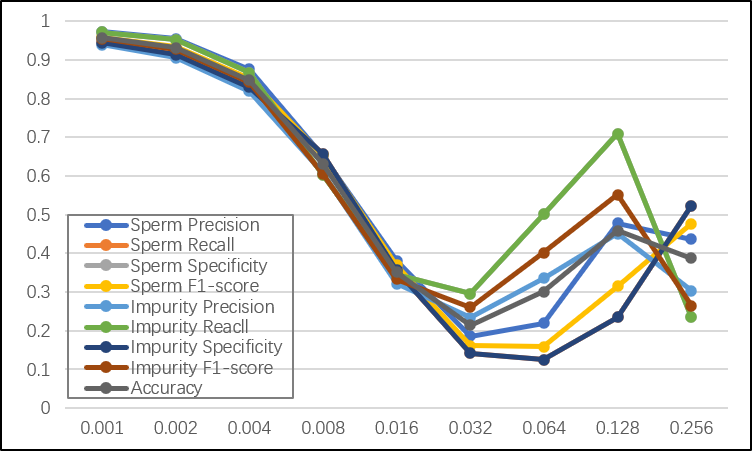}
\end{minipage}
}
\subfigure[ResNet-101]{
\begin{minipage}[t]{0.23\linewidth}
\includegraphics[width=4cm]{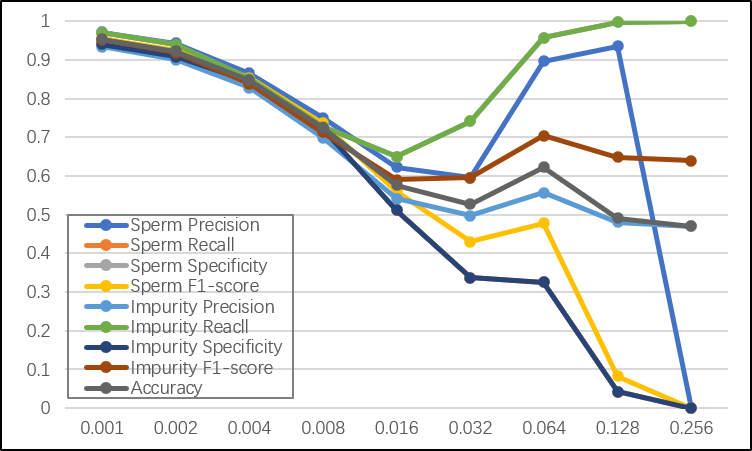}
\end{minipage}
}
\subfigure[GoogleNet]{
\begin{minipage}[t]{0.23\linewidth}
\includegraphics[width=4cm]{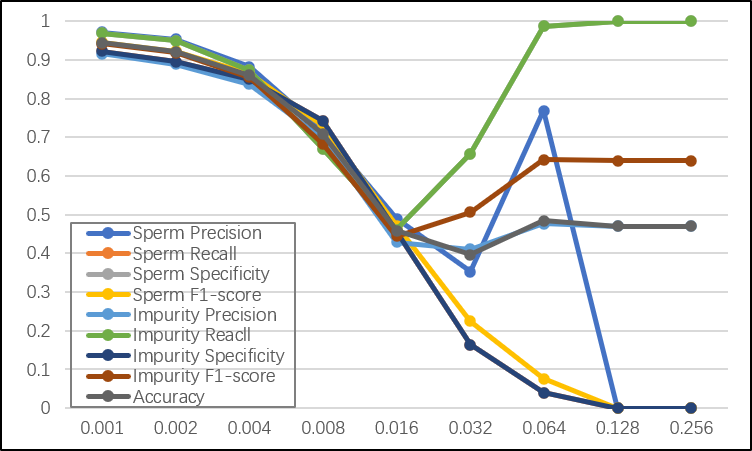}
\end{minipage}
}
\subfigure[DensenNet-121]{
\begin{minipage}[t]{0.23\linewidth}
\includegraphics[width=4cm]{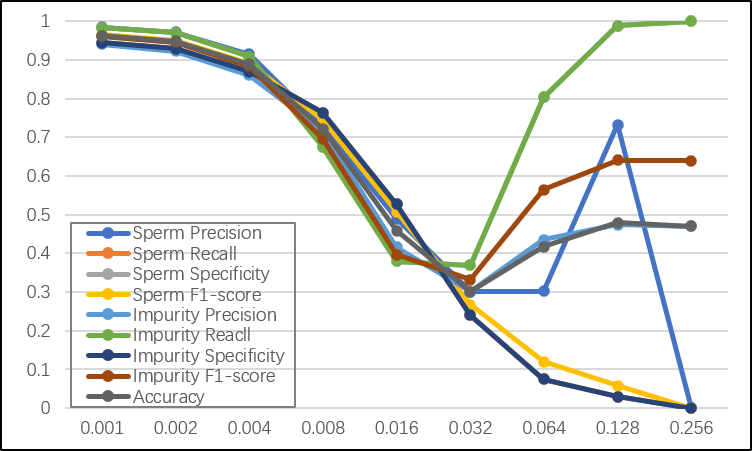}
\end{minipage}
}
\quad
\subfigure[Inception-V3]{
\begin{minipage}[t]{0.23\linewidth}
\includegraphics[width=4cm]{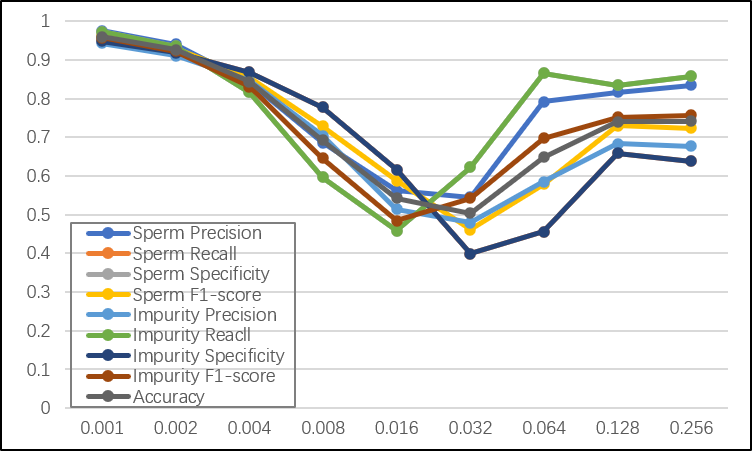}
\end{minipage}
}
\subfigure[MobileNet-V2]{
\begin{minipage}[t]{0.23\linewidth}
\includegraphics[width=4cm]{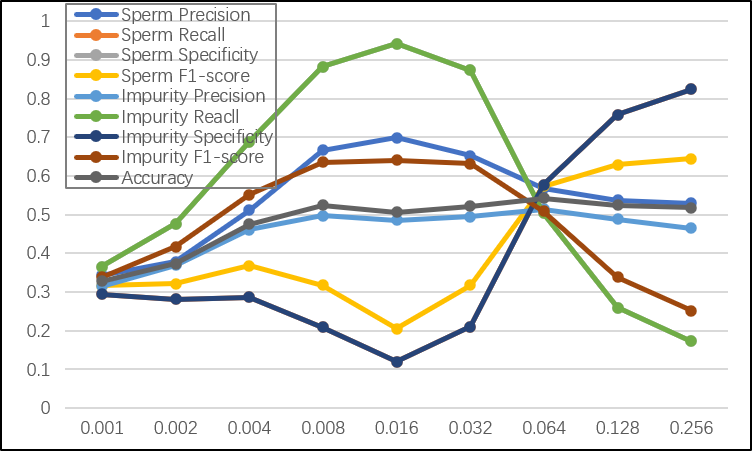}
\end{minipage}
}
\subfigure[ShuffleNet-V2]{
\begin{minipage}[t]{0.23\linewidth}
\includegraphics[width=4cm]{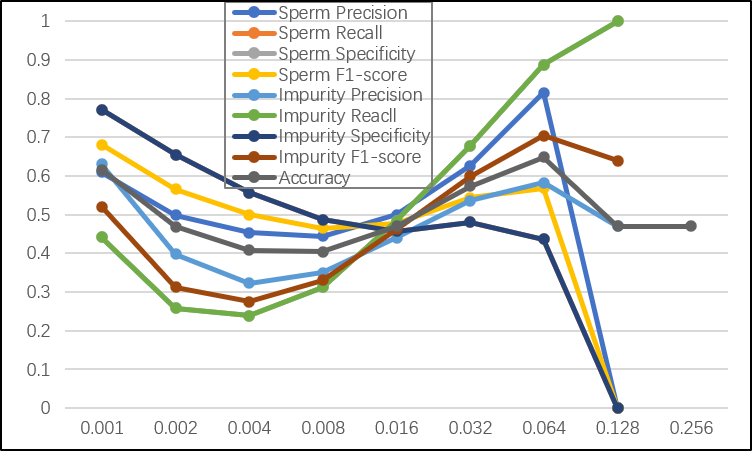}
\end{minipage}
}
\subfigure[Xception]{
\begin{minipage}[t]{0.23\linewidth}
\includegraphics[width=4cm]{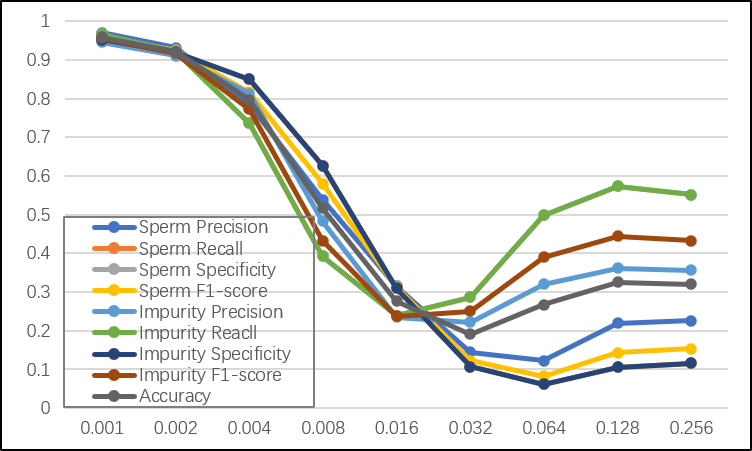}
\end{minipage}
}
\quad
\subfigure[ViT]{
\begin{minipage}[t]{0.23\linewidth}
\includegraphics[width=4cm]{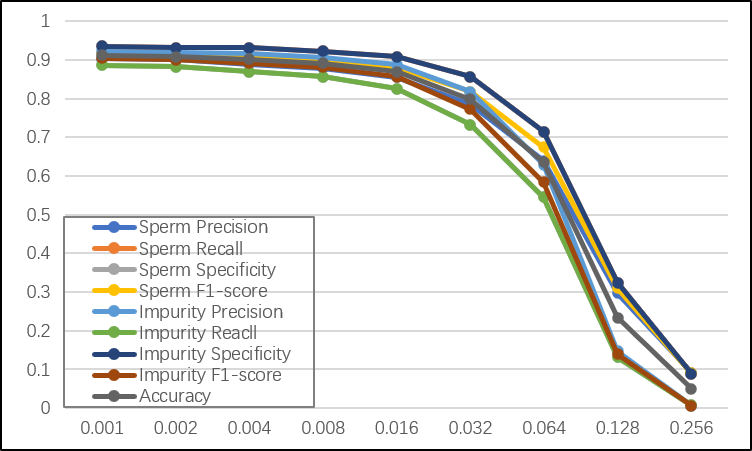}
\end{minipage}
}
\subfigure[BotNet]{
\begin{minipage}[t]{0.23\linewidth}
\includegraphics[width=4cm]{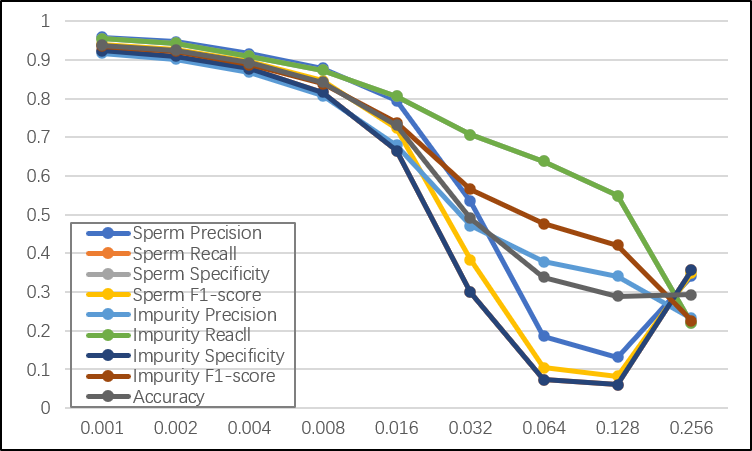}
\end{minipage}
}
\subfigure[DeiT-Base]{
\begin{minipage}[t]{0.23\linewidth}
\includegraphics[width=4cm]{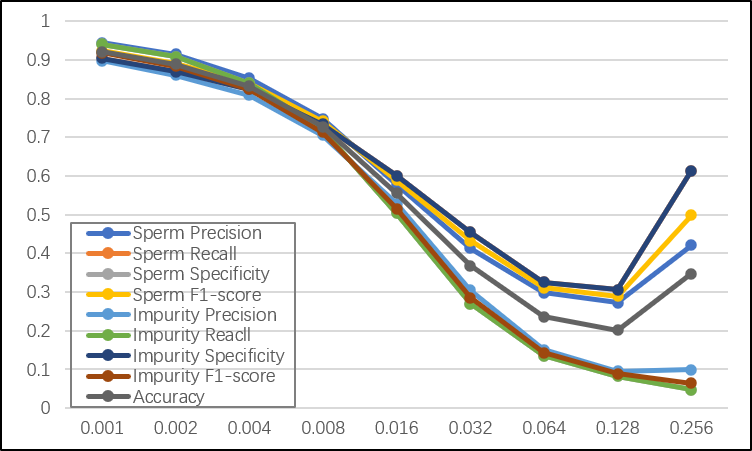}
\end{minipage}
}
\subfigure[DeiT-Tiny]{
\begin{minipage}[t]{0.23\linewidth}
\includegraphics[width=4cm]{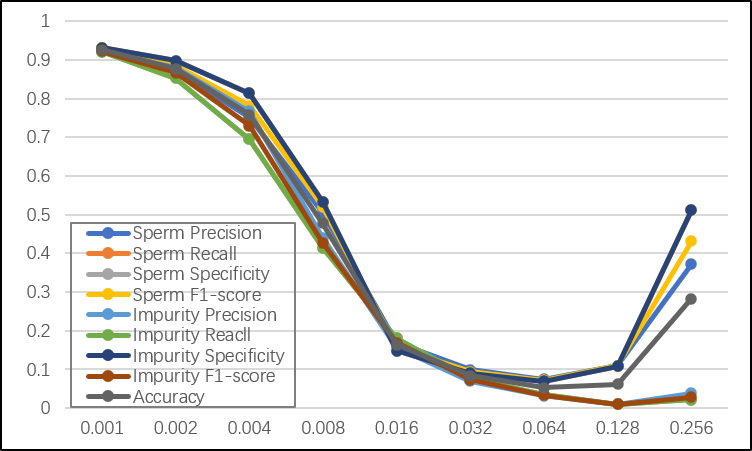}
\end{minipage}
}
\quad
\subfigure[T2T-ViT-t-19]{
\begin{minipage}[t]{0.23\linewidth}
\includegraphics[width=4cm]{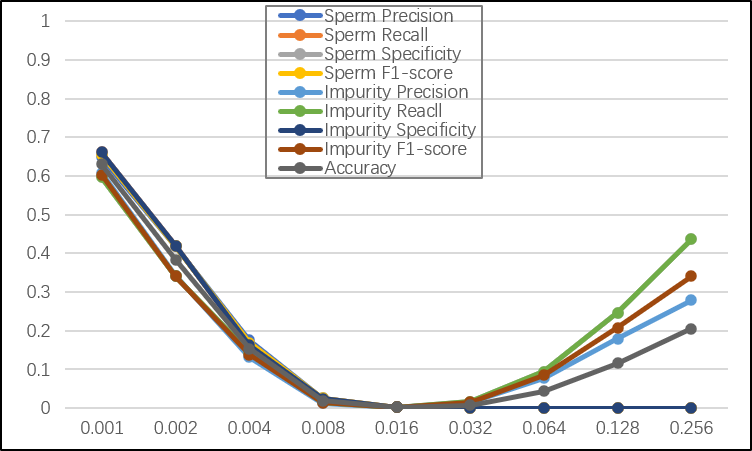}
\end{minipage}
}
\subfigure[T2T-ViT-t-24]{
\begin{minipage}[t]{0.23\linewidth}
\includegraphics[width=4cm]{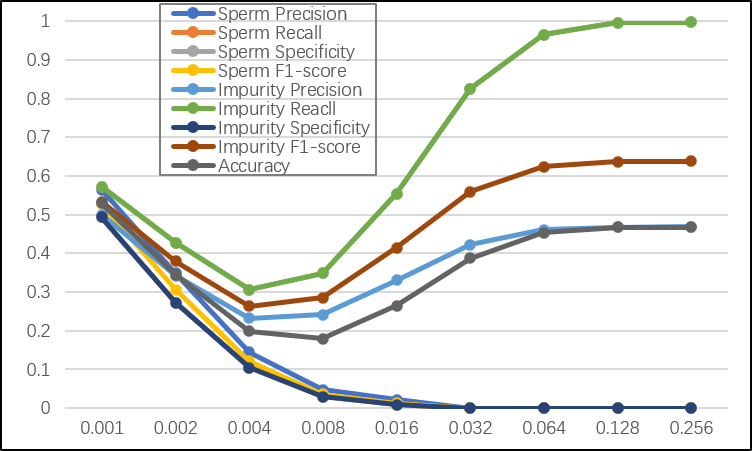}
\end{minipage}
}
\subfigure[T2T-ViT-7]{
\begin{minipage}[t]{0.23\linewidth}
\includegraphics[width=4cm]{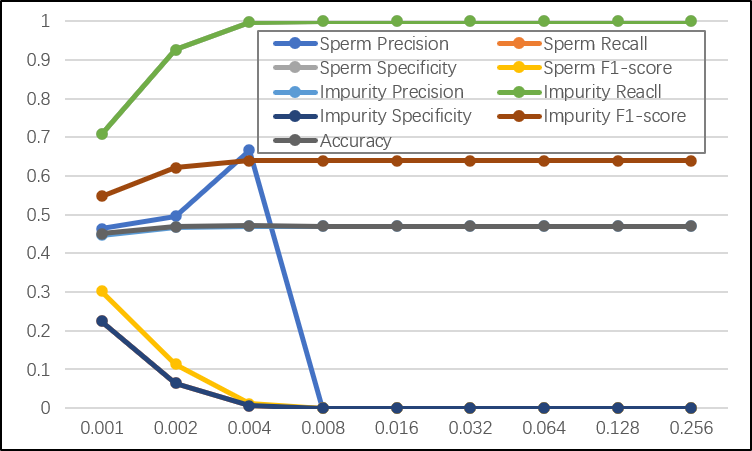}
\end{minipage}
}
\subfigure[T2T-ViT-10]{
\begin{minipage}[t]{0.23\linewidth}
\includegraphics[width=4cm]{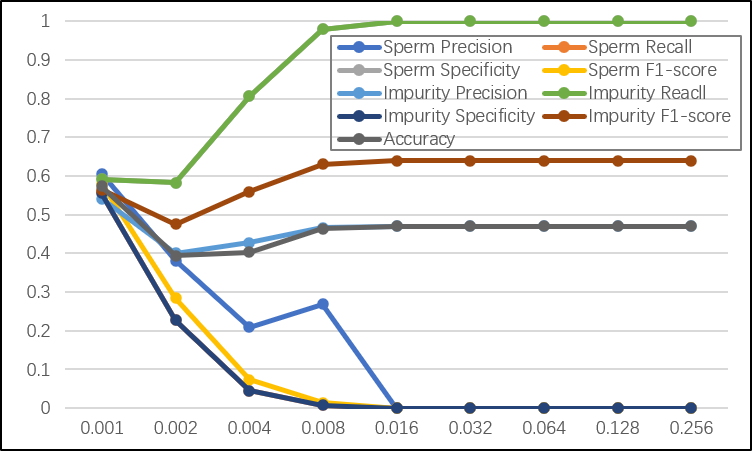}
\end{minipage}
}
\quad
\subfigure[T2T-ViT-12]{
\begin{minipage}[t]{0.23\linewidth}
\includegraphics[width=4cm]{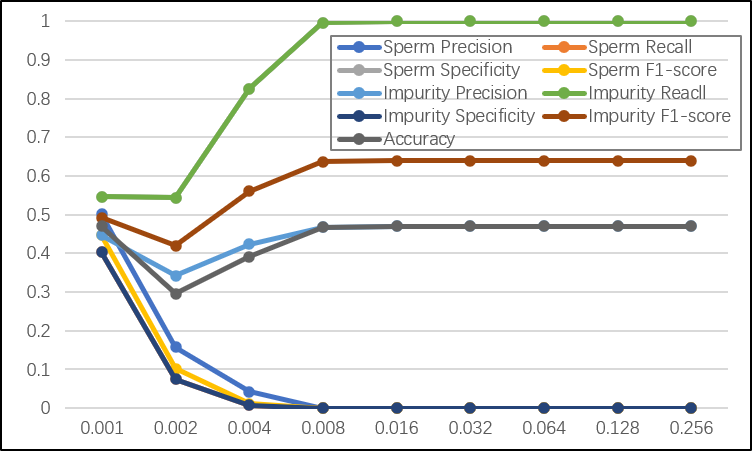}
\end{minipage}
}
\subfigure[T2T-ViT-14]{
\begin{minipage}[t]{0.23\linewidth}
\includegraphics[width=4cm]{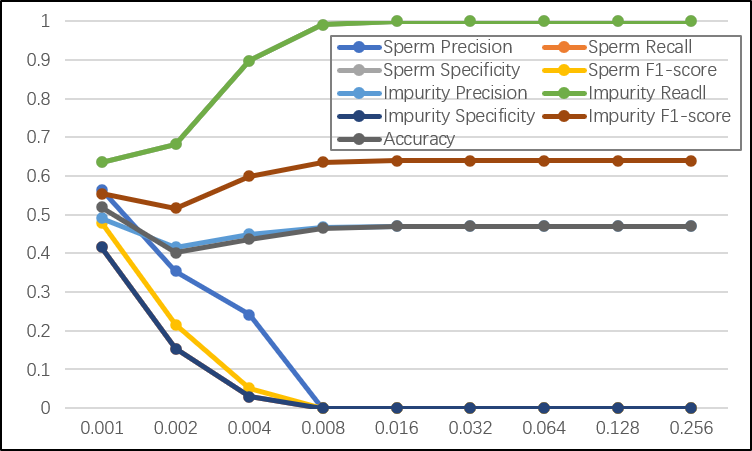}
\end{minipage}
}
\subfigure[T2T-ViT-19]{
\begin{minipage}[t]{0.23\linewidth}
\includegraphics[width=4cm]{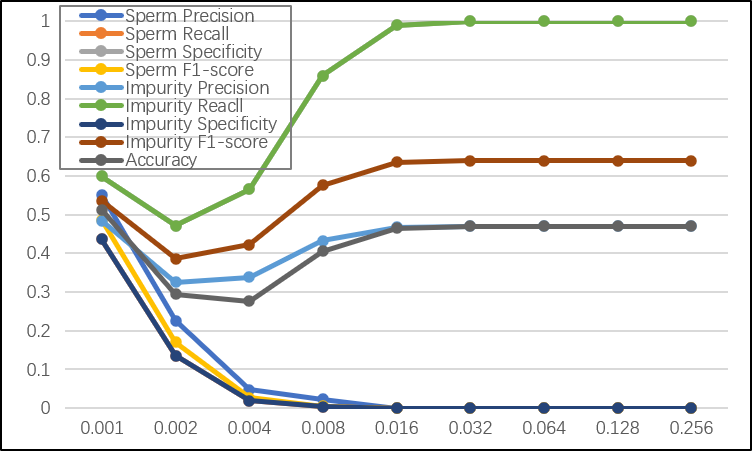}
\end{minipage}
}
\subfigure[T2T-ViT-24]{
\begin{minipage}[t]{0.23\linewidth}
\includegraphics[width=4cm]{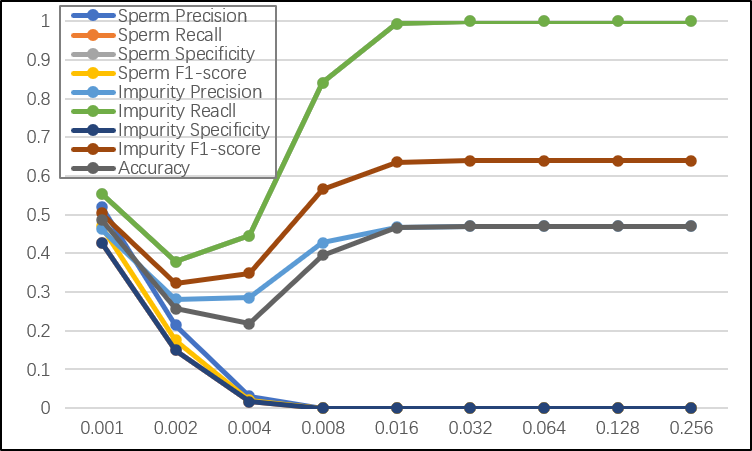}
\end{minipage}
}
\caption{Evaluation metrics curves of deep learning models under FGSM. Here, (a) exhibits the evaluation  metrics curves of AlexNet model. Similarly, (b) presents the evaluation metrics curves of VGG-11 model, (c) is generated on VGG-16, (d) is on VGG-19, (e) is on ResNet-50,  (f) is on ResNet-101, (g) is on GoogleNet, (h) is on DensenNet-121, (i) is on Inception-V3, (j) is on MobileNet-V2, (k) is on ShuffleNet-V2, (l) is on Xception, (m) is on ViT, (n) is on BotNet, (o) is on DeiT-Base, (p) is on DeiT-Tiny, (q) is on T2T-ViT-t-19, (r) is on T2T-ViT-t-24, (s) is on T2T-ViT-7, (t) is on T2T-ViT-10, (u) is on T2T-ViT-12, (v) is on T2T-ViT-14, (w) is on T2T-ViT-19, (x) is on T2T-ViT-24.}
\label{FIG:6}
\end{figure*}

\paragraph{Evaluation of different deep learning models under FGSM}~{} 

Fig.~\ref{FIG:6} shows the variation of each evaluation metric in CNN and VT models with the increase of $\epsilon$ under the FGSM adversarial attack. We observe the evaluation metrics curves of each model in Fig.~\ref{FIG:6}, and it can be known that the change trend of evaluation metrics of all models are large fluctuation. Therefore, we can known that under FGSM adversarial attack, as $\epsilon$ increases in multiplication from 0.01 to 0.256, in the CNN and VT models used in this paper, all models have poor anti-noise robustness for the classification of tiny object (sperm and impurity) image dataset.

\paragraph{Evaluation of different deep learning models under I-FGSM}~{} 

\begin{figure*}
\centering
\subfigure[AlexNet]{
\begin{minipage}[t]{0.23\linewidth}
\includegraphics[width=4cm]{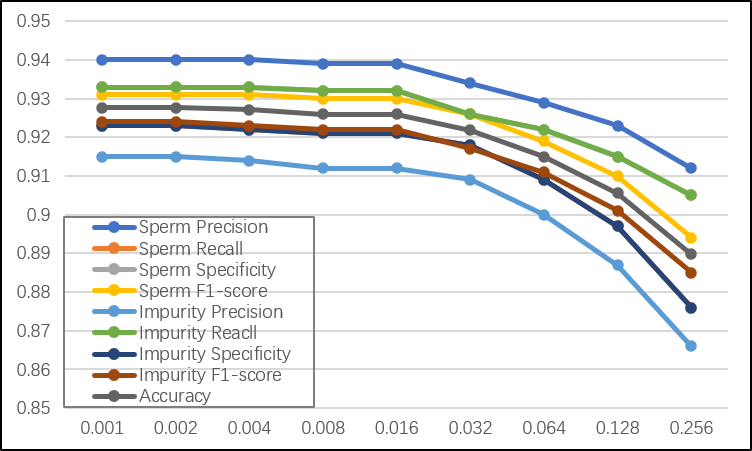}
\end{minipage}
}
\subfigure[VGG-11]{
\begin{minipage}[t]{0.23\linewidth}
\includegraphics[width=4cm]{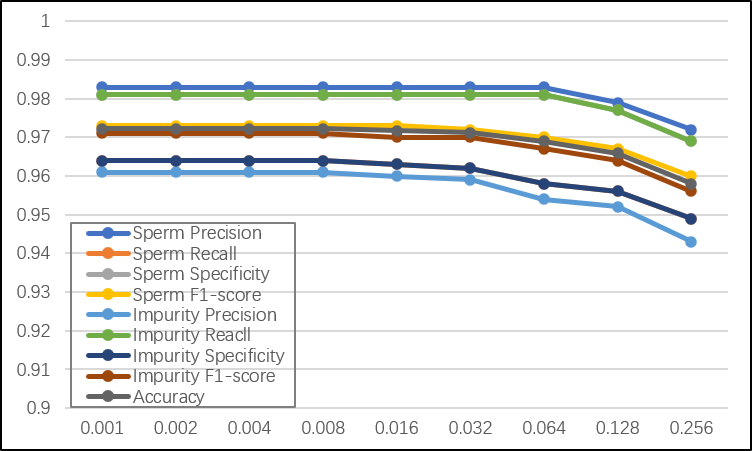}
\end{minipage}
}
\subfigure[VGG-16]{
\begin{minipage}[t]{0.23\linewidth}
\includegraphics[width=4cm]{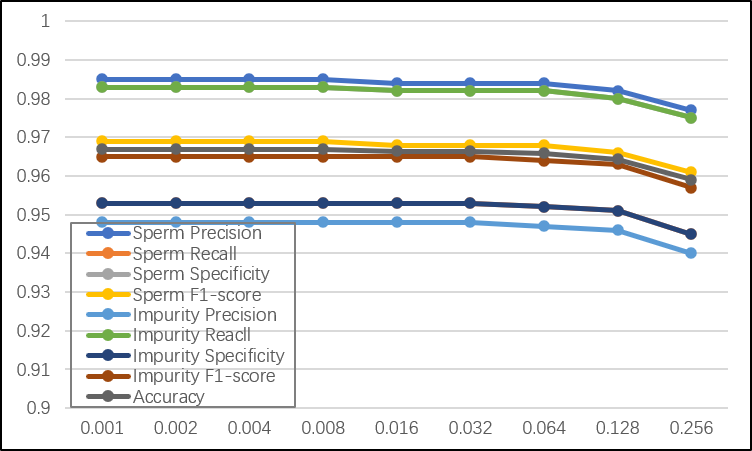}
\end{minipage}
}
\subfigure[VGG-19]{
\begin{minipage}[t]{0.23\linewidth}
\includegraphics[width=4cm]{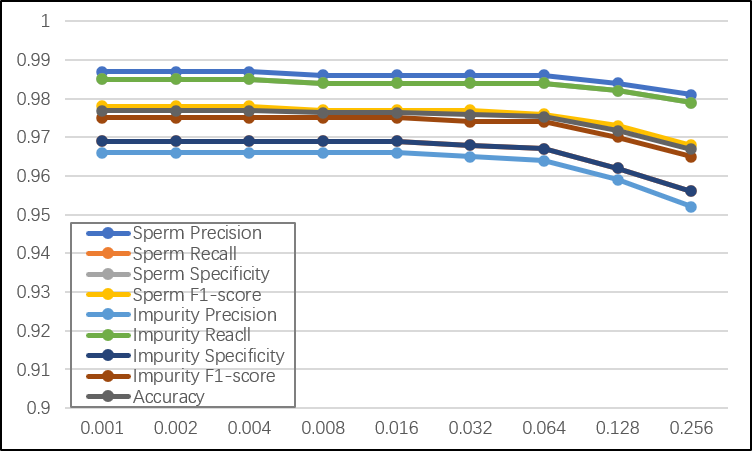}
\end{minipage}
}
\quad
\subfigure[ResNet-50]{
\begin{minipage}[t]{0.23\linewidth}
\includegraphics[width=4cm]{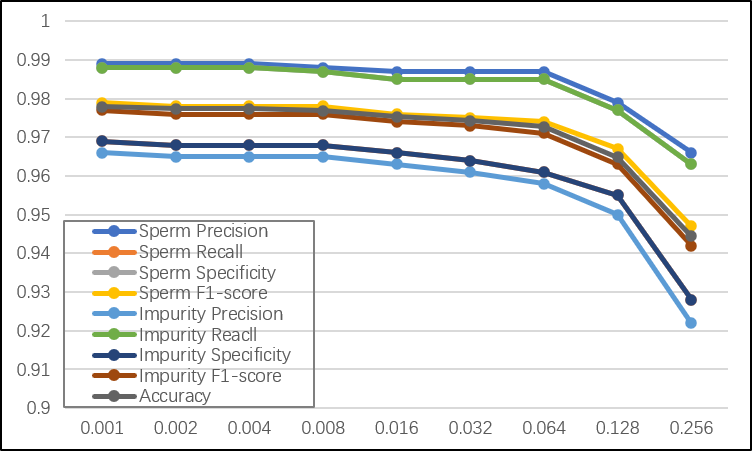}
\end{minipage}
}
\subfigure[ResNet-101]{
\begin{minipage}[t]{0.23\linewidth}
\includegraphics[width=4cm]{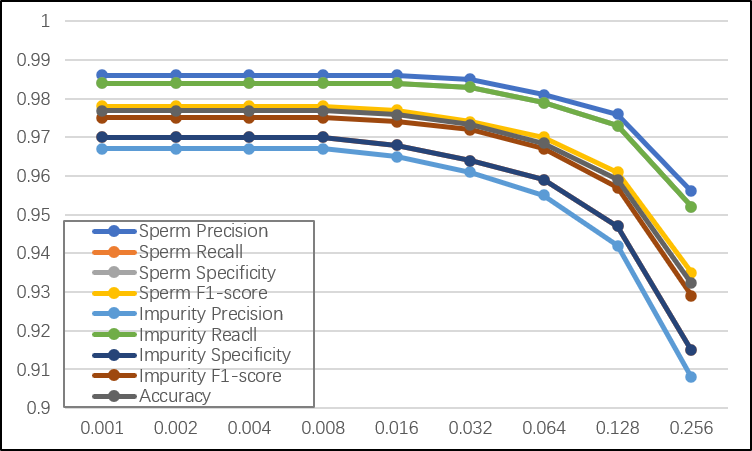}
\end{minipage}
}
\subfigure[GoogleNet]{
\begin{minipage}[t]{0.23\linewidth}
\includegraphics[width=4cm]{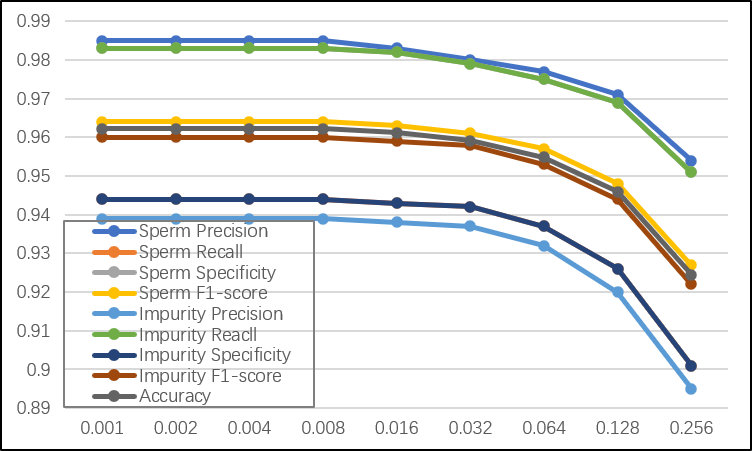}
\end{minipage}
}
\subfigure[DensenNet-121]{
\begin{minipage}[t]{0.23\linewidth}
\includegraphics[width=4cm]{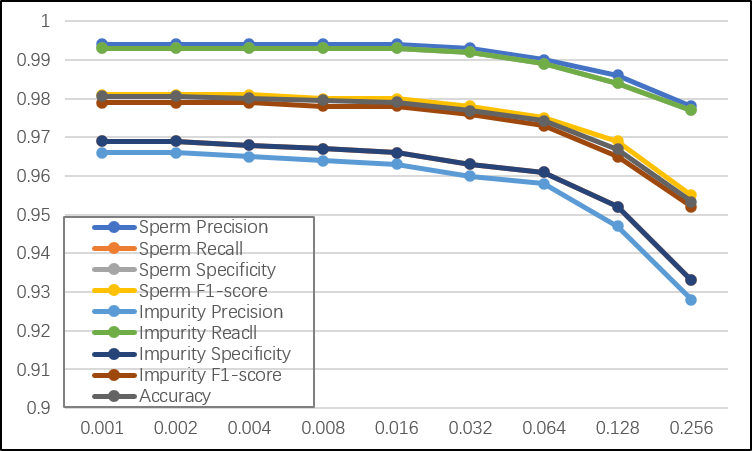}
\end{minipage}
}
\quad
\subfigure[Inception-V3]{
\begin{minipage}[t]{0.23\linewidth}
\includegraphics[width=4cm]{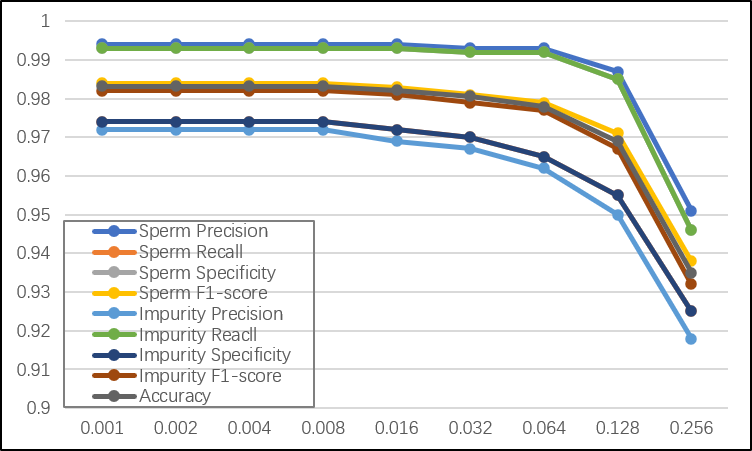}
\end{minipage}
}
\subfigure[MobileNet-V2]{
\begin{minipage}[t]{0.23\linewidth}
\includegraphics[width=4cm]{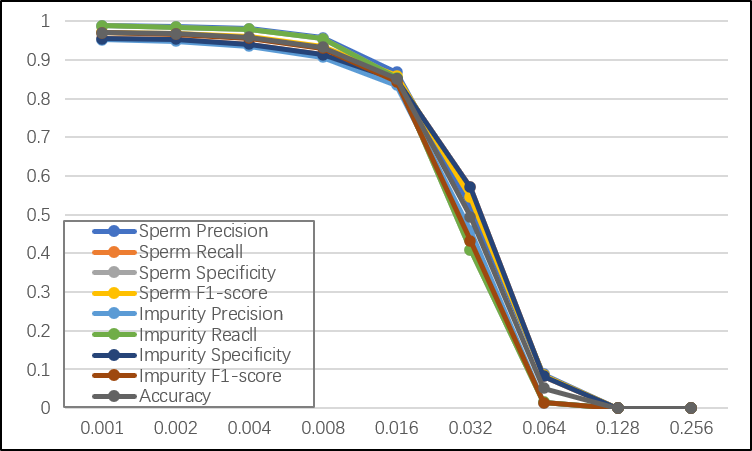}
\end{minipage}
}
\subfigure[ShuffleNet-V2]{
\begin{minipage}[t]{0.23\linewidth}
\includegraphics[width=4cm]{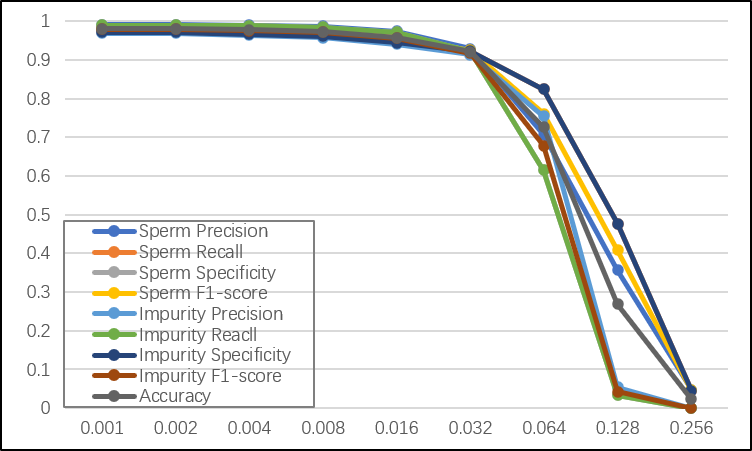}
\end{minipage}
}
\subfigure[Xception]{
\begin{minipage}[t]{0.23\linewidth}
\includegraphics[width=4cm]{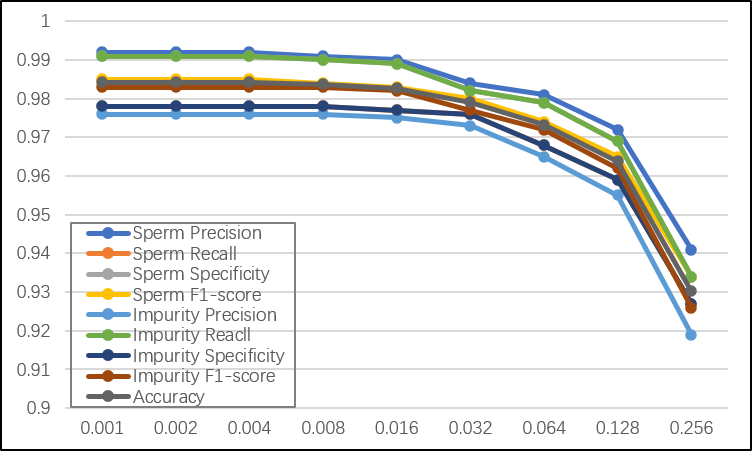}
\end{minipage}
}
\quad
\subfigure[ViT]{
\begin{minipage}[t]{0.23\linewidth}
\includegraphics[width=4cm]{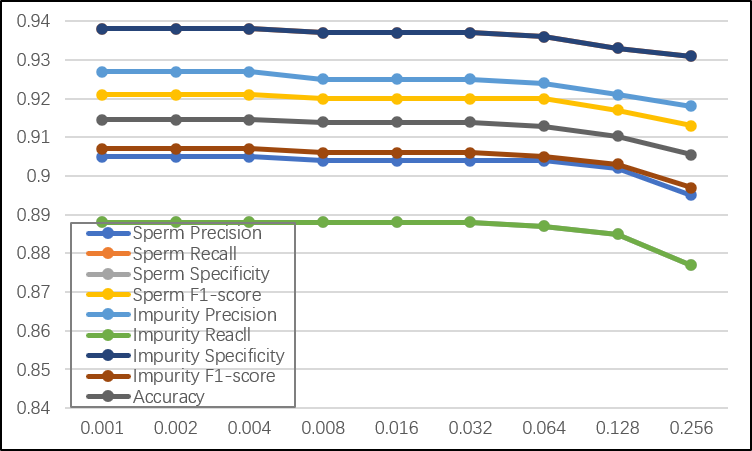}
\end{minipage}
}
\subfigure[BotNet]{
\begin{minipage}[t]{0.23\linewidth}
\includegraphics[width=4cm]{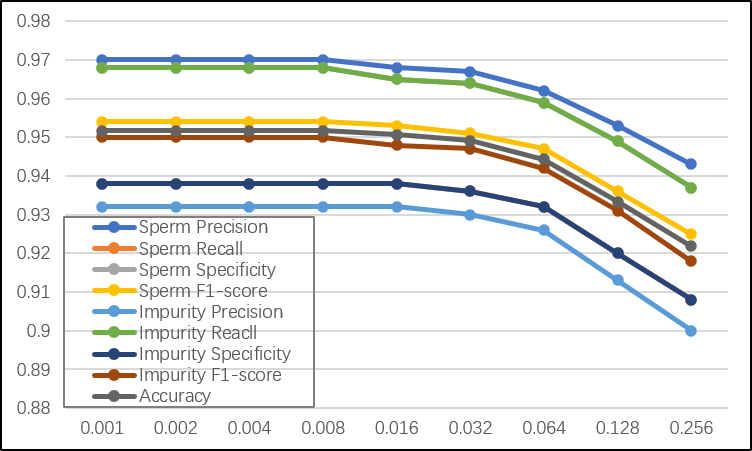}
\end{minipage}
}
\subfigure[DeiT-Base]{
\begin{minipage}[t]{0.23\linewidth}
\includegraphics[width=4cm]{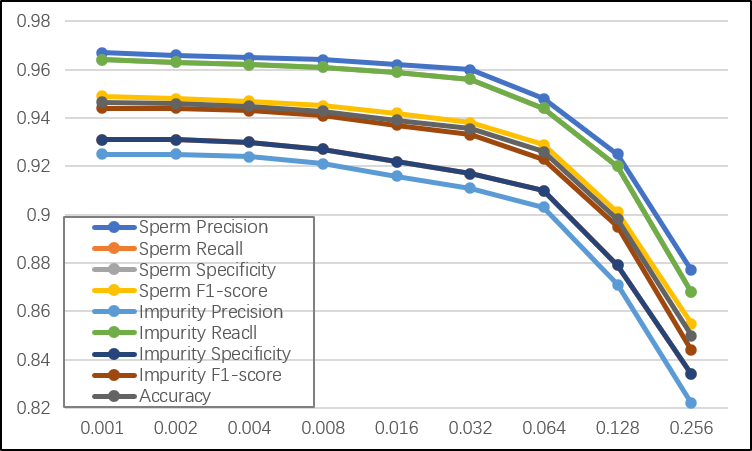}
\end{minipage}
}
\subfigure[DeiT-Tiny]{
\begin{minipage}[t]{0.23\linewidth}
\includegraphics[width=4cm]{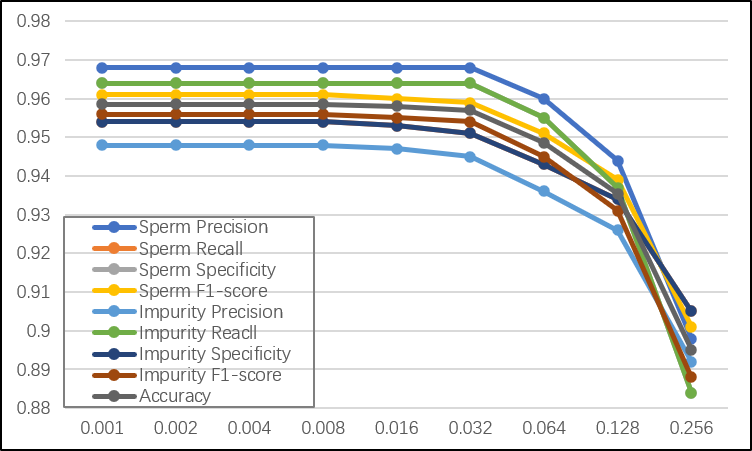}
\end{minipage}
}
\quad
\subfigure[T2T-ViT-t-19]{
\begin{minipage}[t]{0.23\linewidth}
\includegraphics[width=4cm]{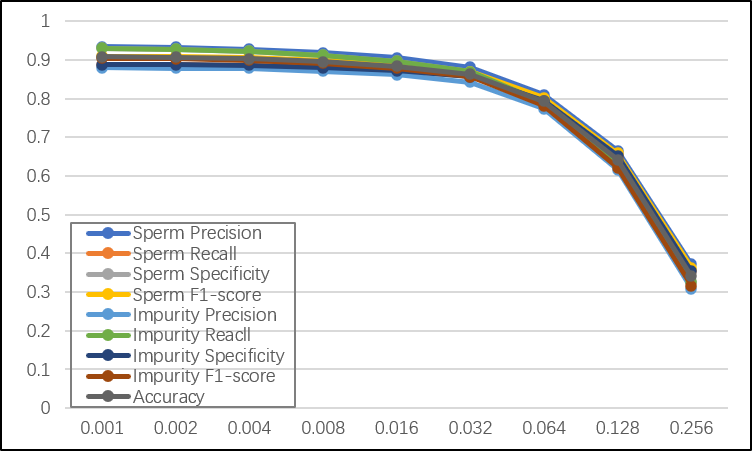}
\end{minipage}
}
\subfigure[T2T-ViT-t-24]{
\begin{minipage}[t]{0.23\linewidth}
\includegraphics[width=4cm]{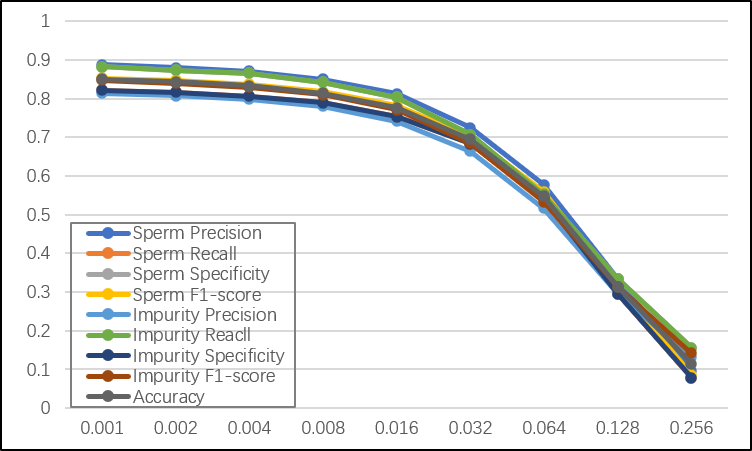}
\end{minipage}
}
\subfigure[T2T-ViT-7]{
\begin{minipage}[t]{0.23\linewidth}
\includegraphics[width=4cm]{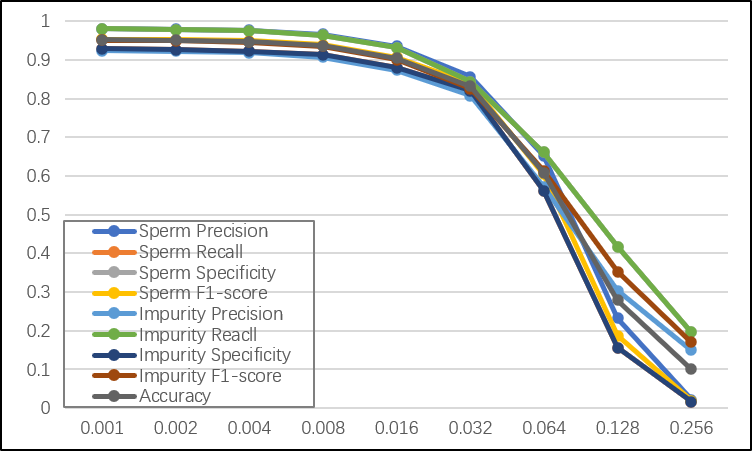}
\end{minipage}
}
\subfigure[T2T-ViT-10]{
\begin{minipage}[t]{0.23\linewidth}
\includegraphics[width=4cm]{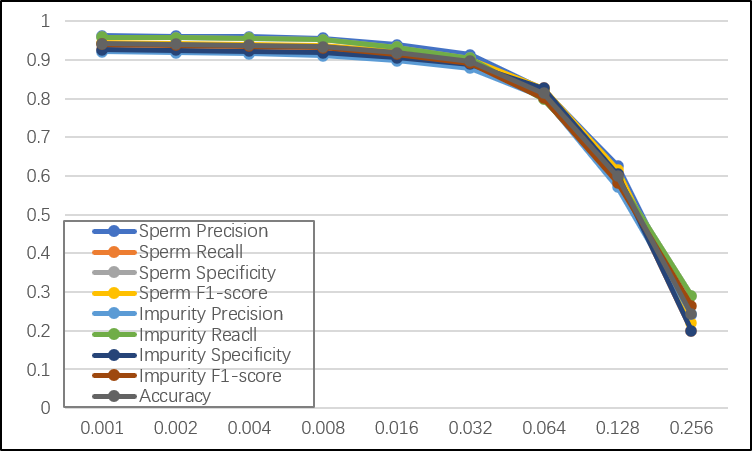}
\end{minipage}
}
\quad
\subfigure[T2T-ViT-12]{
\begin{minipage}[t]{0.23\linewidth}
\includegraphics[width=4cm]{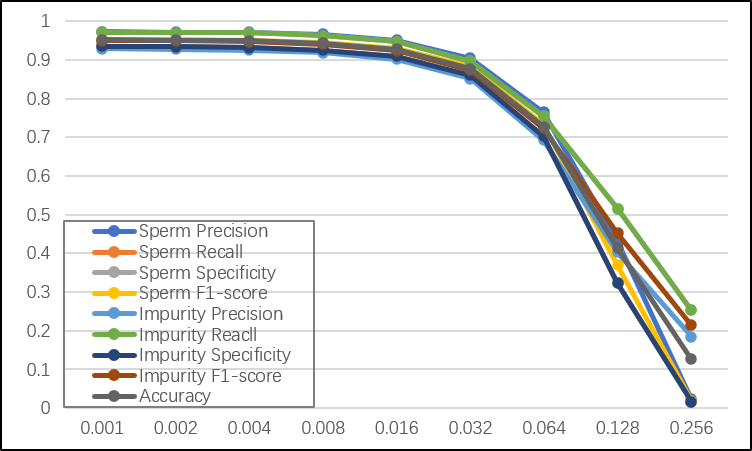}
\end{minipage}
}
\subfigure[T2T-ViT-14]{
\begin{minipage}[t]{0.23\linewidth}
\includegraphics[width=4cm]{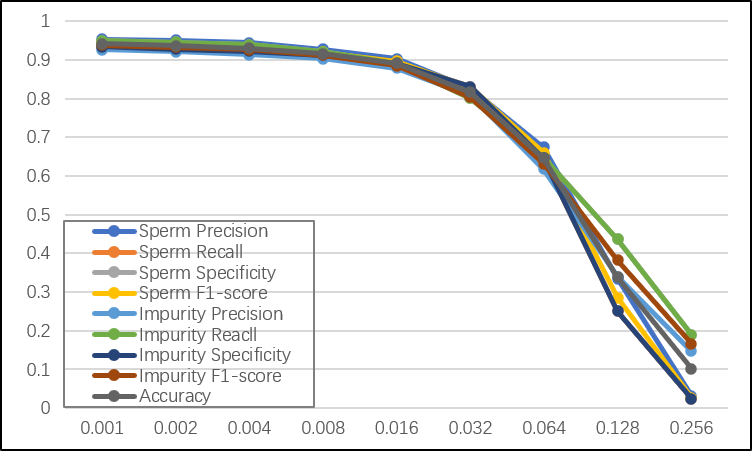}
\end{minipage}
}
\subfigure[T2T-ViT-19]{
\begin{minipage}[t]{0.23\linewidth}
\includegraphics[width=4cm]{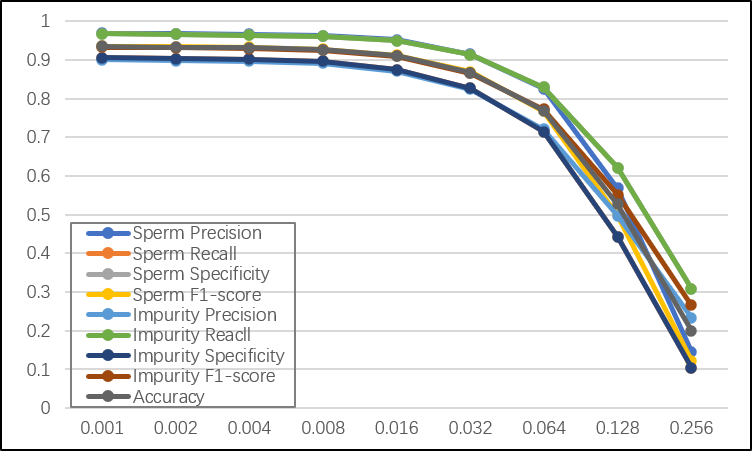}
\end{minipage}
}
\subfigure[T2T-ViT-24]{
\begin{minipage}[t]{0.23\linewidth}
\includegraphics[width=4cm]{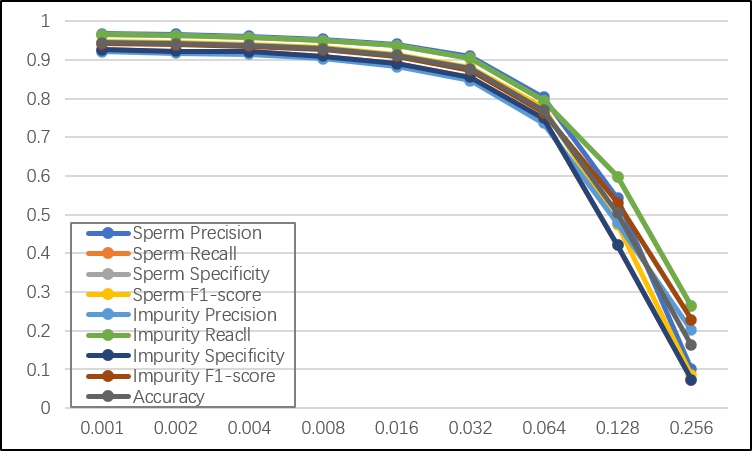}
\end{minipage}
}
\caption{Evaluation metrics curves of deep learning models under I-FGSM. Here, (a) exhibits the evaluation  metrics curves of AlexNet model. Similarly, (b) presents the evaluation metrics curves of VGG-11 model, (c) is generated on VGG-16, (d) is on VGG-19, (e) is on ResNet-50,  (f) is on ResNet-101, (g) is on GoogleNet, (h) is on DensenNet-121, (i) is on Inception-V3, (j) is on MobileNet-V2, (k) is on ShuffleNet-V2, (l) is on Xception, (m) is on ViT, (n) is on BotNet, (o) is on DeiT-Base, (p) is on DeiT-Tiny, (q) is on T2T-ViT-t-19, (r) is on T2T-ViT-t-24, (s) is on T2T-ViT-7, (t) is on T2T-ViT-10, (u) is on T2T-ViT-12, (v) is on T2T-ViT-14, (w) is on T2T-ViT-19, (x) is on T2T-ViT-24.}
\label{FIG:7}
\end{figure*}

Fig.~\ref{FIG:7} shows the variation of each evaluation metric in CNN and VT models with the increase of $\epsilon$ under the I-FGSM adversarial attack. First of all, we observe the evaluation metrics curves of each model in Fig.~\ref{FIG:7}, and it can be known that the change trend of evaluation metrics of AlexNet, VGG-11, VGG-16, VGG-19, ResNet-50, ResNet-101, GoogleNet, DensenNet-121, Inception-V3, Xception, ViT, BotNet, DeiT-Base and DeiT-Tiny are relatively smooth.\par

Then, when we focus on these models, it can be clearly seen from these histograms (as shown in Fig.~\ref{FIG:19} (a) \~{} (i)) in Appendix: when the parameter $\epsilon$ increases from 0.01 to 0.256 by multiplication, we can find that under AlexNet, sperm recall, impurity precision and impurity specificity change about 4.8 \%, sperm F1-score, impurity F1-score and accuracy change about 4 \%, and other evaluation metrics change 2.8 \%. Under VGG-11, sperm precision, sperm specificity and impurity recall changes 1.1 \%, 1.2 \% and 1.2 \%, respectively, sperm F1-score, impurity precision and accuracy changes 1.3 \%, 1.8 \% and 1.42 \%, respectively, other evaluation metrics change 1.5 \%. Under VGG-16, all evaluation metrics change 0.8 \%. Under VGG-19, sperm precision, sperm specificity and impurity recall change 0.6 \%, sperm F1-score, impurity F1-score and accuracy change 1 \%, other evaluation metrics change about 1.3 \%. Under ResNet-50, sperm precision, sperm specificity and impurity recall change about 2.5 \%, sperm F1-score, impurity F1-score and accuracy change about 3.4 \%, other evaluation metrics change about 4.2 \%. Under ResNet-101, sperm precision, sperm specificity and impurity recall change 3.1 \%, sperm F1-score, impurity F1-score and accuracy change about 4.5 \%, other evaluation metrics change about 5.7 \%. Under GoogleNet, sperm precision, sperm specificity and impurity recall changes 3.1 \%, 3.2 \% and 3.2, respectively, sperm F1-score, impurity F1-score and accuracy change about 3.8 \%, other evaluation metrics change about 4.3 \%. Under DensenNet-121, sperm precision, sperm specificity and impurity recall change 1.6 \%, sperm F1-score, impurity F1-score and accuracy change about 2.7 \%, other evaluation metrics change about 3.7 \%. Under Inception-V3, expect sperm precision changes 4.3 \%, other evaluation metrics change about 5 \%. Under Xception, sperm precision, sperm recall sperm F1-score and impurity specificity change 5.1 \%, accuracy changes 5.41 \%, other evaluation metrics change 5.7 \%.\par 

Meanwhile, under ViT, sperm recall and impurity specificity change 0.7 \%, other evaluation metrics change about 1 \%. Under BotNet, sperm precision changes 2.7 \%, other evaluation metrics change about 3 \%. Under DeiT-Base, sperm precision, sperm specificity and impurity recall changes 9.1 \%, 9.6 \% and 9.6 \%,respectively, other evaluation metrics change more than 10 \%. Under DeiT-Tiny, sperm recall and impurity specificity change 2.6 \%, sperm precision, sperm F1-score, impurity precision, impurity F1-score and accuracy changes 6.9 \%, 4.8 \%, 3.3 \%, 5.6 \% and 5.14 \%, respectively, other evaluation metrics change 8 \%.\par

Therefore, it can be known that under I-FGSM adversarial attack, as $\epsilon$ increases in multiplication from 0.01 to 0.256, in the CNN and VT models used in this paper, VGG-11/16/19 and ViT have strong anti-noise robustness for the classification of tiny object (sperm and impurity) image dataset.

\subsection{Discussion}
In Section 4.2, we compare the anti-noise robustness of the deep learning classification methods on a tiny object (sperm and impurity) image dataset. Moreover, we summarize the analysis of the experimental results in subsection 4.2. It can be known that all the models in this paper have poor robustness for the classification of tiny object (sperm and impurity) image dataset under Gaussian noise with the mean value of 0.2 and the variance of 0.01 or 0.05, periodic noise, Rayleigh noise, uniform noise, exponential noise, FGSM and DeepFool. However, ViT has strong robustness for the classification of tiny object (sperm and impurity) image dataset under other conventional noises and adversarial attacks. \par
Why does ViT strong robustness for the classification of tiny object (sperm and impurity) image dataset under some conventional noises and adversarial attacks? Our discussion is mainly as follows. In the work of image classification, ViT first divides the original image into blocks and develops them into sequences. Then, the sequences are input into the encoder part of the original transformer model. Finally, a full connection layer is connected to classify the images. By simply analyzing the mechanism of ViT for image classification, we can known that ViT uses attention to capture the global context information so as to establish a long-distance dependence on the object, and extract more powerful features. Moreover, it pays more attention to the global information, ignoring part of the noise interference. Meanwhile, the size of tiny objects in the image of tiny object image dataset is large, which leads to more attention to global information, and there is not much close connection between local areas.

\section{Conclusion and Future Work}
Tiny object classification is a challenging problem in the field of computer vision, and has attracted the attention of many researchers. Due to the development of deep learning, the image classification of tiny objects is developing faster and faster. In this paper, we use 12 kinds of CNN models and 12 kinds of VT models. In VT models, the performer module is used in t2t-vit-7 / 10 / 12 / 19 / 24. Meanwhile, 14 conventional noises and 4 kinds of adversarial attacks are introduced, and different evaluation metrics are used to compare the anti-noise robustness. The obtained results confirm that ViT is the most stable model in this paper.\par
Although CNNs and VTs have achieved good results in image classification tasks, there is still much room for development for tiny objects classification. First, researchers can choose the top-level CNN and VT models to find their respective advantages and combine their "best part" to enhance the final classification, so as to develop a model that combines the advantages of CNN and VT to analyze the classification performance of tiny objects. Second, there is a lack of publicly accessible tiny objects image datasets. Therefore, developing a publicly accessible tiny object image database will benefit future researchers.\par
In the prospect, we intend to develop a more effective structure combined with CNN and VT, in order to obtain better classification performance of tiny object images and improve the  anti-noise robustness of tiny object images in convention noises and adversarial attack. In addition, we will publish the dataset of tiny objects that have been produced and continue to produce datasets about tiny objects.
\section{Acknowledgements}
This work is supported by the "National Natural Science Foundation of China" (No. 61806047) and the "Fundamental Research Funds for the Central Universities" (No. N2019003). 
We also thank Miss. Zixian Li and Mr. Guoxian Li for their important discussion in this work.
\bibliographystyle{unsrt}
\bibliography{cas-dc-template}

 \onecolumn 
\section*{Appendix}
\begin{appendix}
\subsection*{Histograms of Each Evaluation Metrics for the Original Test Set and Test Set with Conventional Noises and Adversarial Attacks in CNN and VT models.}

\begin{figure}[h]
\flushleft
\includegraphics[scale=0.5]{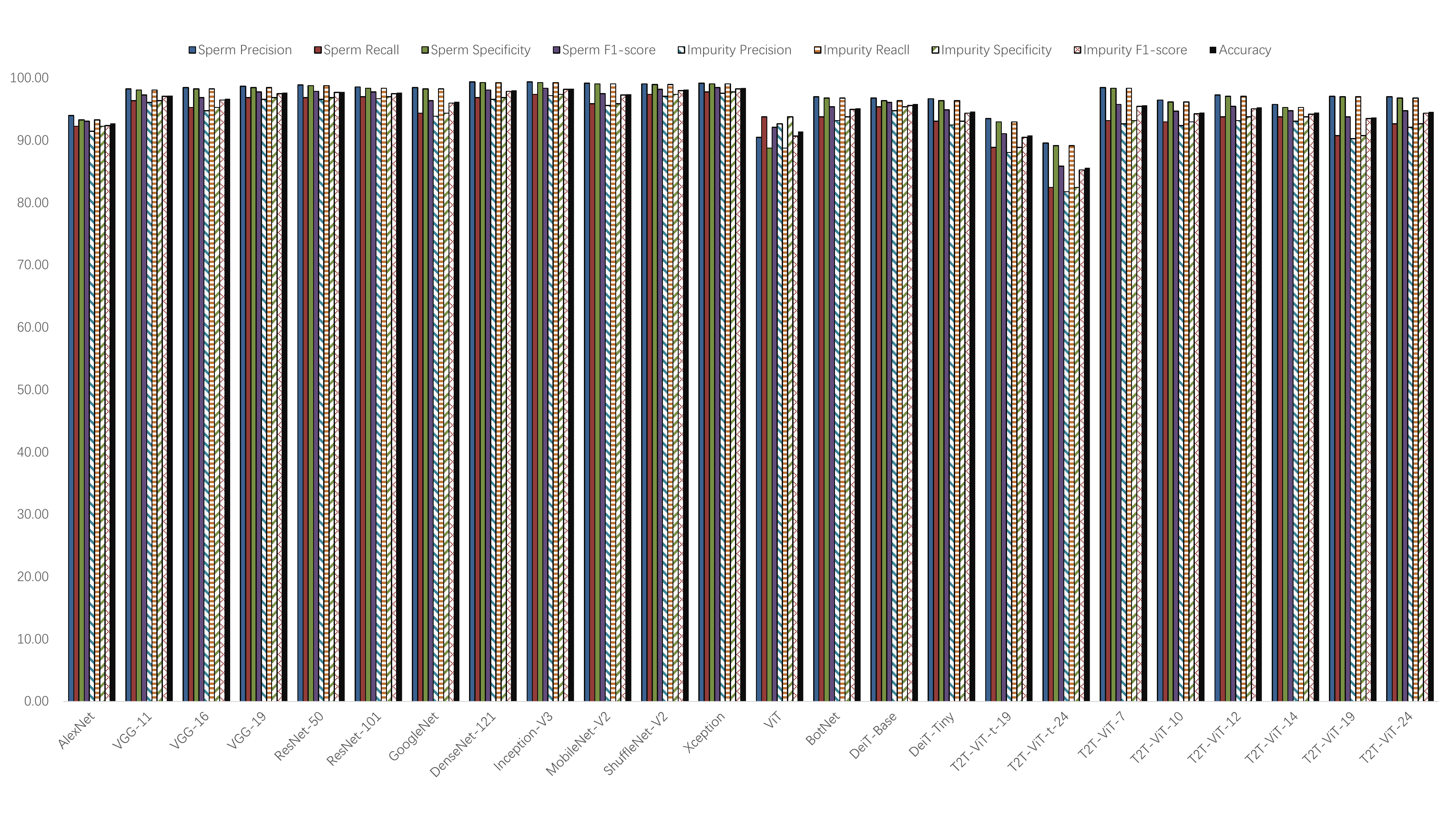}
\caption{The histogram of the evaluation metrics of the test dataset in each model.}
\label{FIG:8}
\end{figure} 

\begin{figure*}
\flushleft
\subfigure[Gaussian: A =$0$, B = $0.01$.]{
\includegraphics[scale=0.5]{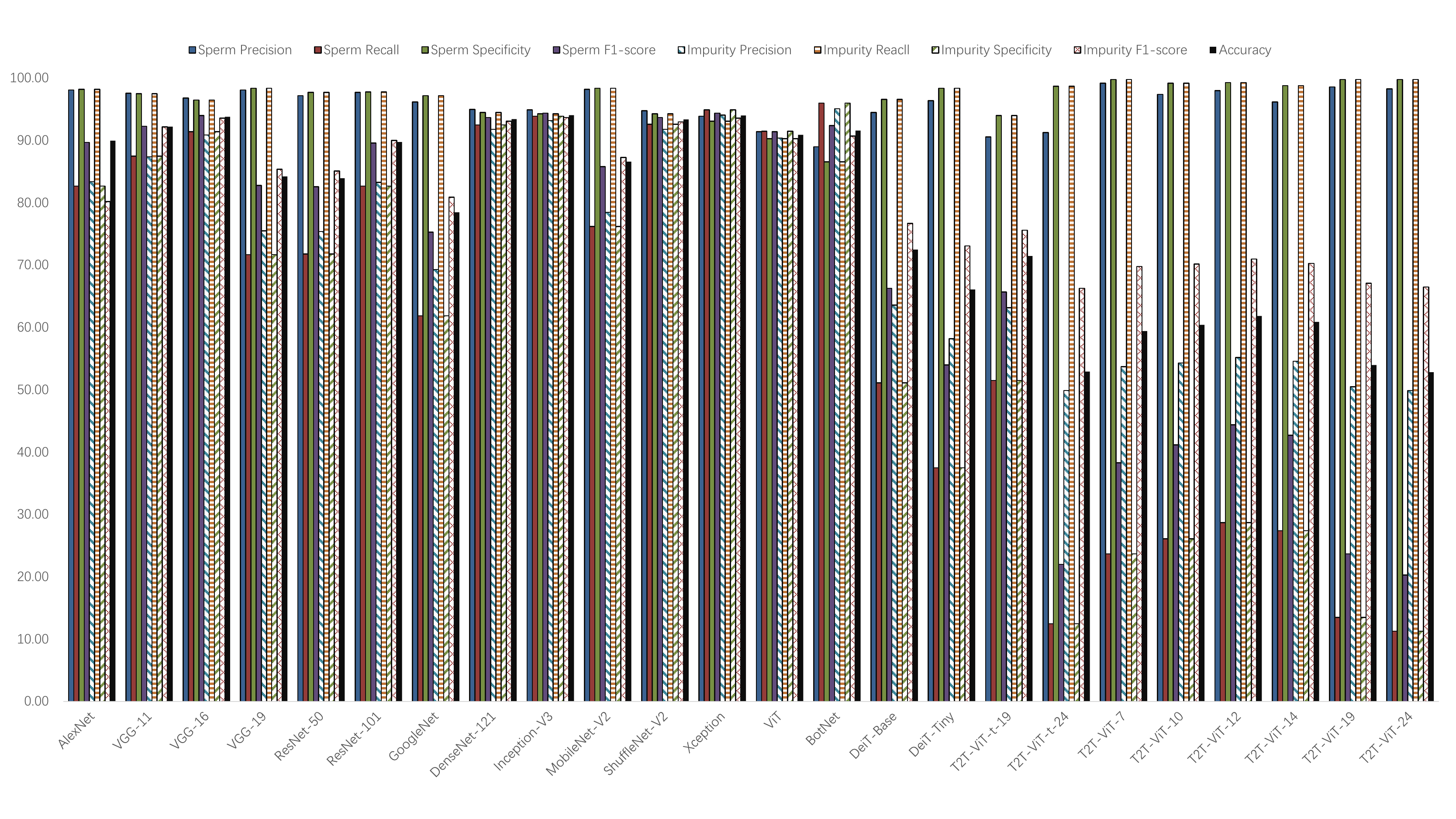}
}
\subfigure[Gaussian: A = $0$, B = $0.05$.]{
\includegraphics[scale=0.5]{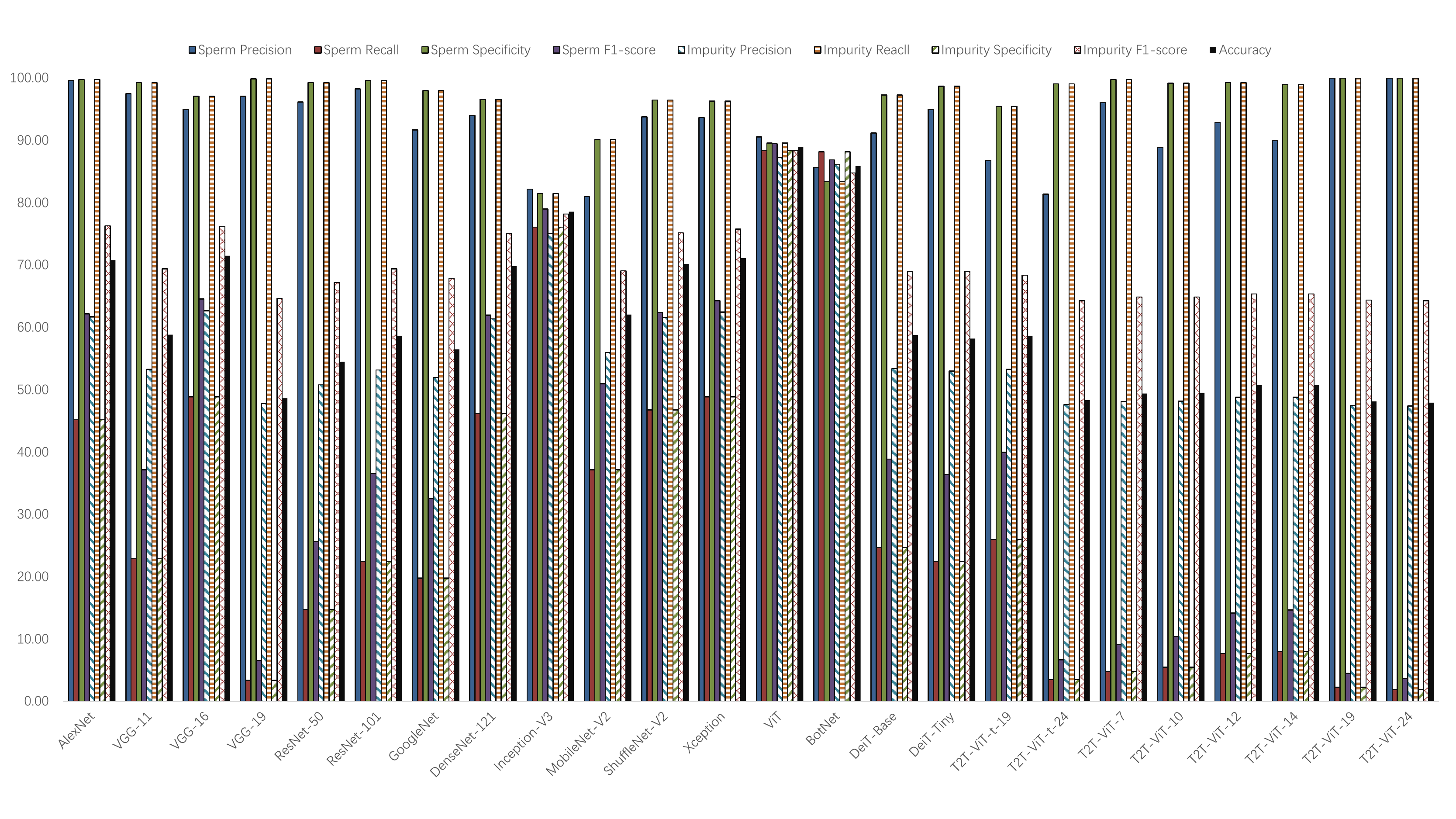}
}
\caption{The histogram of the evaluation metrics of the test dataset after adding Gaussian noise in each model. Here, A and B represent the mean value and variance respectively, (a) exhibit the evaluation metrics histogram of each model under the influence of Gaussian noise with mean value of $0$ and a variance of $0.01$, (b) exhibit the evaluation metrics histogram of each model under the influence of Gaussian noise with mean value of $0$ and a variance of $0.05$, (c) exhibit the evaluation metrics histogram of each model under the influence of Gaussian noise with mean value of $0.2$ and a variance of $0.01$, (d) exhibit the evaluation metrics histogram of each model under the influence of Gaussian noise with mean value of $0.2$ and a variance of $0.05$.}
\label{FIG:9}
\end{figure*} 
\addtocounter{figure}{-1}       
\begin{figure*} 
\flushleft
\addtocounter{subfigure}{2}      
\subfigure[Gaussian: A = $0.2$, B = $0.01$.]{
\includegraphics[scale=0.5]{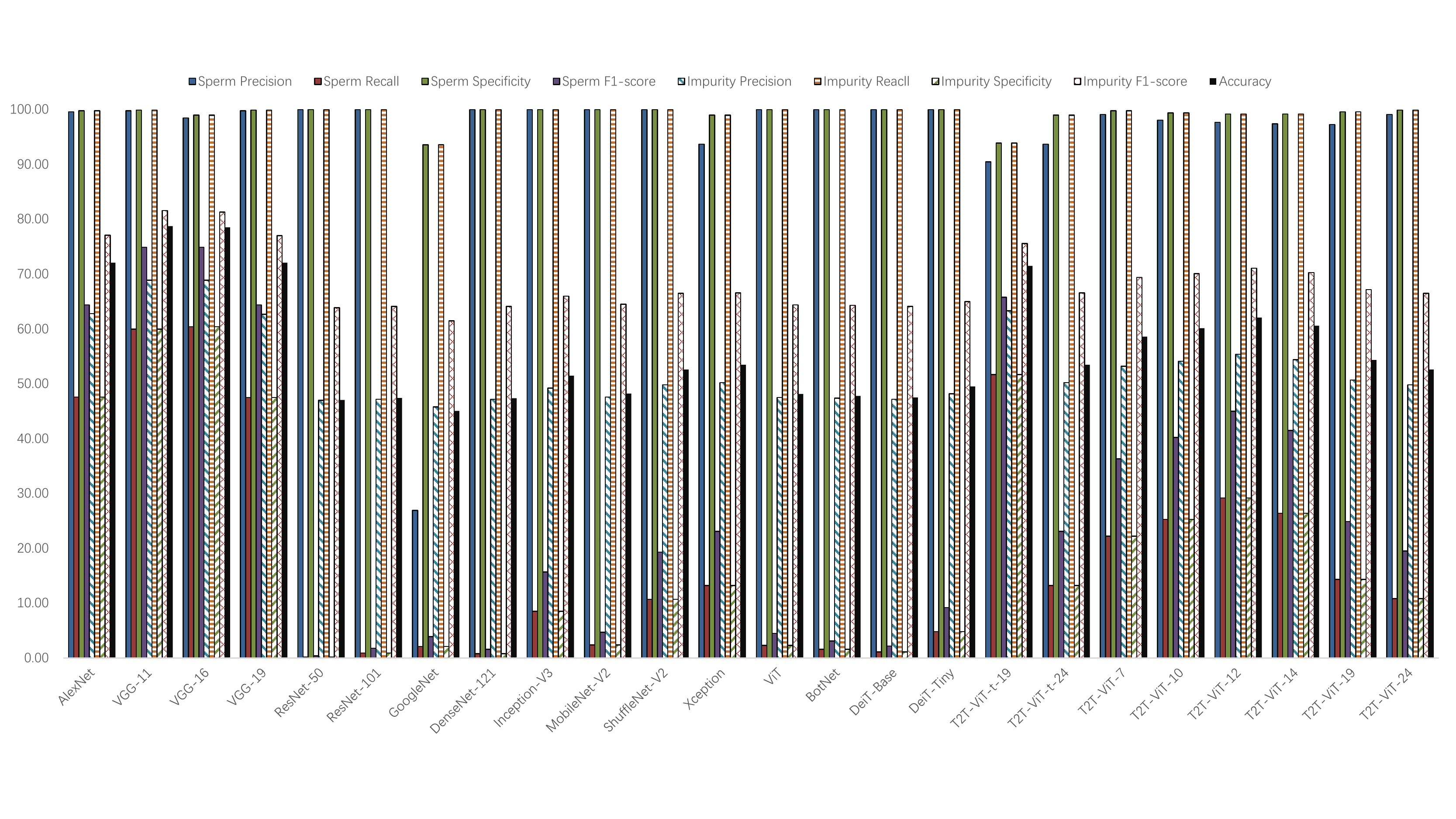}
}
\quad
\subfigure[Gaussian: A = $0.2$, B = $0.05$.]{
\includegraphics[scale=0.5]{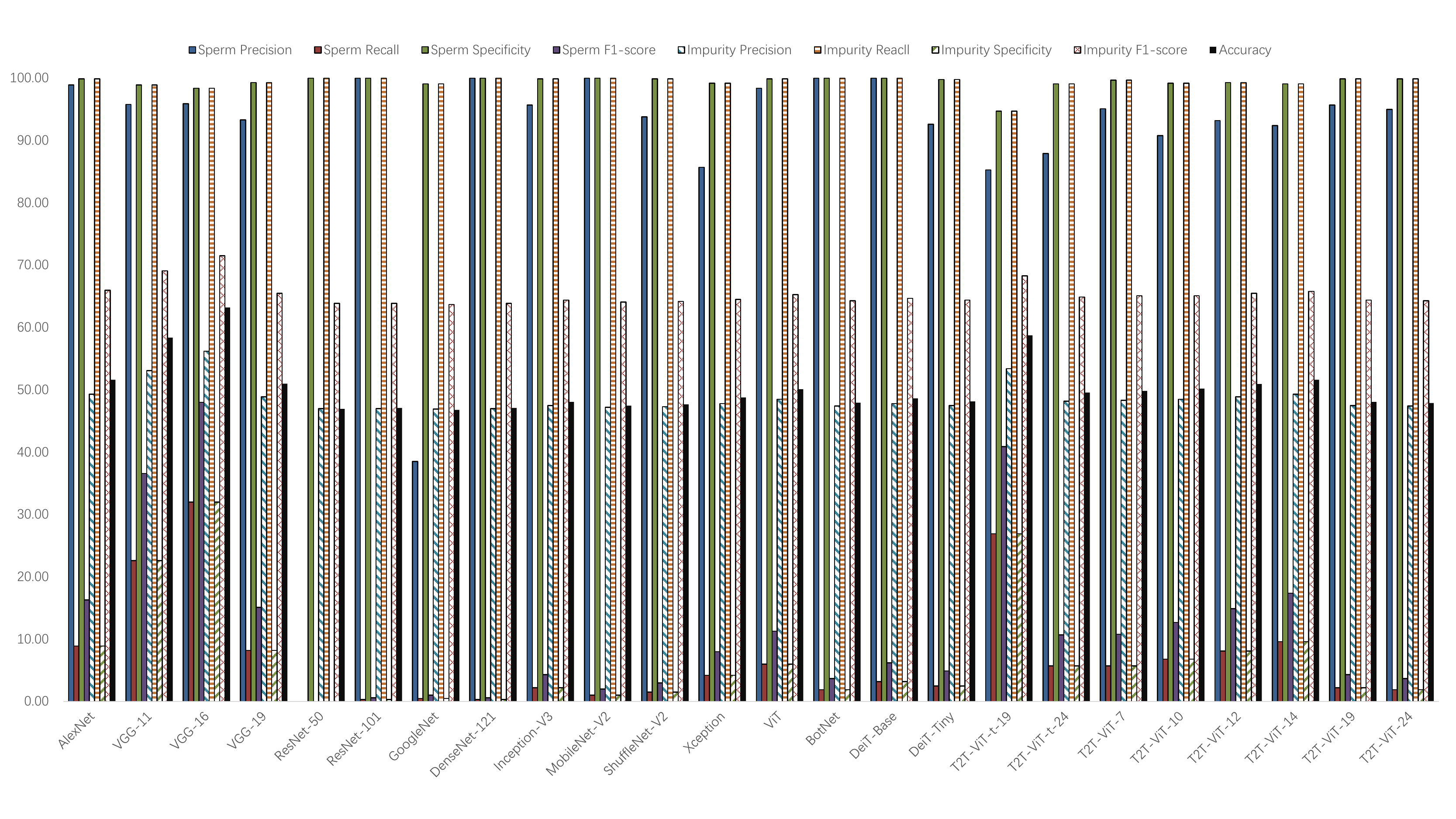}
}
\caption{The histogram of the evaluation metrics of the test dataset after adding Gaussian noise in each model. Here, A and B represent the mean value and variance respectively, (a) exhibit the evaluation metrics histogram of each model under the influence of Gaussian noise with mean value of $0$ and a variance of $0.01$, (b) exhibit the evaluation metrics histogram of each model under the influence of Gaussian noise with mean value of $0$ and a variance of $0.05$, (c) exhibit the evaluation metrics histogram of each model under the influence of Gaussian noise with mean value of $0.2$ and a variance of $0.01$, (d) exhibit the evaluation metrics histogram of each model under the influence of Gaussian noise with mean value of $0.2$ and a variance of $0.05$.}
\label{FIG:9}
\end{figure*}

\begin{figure*}
\flushleft
\subfigure[Speckle: A =$0$, B = $0.1$.]{
\includegraphics[scale=0.5]{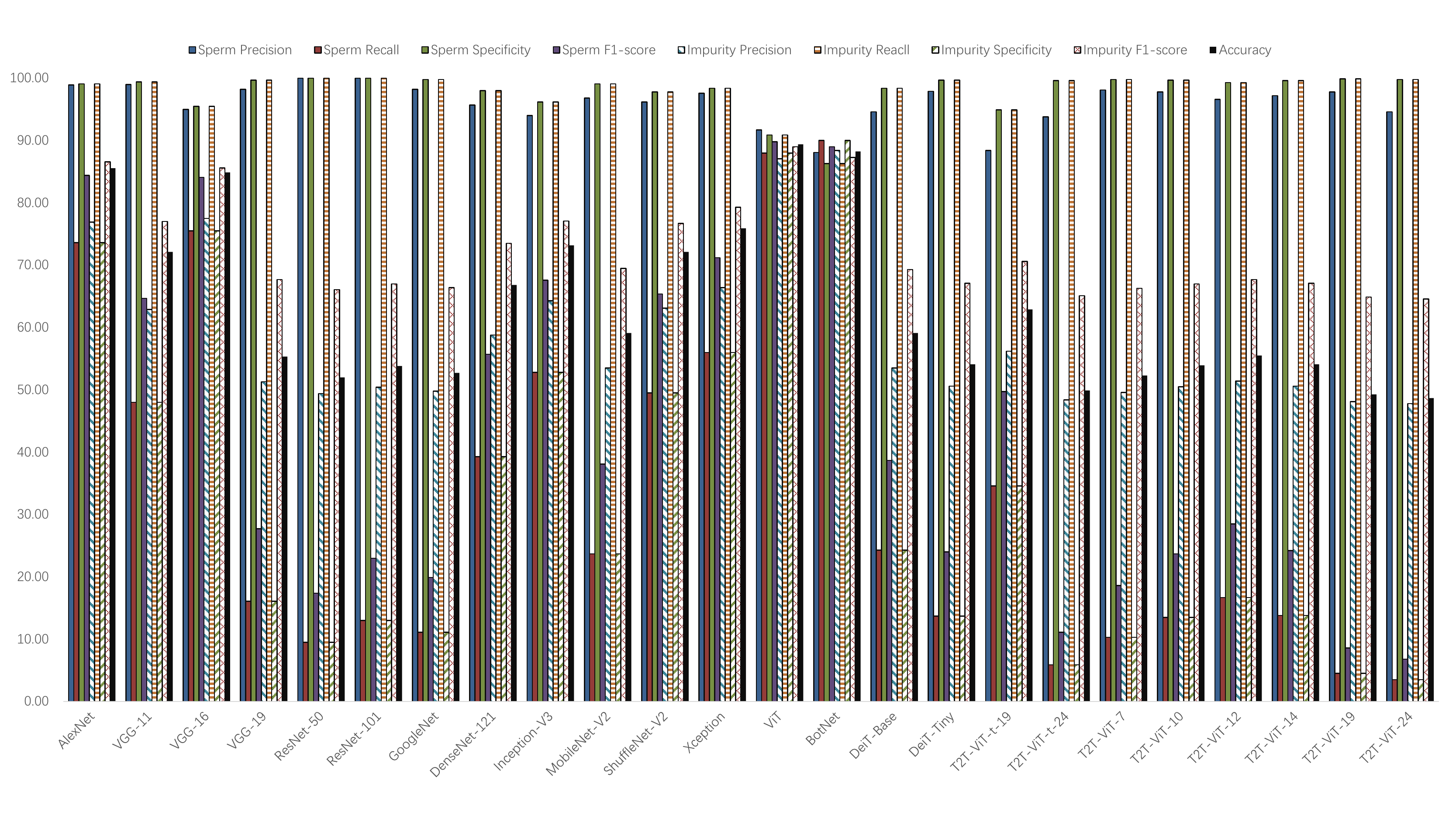}
}
\subfigure[Speckle: A = $0$, B = $0.05$.]{
\includegraphics[scale=0.5]{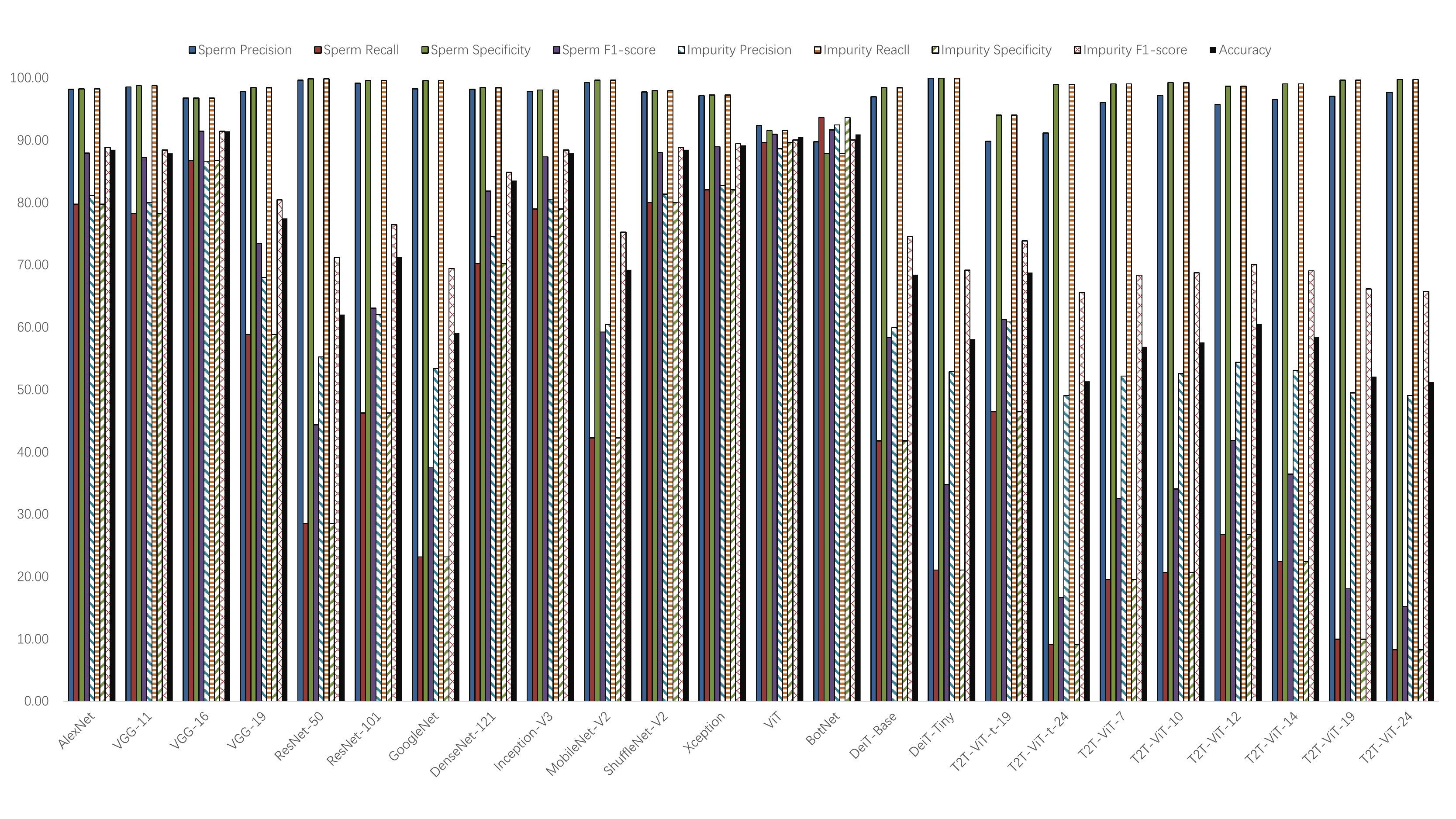}
}
\caption{The histogram of the evaluation metrics of the test dataset after adding Speckle noise in each model. Here, A and B represent the mean value and variance respectively, (a) exhibit the evaluation metrics histogram of each model under the influence of Speckle noise with mean value of $0$ and a variance of $0.1$, (b) exhibit the evaluation metrics histogram of each model under the influence of Speckle noise with mean value of $0$ and a variance of $0.05$.}
\label{FIG:10}
\end{figure*} 

\begin{figure*}
\flushleft
\subfigure[Salt and pepper: A = $0.1$.]{
\includegraphics[scale=0.5]{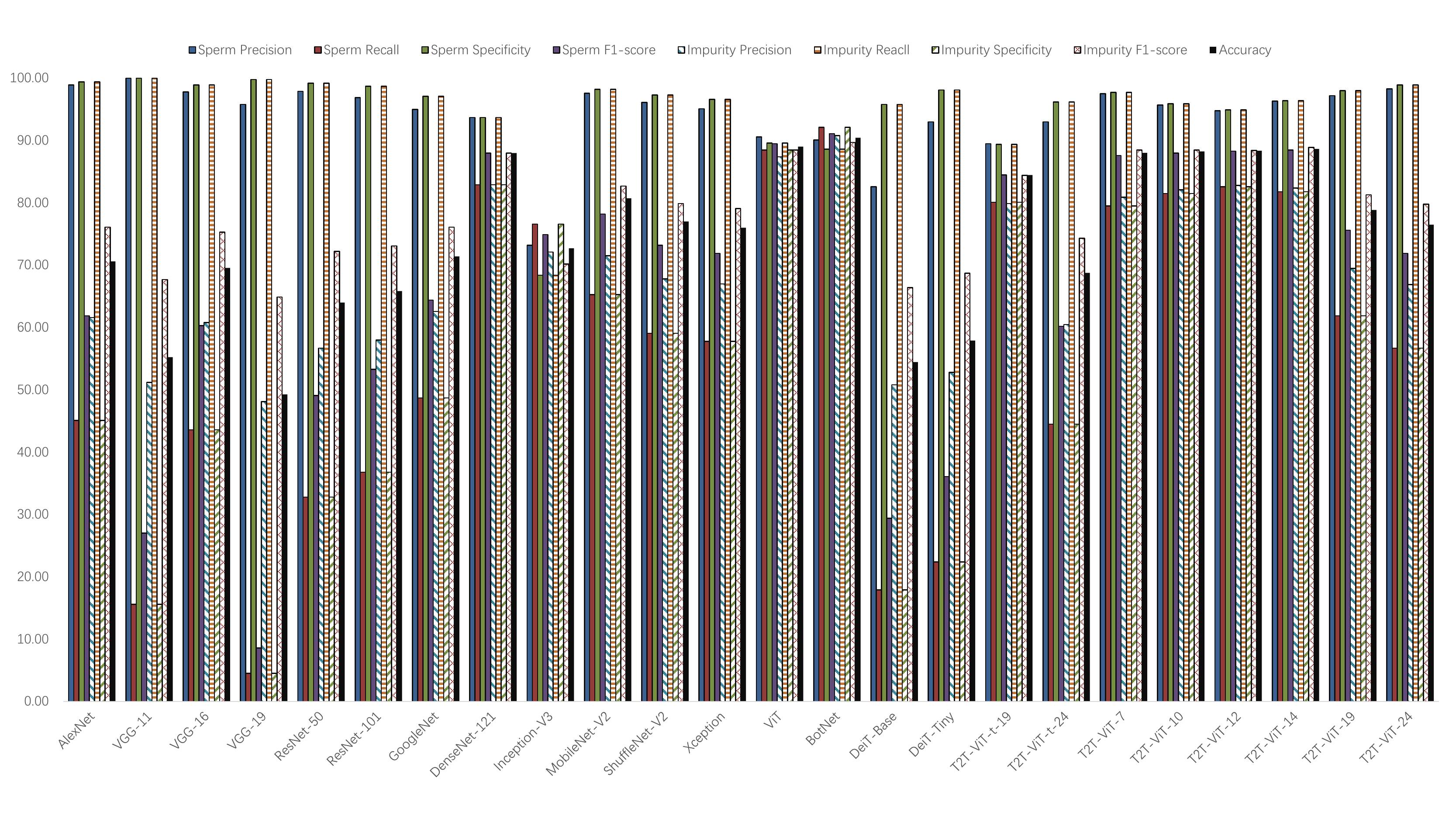}
}
\subfigure[Salt and pepper: A = $0.05$.]{
\includegraphics[scale=0.5]{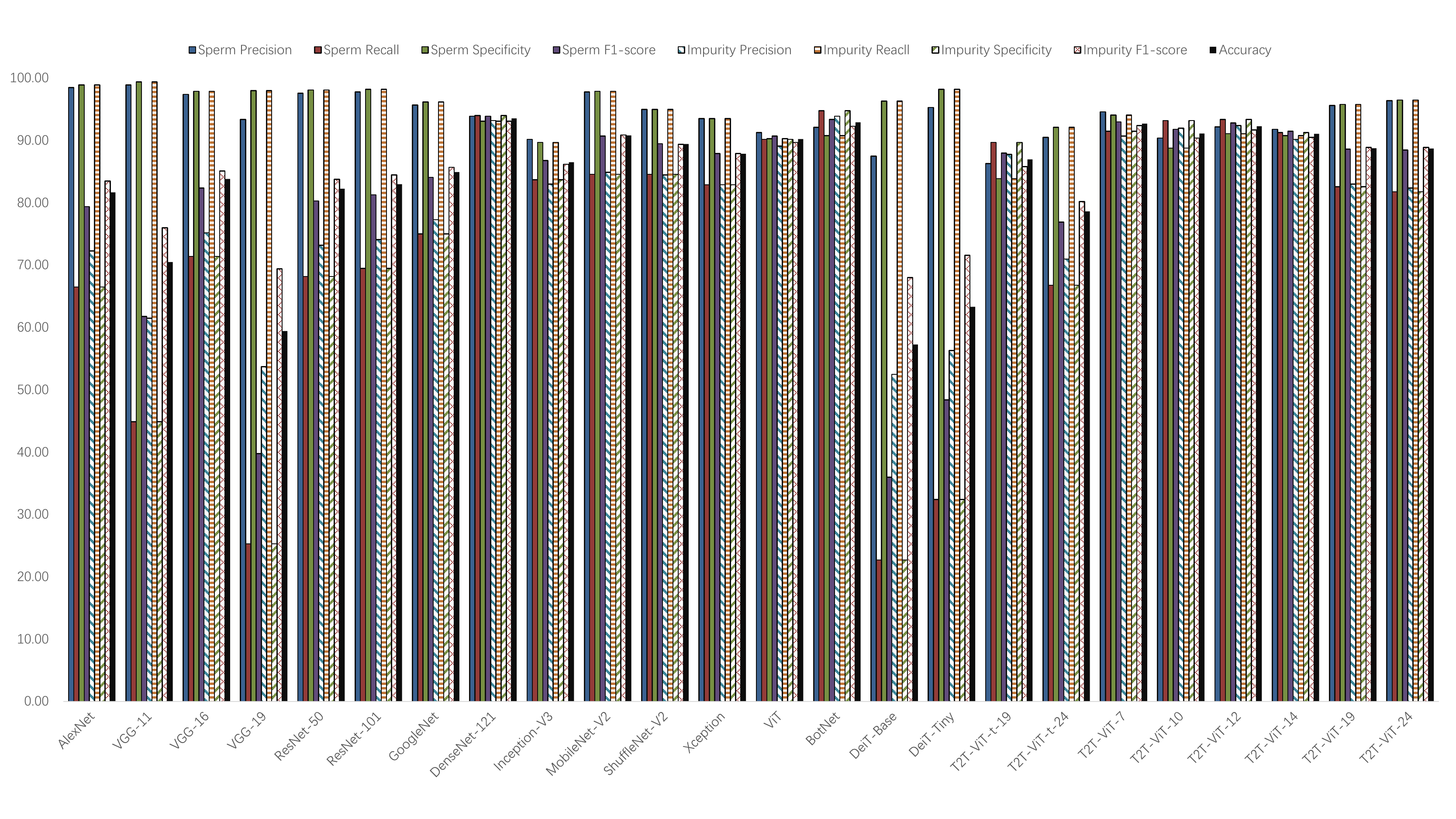}
}
\caption{The histogram of the evaluation metrics of the test dataset after adding Salt and pepper noise in each model. Here, A represent the mean value, (a) exhibit the evaluation metrics histogram of each model under the influence of Gaussian noise with mean value of $0.1$, (b) exhibit the evaluation metrics histogram of each model under the influence of Salt and pepper noise with mean value of $0.05$.}
\label{FIG:11}
\end{figure*} 

\begin{figure*}
\flushleft
\subfigure[Periodic: A = $50$, B = $50$.]{
\includegraphics[scale=0.5]{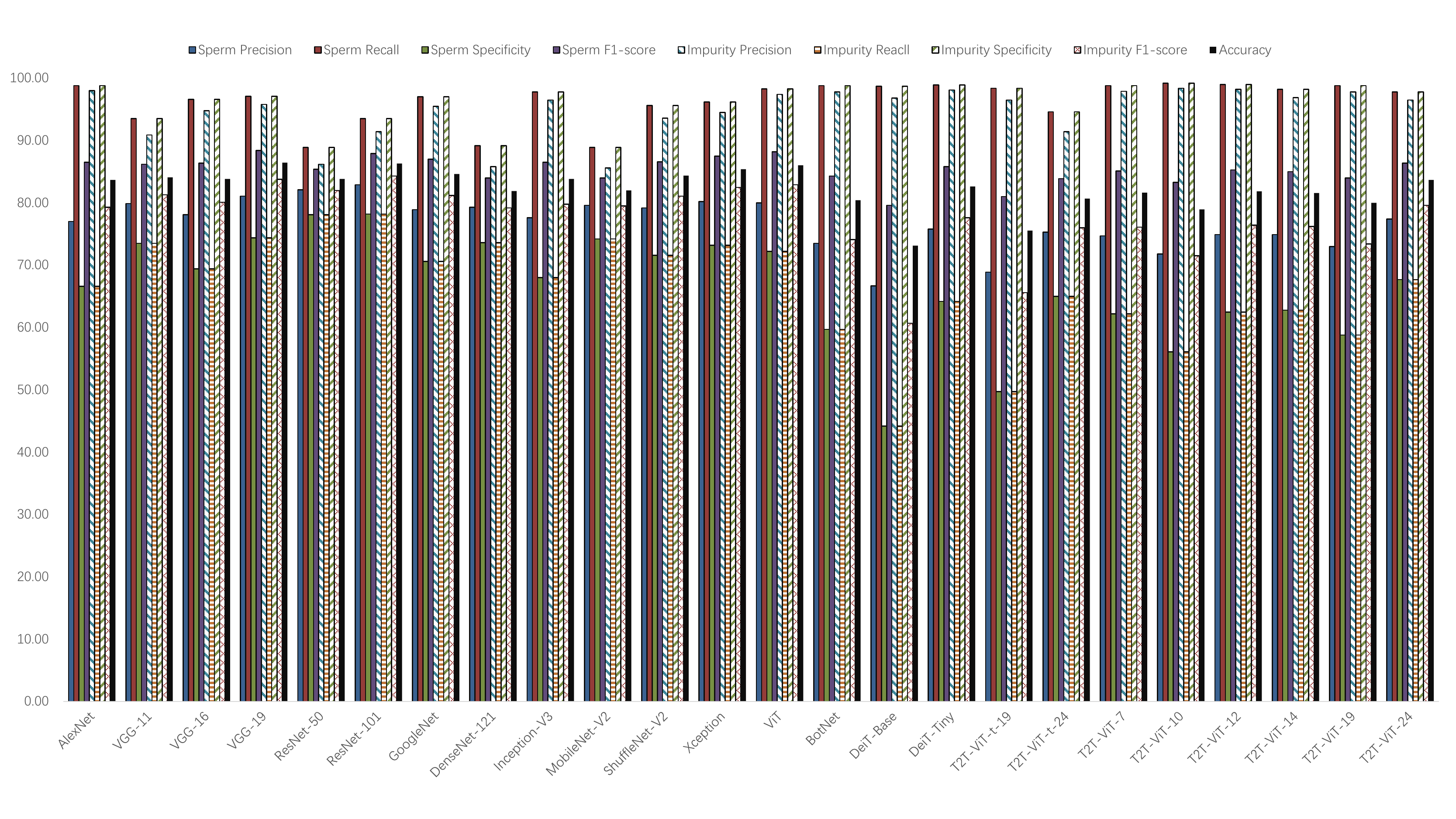}
}
\subfigure[Periodic:A = $50$, B = $40$.]{
\includegraphics[scale=0.5]{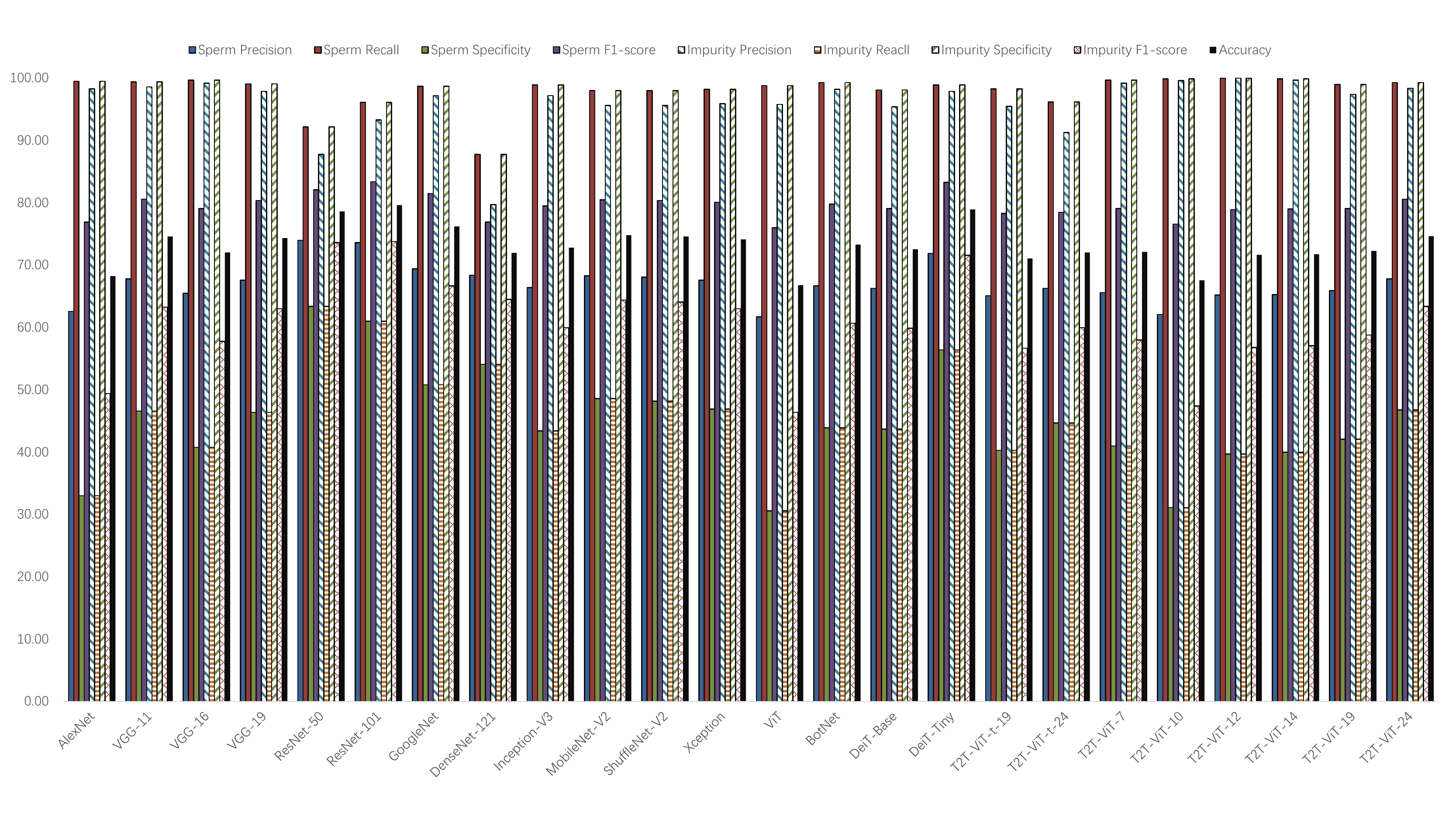}
}
\caption{The histogram of the evaluation metrics of the test dataset after adding Periodic noise in each model. Here, A and B represent sin amplitude and angle respectively, (a) exhibit the evaluation metrics histogram of each model under the influence of Periodic noise with sin amplitude value of $50$ and angle value of $50$, (b) exhibit the evaluation metrics histogram of each model under the influence of Periodic noise with sin amplitude value of $50$ and angle value of $40$.}
\label{FIG:12}
\end{figure*} 

\begin{figure*}
\flushleft
\subfigure[Uniform: A = $0$, B = $1$.]{
\includegraphics[scale=0.5]{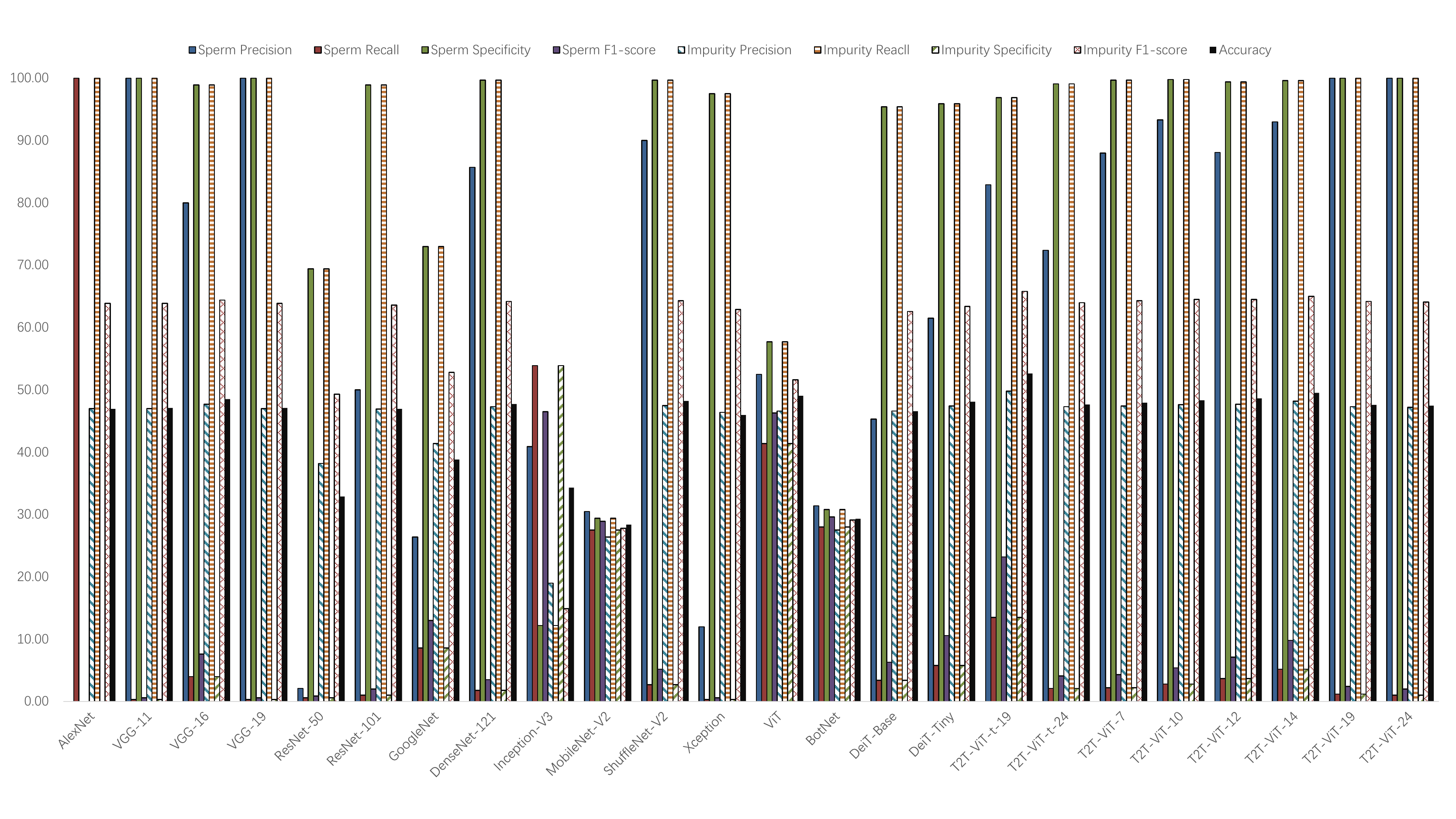}
}
\caption{The histogram of the evaluation metrics of the test dataset after adding Uniform noise in each model. Here, A and B represent  the mean value and variance respectively, (a) exhibit the evaluation metrics histogram of each model under the influence of Uniform noise with mean value of $0$ and a variance of $1$.}
\label{FIG:13}
\end{figure*} 

\begin{figure*}
\flushleft
\subfigure[Rayleigh: A = $0$, B = $1$.]{
\includegraphics[scale=0.5]{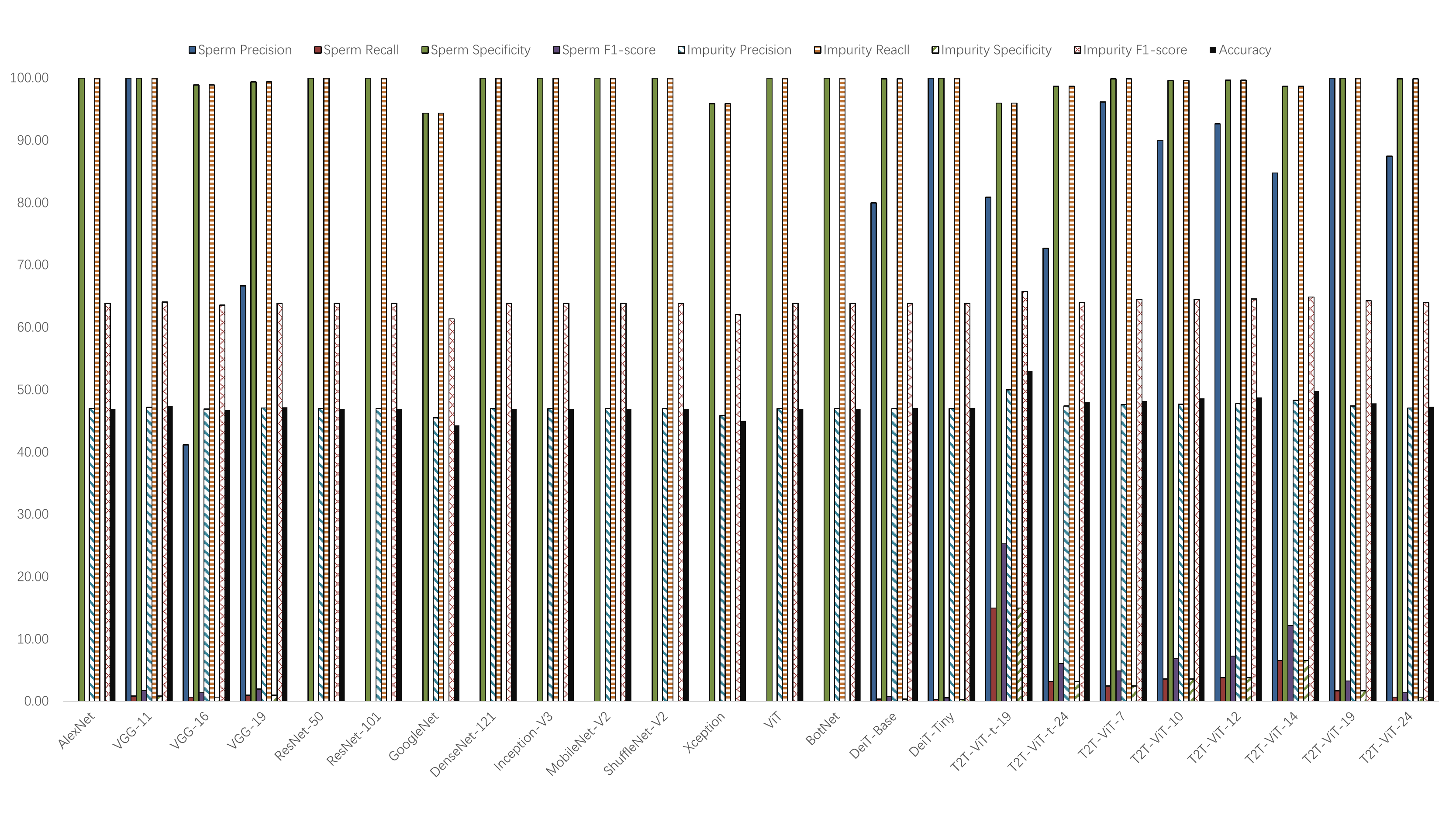}
}
\caption{The histogram of the evaluation metrics of the test dataset after adding Rayleigh noise in each model. Here, A and B represent  the mean value and variance respectively, (a) exhibit the evaluation metrics histogram of each model under the influence of Rayleigh noise with mean value of $0$ and a variance of $1$.}
\label{FIG:14}
\end{figure*} 

\begin{figure*}
\flushleft
\subfigure[Exponential: A = $1$.]{
\includegraphics[scale=0.5]{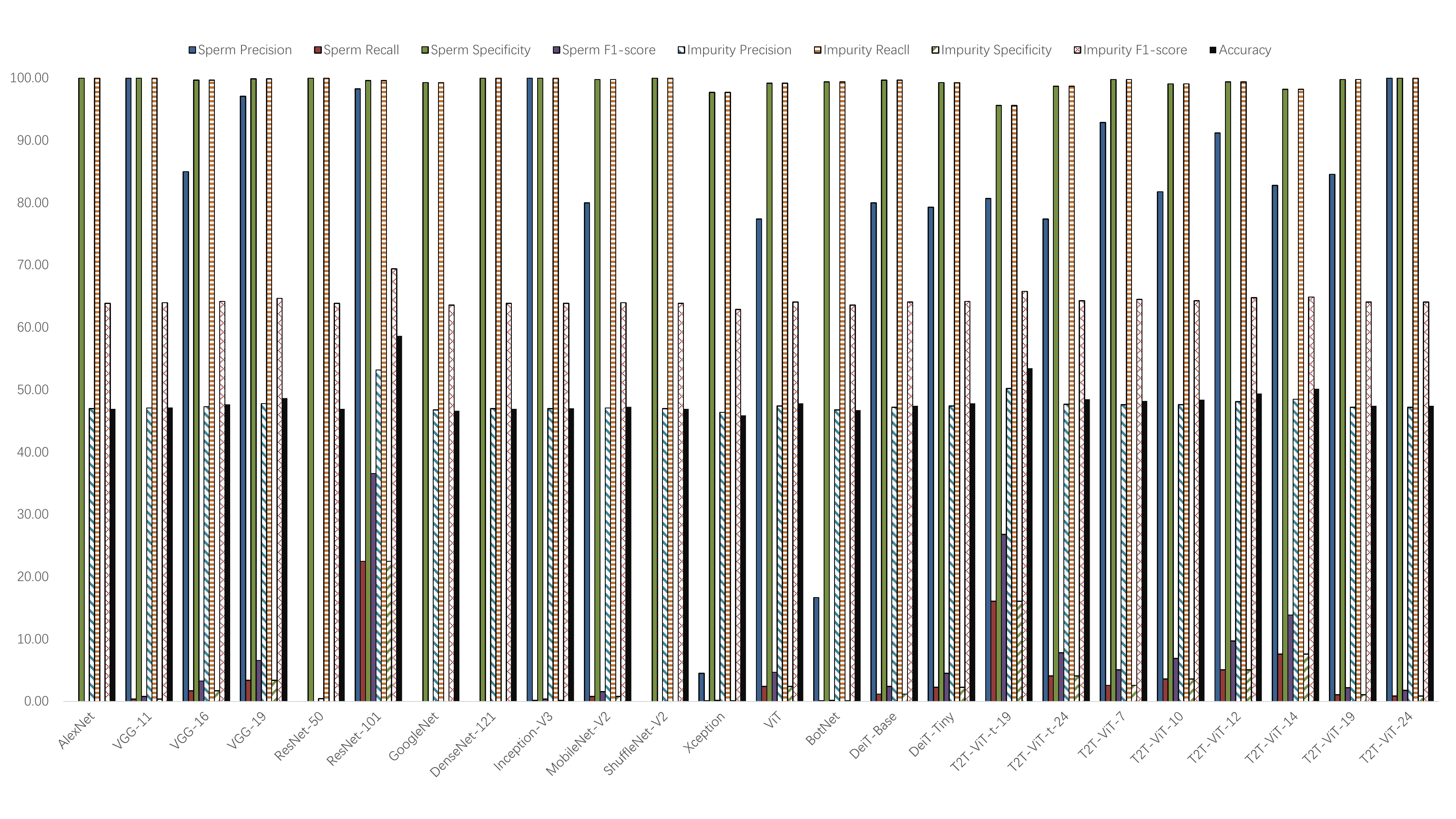}
}
\caption{The histogram of the evaluation metrics of the test dataset after adding Exponential noise in each model. Here, A represent  the mean value, (a) exhibit the evaluation metrics histogram of each model under the influence of Exponential noise with mean value of $1$.}
\label{FIG:15}
\end{figure*} 

\begin{figure*}
\flushleft
\subfigure[Poisson]{
\includegraphics[scale=0.5]{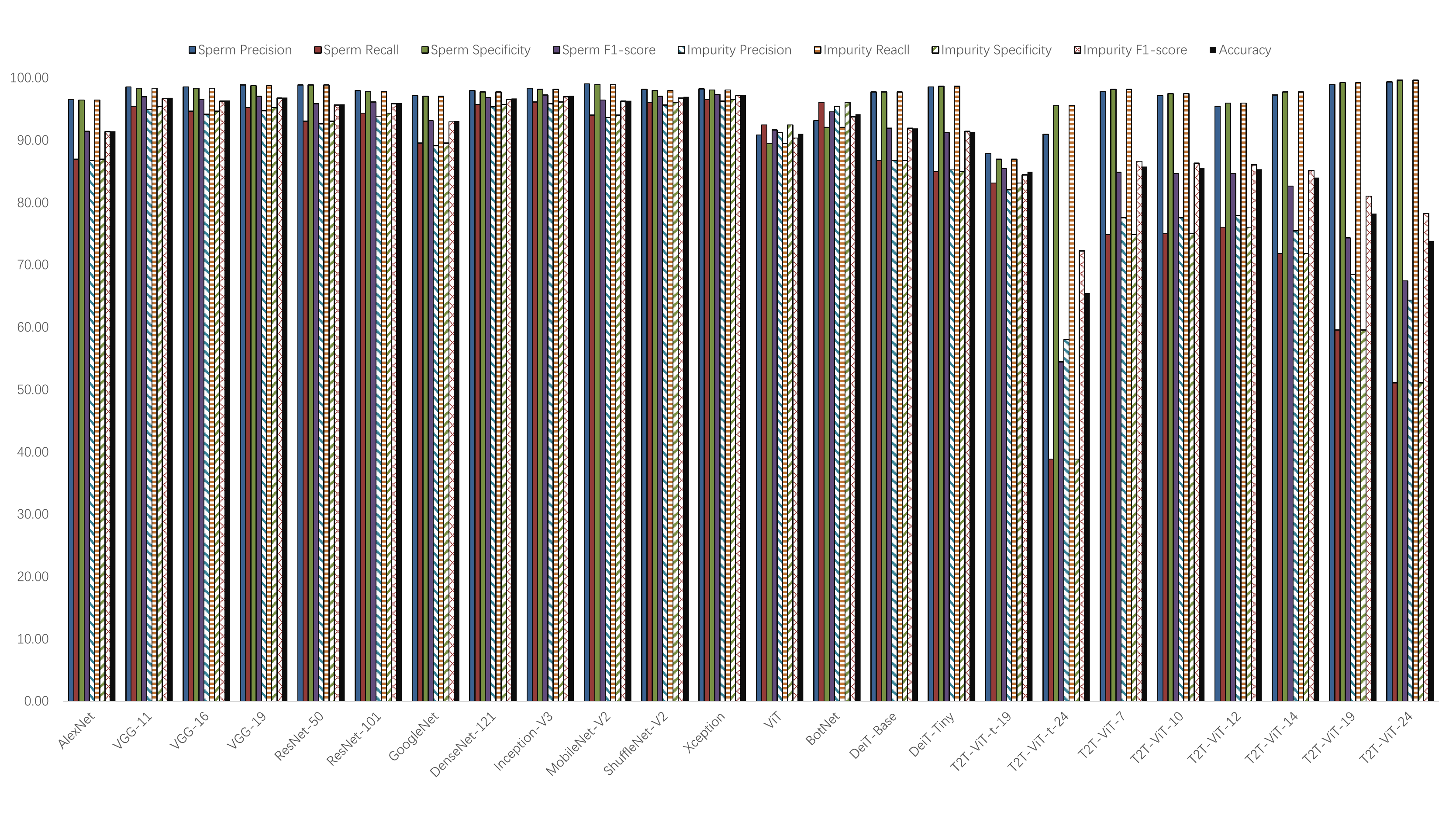}
}
\caption{The histogram of the evaluation metrics of the test dataset after adding Poisson noise in each model. Here, (a) exhibit the evaluation metrics histogram of each model under the influence of Poisson noise with default values}
\label{FIG:16}
\end{figure*} 

\begin{figure*}
\flushleft
\subfigure[FGM: $\epsilon$ = $0.001$.]{
\includegraphics[scale=0.5]{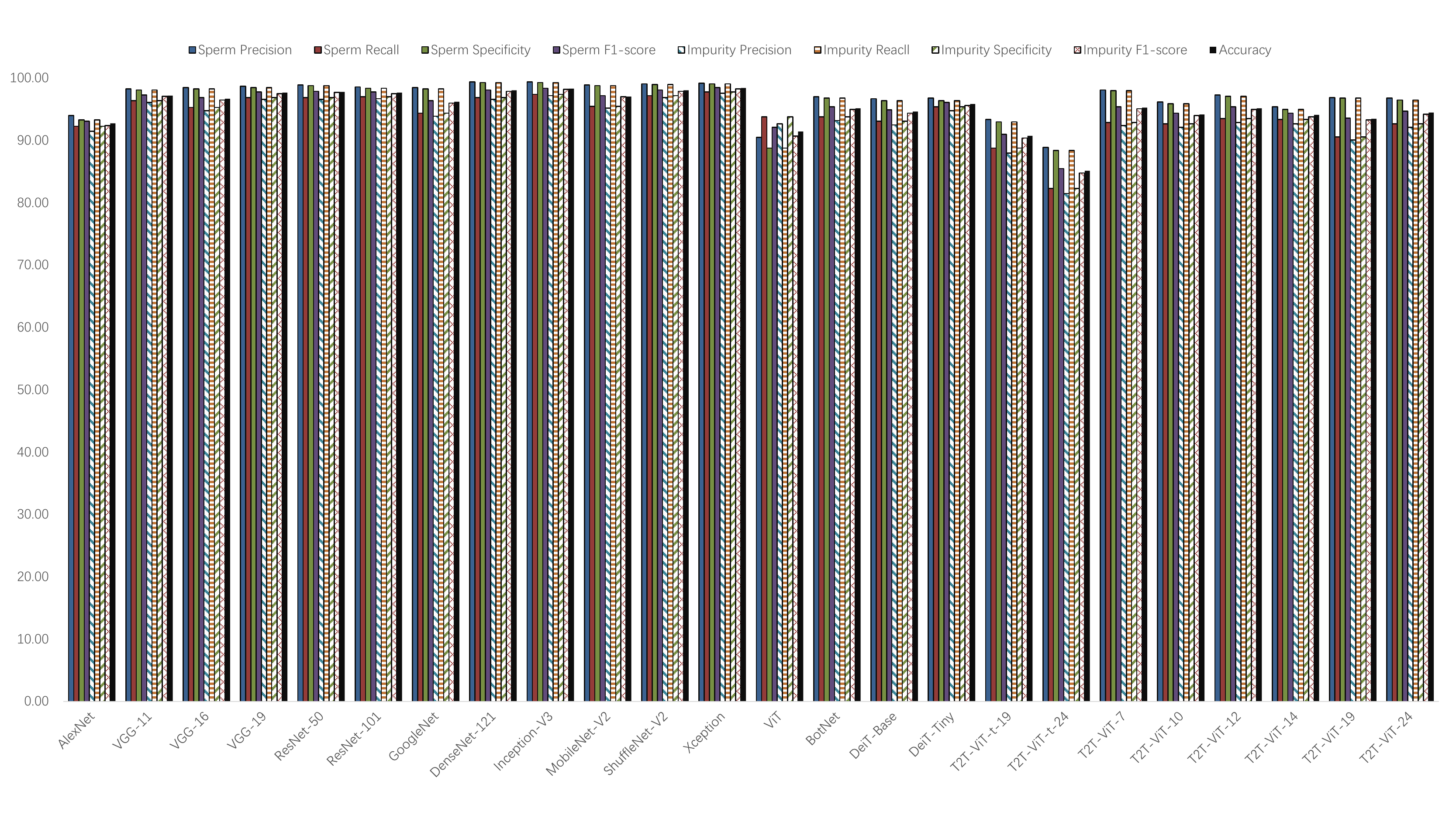}
}
\subfigure[FGM: $\epsilon$ = $0.002$.]{
\includegraphics[scale=0.5]{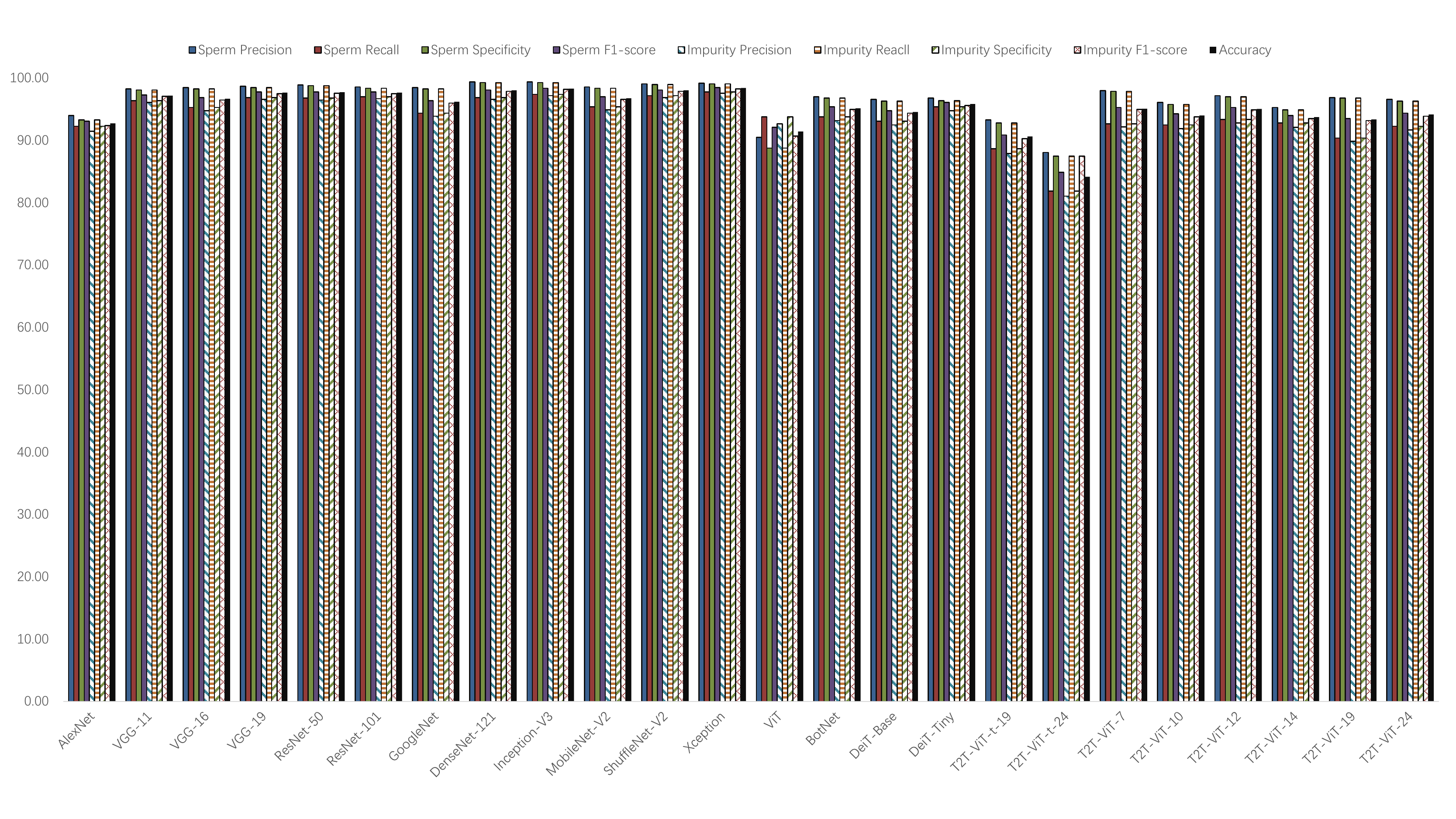}
}
\caption{The histograms of each evaluation metric of test set of CNN and VT models under FGM. }
\label{FIG:}
\end{figure*} 
\addtocounter{figure}{-1}       
\begin{figure*} 
\flushleft
\addtocounter{subfigure}{2}      
\subfigure[FGM: $\epsilon$ = $0.004$.]{
\includegraphics[scale=0.5]{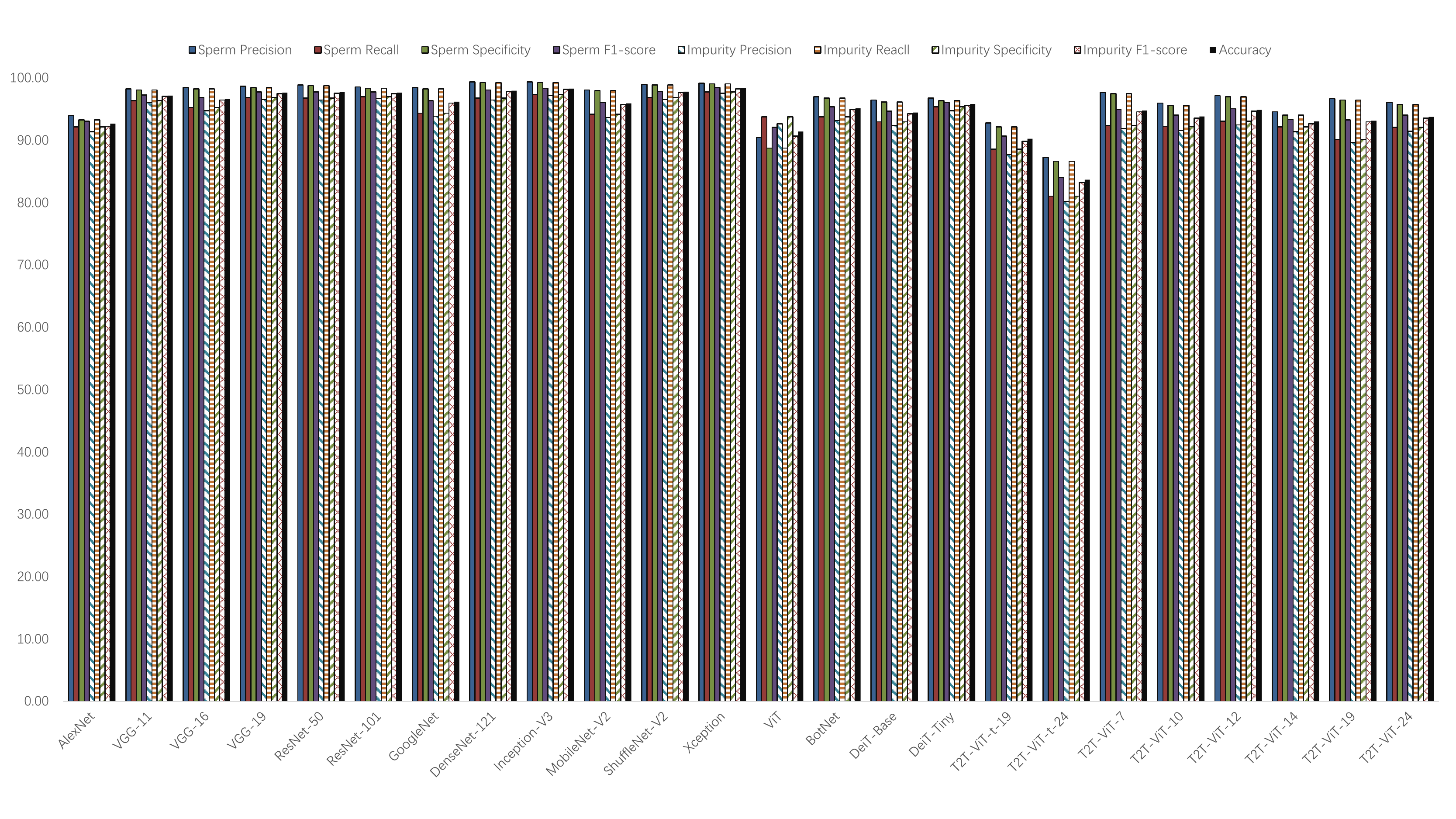}
}
\quad
\subfigure[FGM: $\epsilon$ = $0.008$.]{
\includegraphics[scale=0.5]{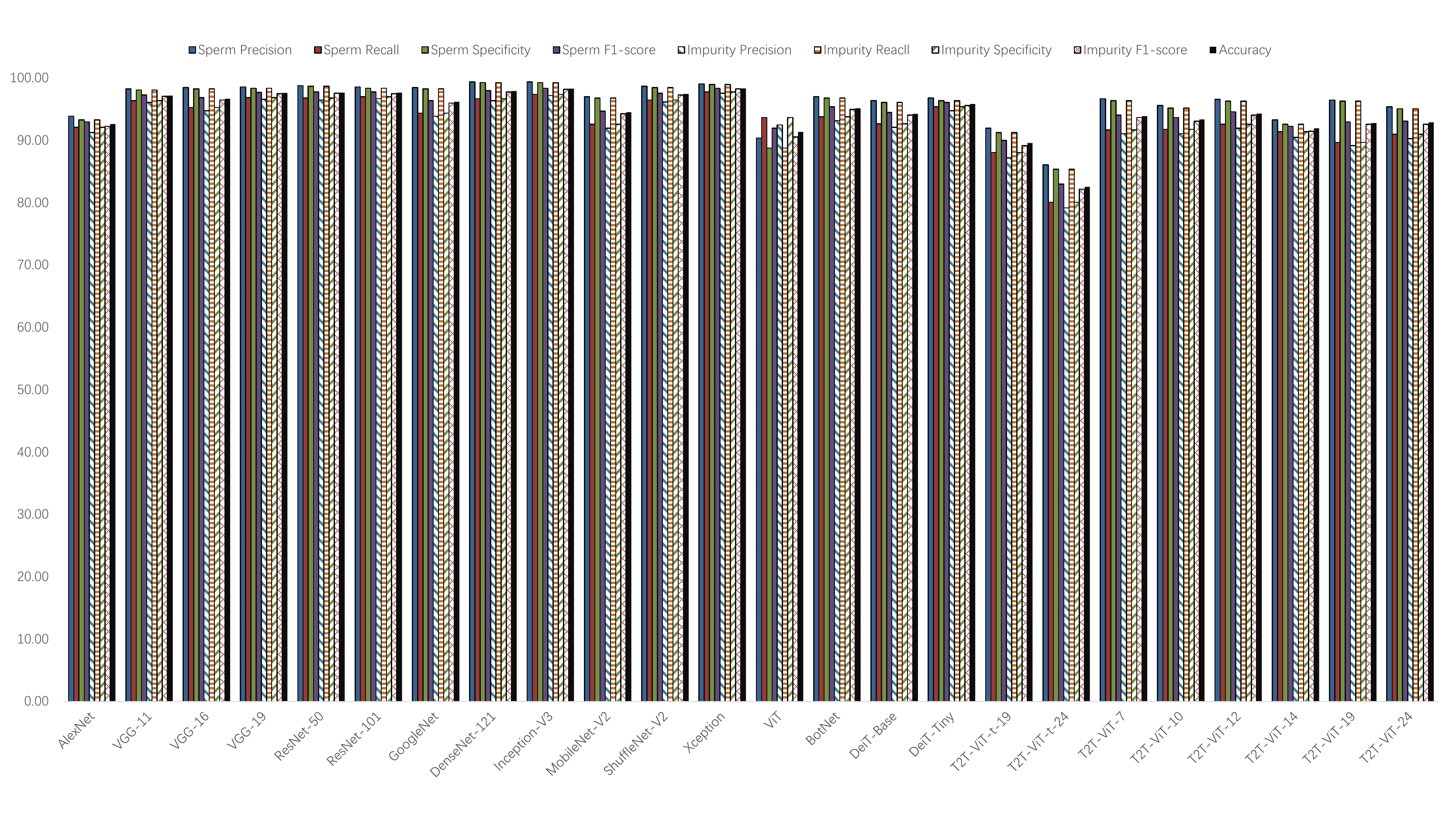}
}
\caption{The histograms of each evaluation metric of test set of CNN and VT models under FGM. }
\label{FIG:}
\end{figure*}
\addtocounter{figure}{-1}       
\begin{figure*} 
\flushleft
\addtocounter{subfigure}{4}      
\subfigure[FGM: $\epsilon$ = $0.016$.]{
\includegraphics[scale=0.5]{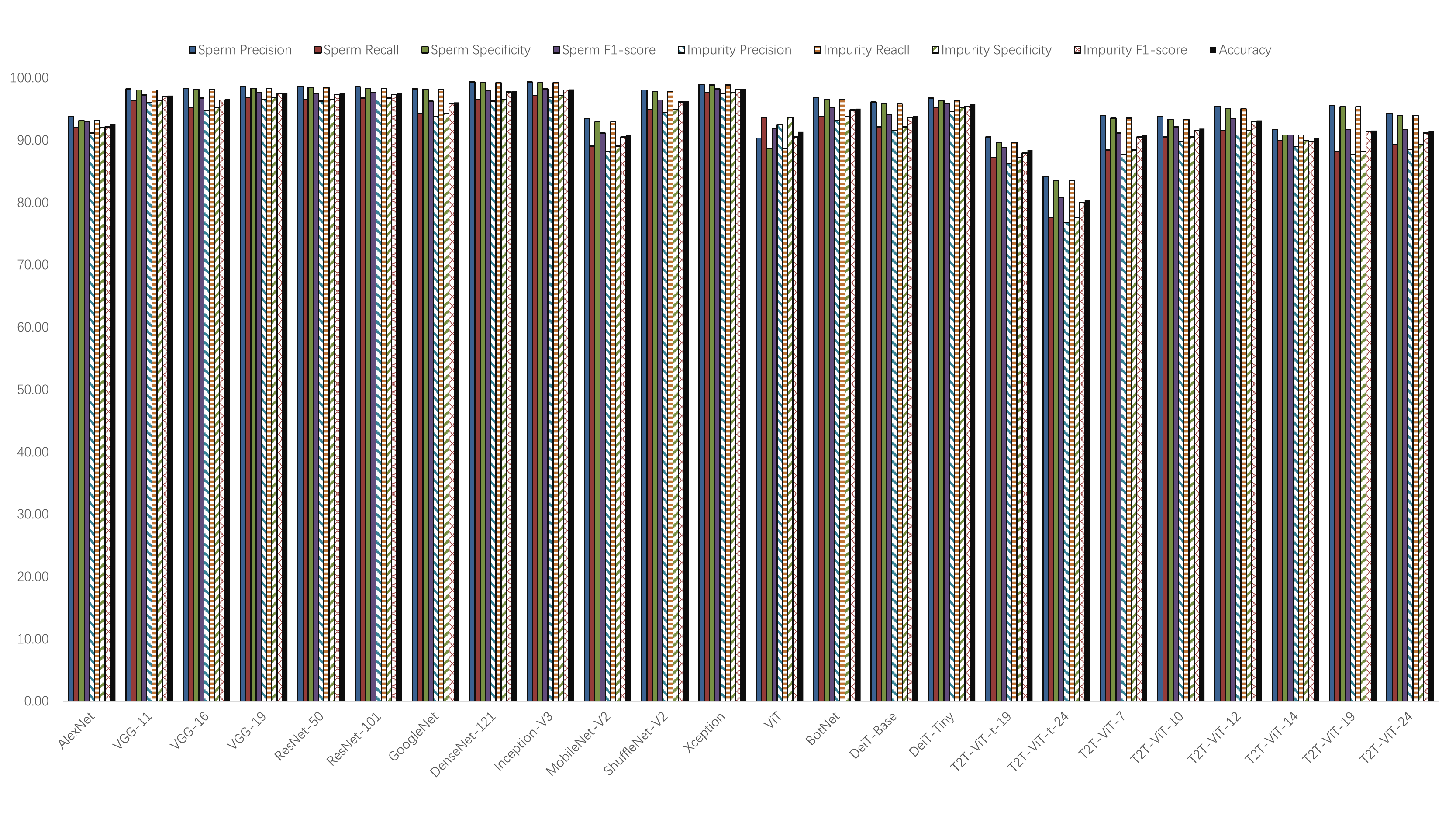}
}
\quad
\subfigure[FGM: $\epsilon$ = $0.032$.]{
\includegraphics[scale=0.5]{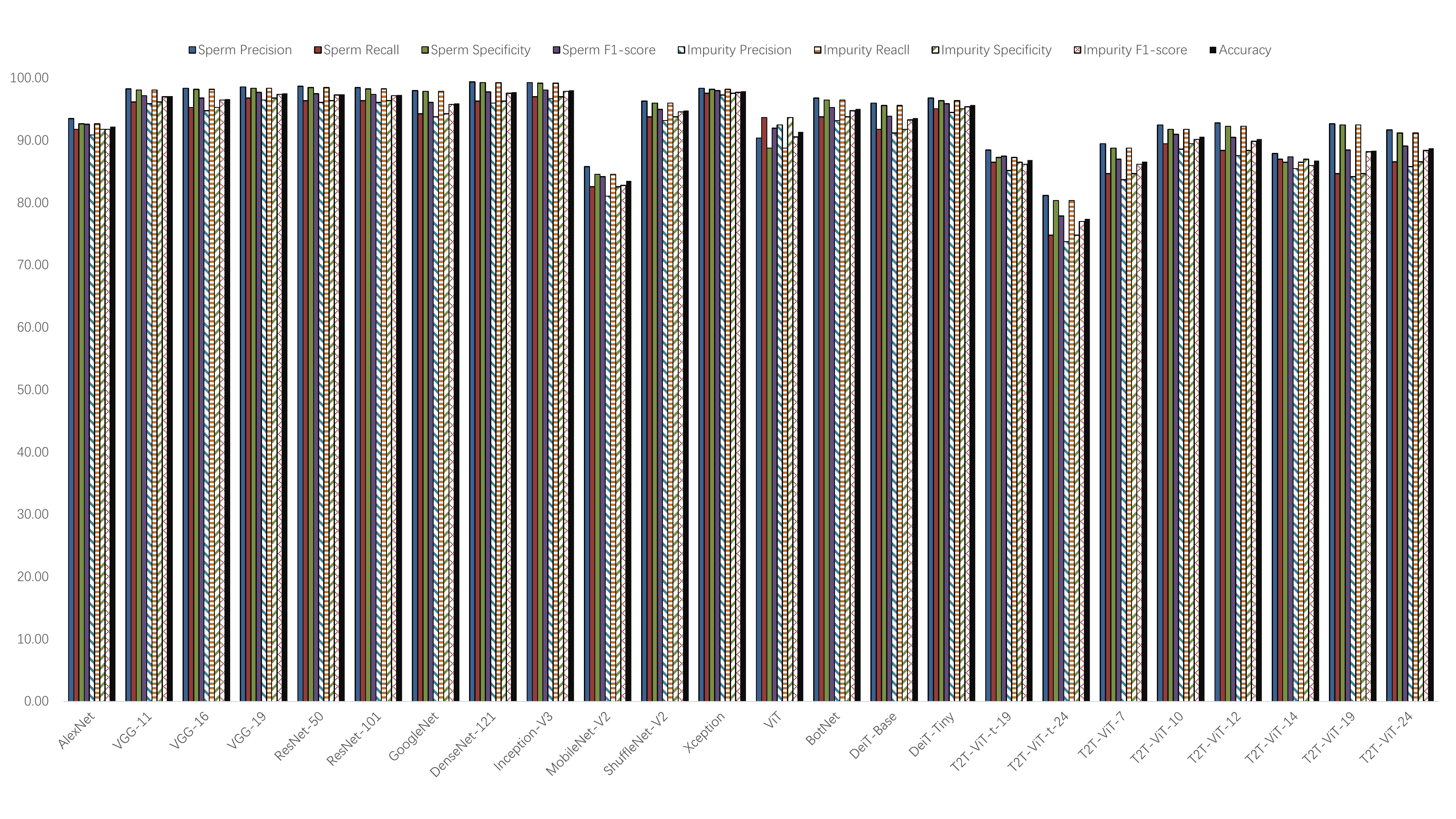}
}
\caption{The histograms of each evaluation metric of test set of CNN and VT models under FGM. }
\label{FIG:}
\end{figure*}
\addtocounter{figure}{-1}       
\begin{figure*} 
\flushleft
\addtocounter{subfigure}{6}      
\subfigure[FGM: $\epsilon$ = $0.64$.]{
\includegraphics[scale=0.5]{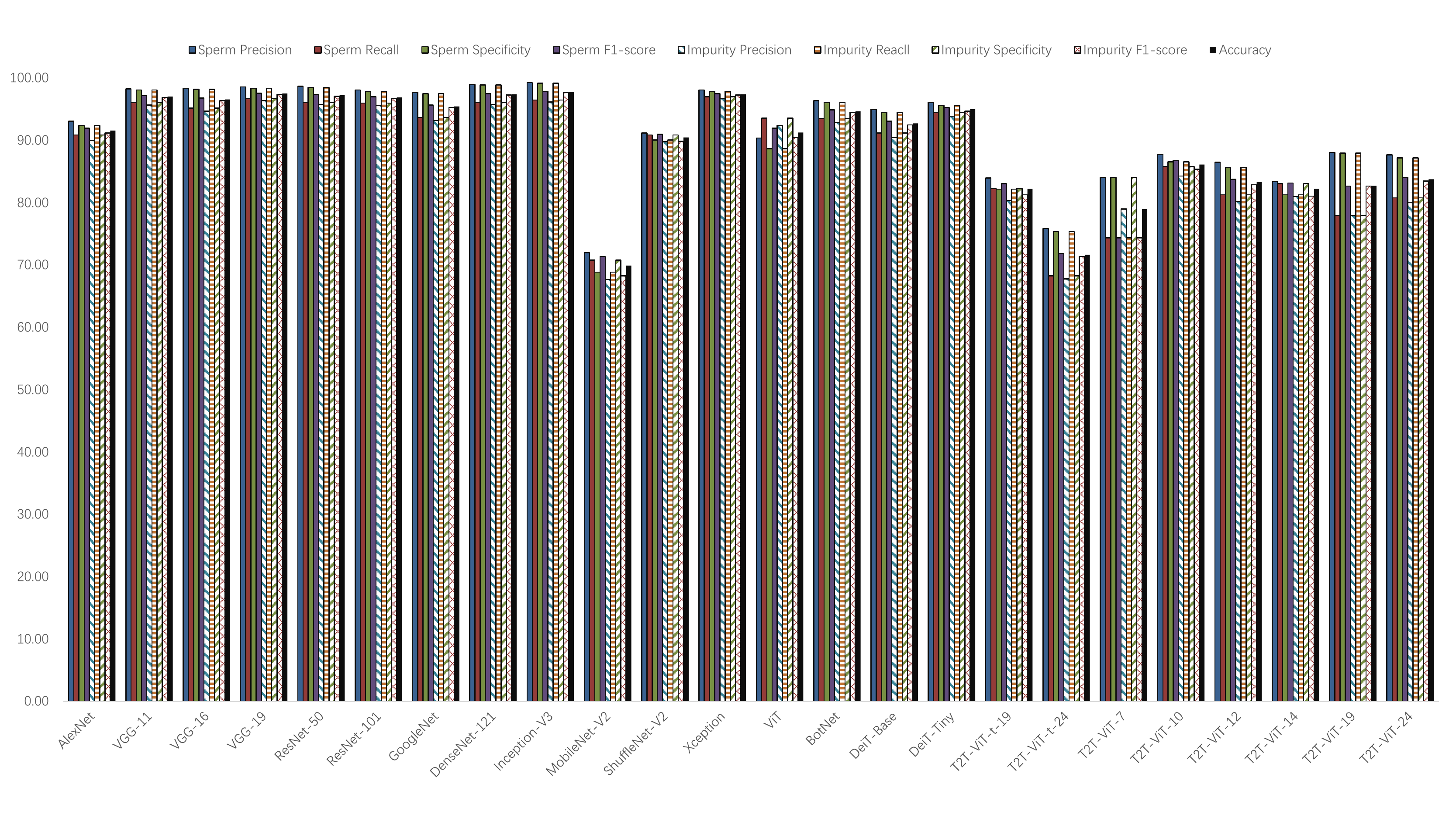}
}
\quad
\subfigure[FGM: $\epsilon$ = $0.128$.]{
\includegraphics[scale=0.5]{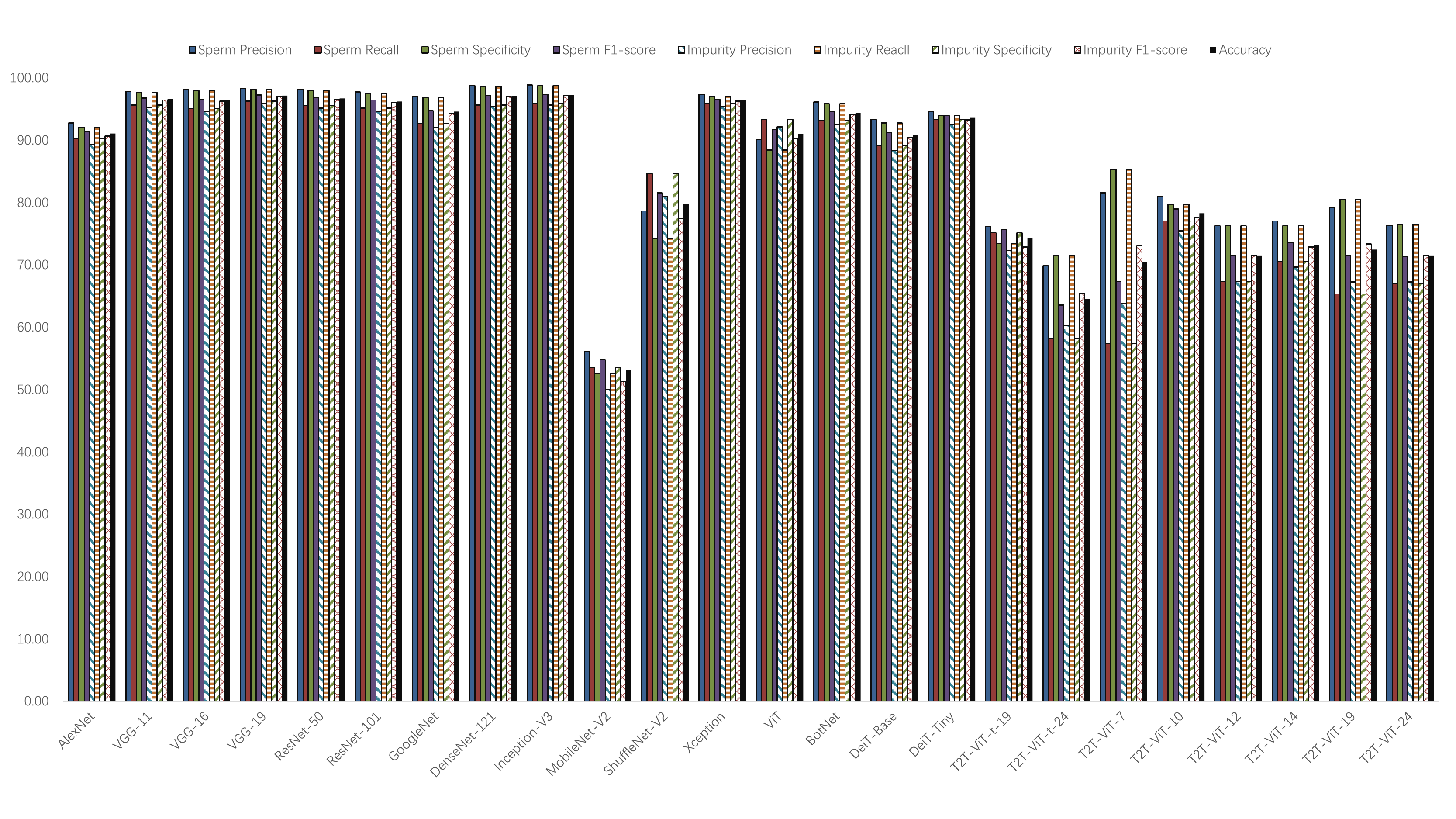}
}
\caption{The histograms of each evaluation metric of test set of CNN and VT models under FGM. }
\label{FIG:}
\end{figure*}
\addtocounter{figure}{-1}       
\begin{figure*} 
\flushleft
\addtocounter{subfigure}{8}      
\subfigure[FGM: $\epsilon$ = $0.256$.]{
\includegraphics[scale=0.5]{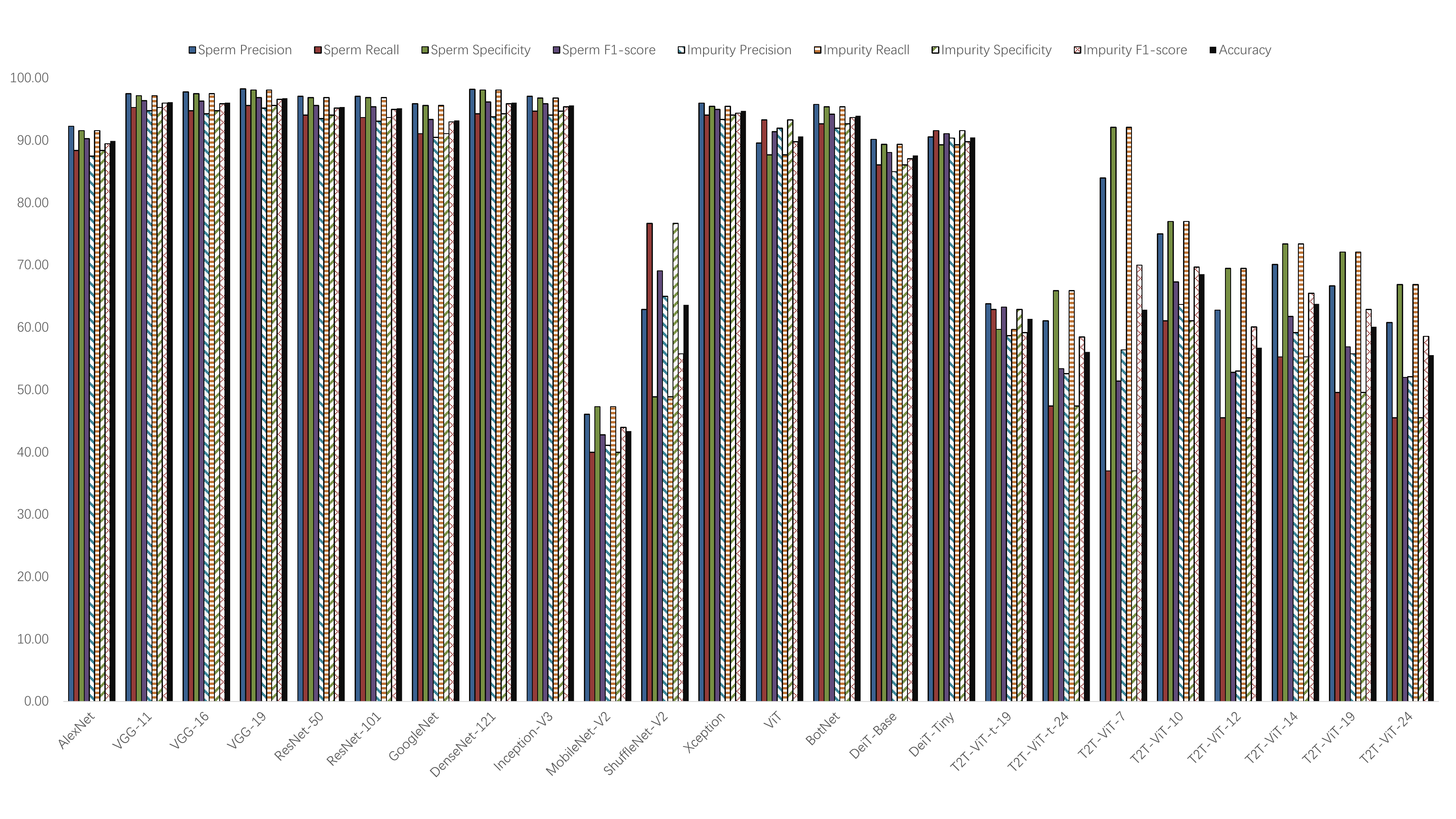}
}
\caption{The histograms of each evaluation metric of test set of CNN and VT models under FGM. }
\label{FIG:17}
\end{figure*}

\begin{figure*}
\flushleft
\subfigure[FGSM: $\epsilon$ = $0.001$.]{
\includegraphics[scale=0.5]{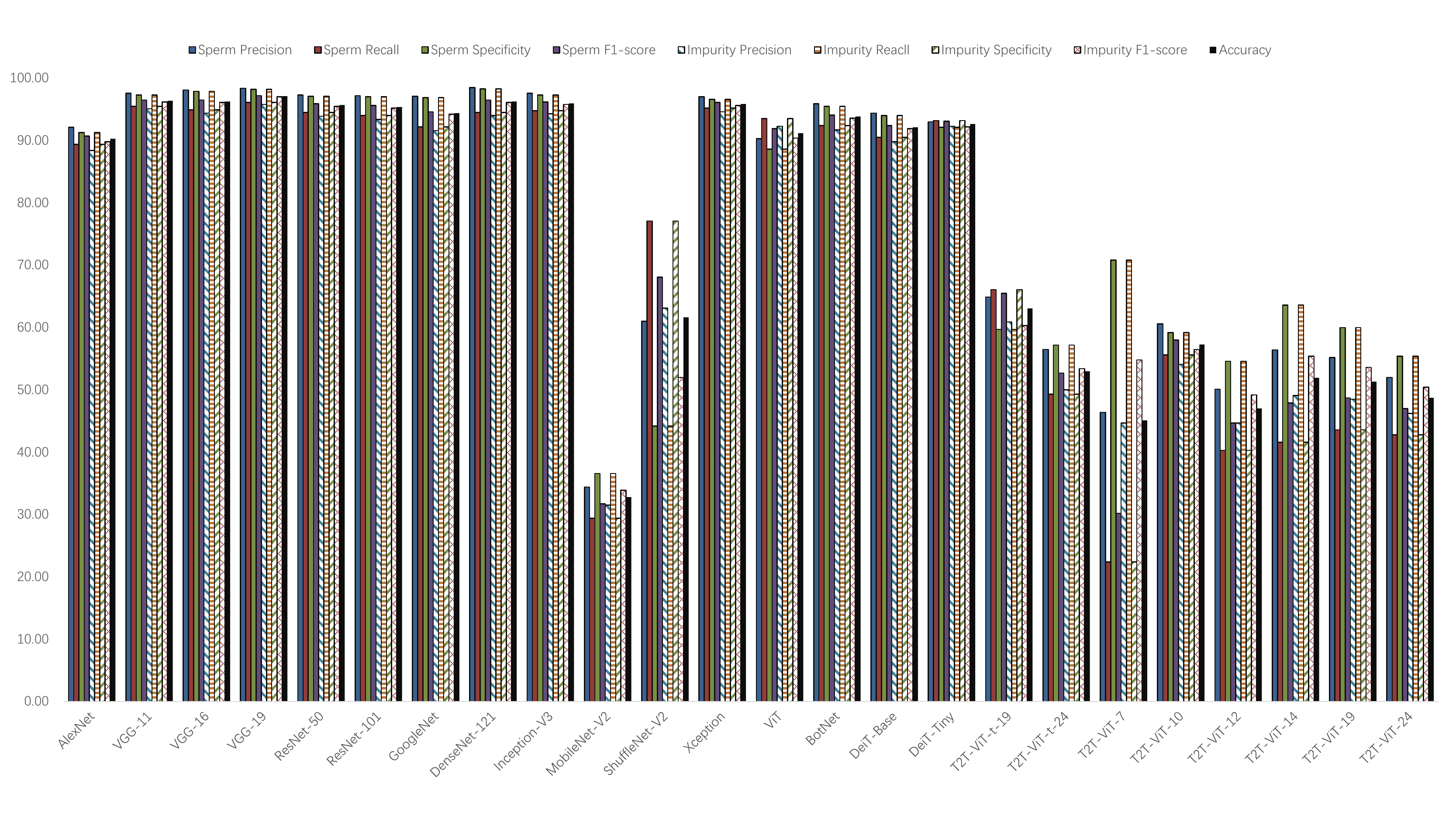}
}
\subfigure[FGSM: $\epsilon$ = $0.002$.]{
\includegraphics[scale=0.5]{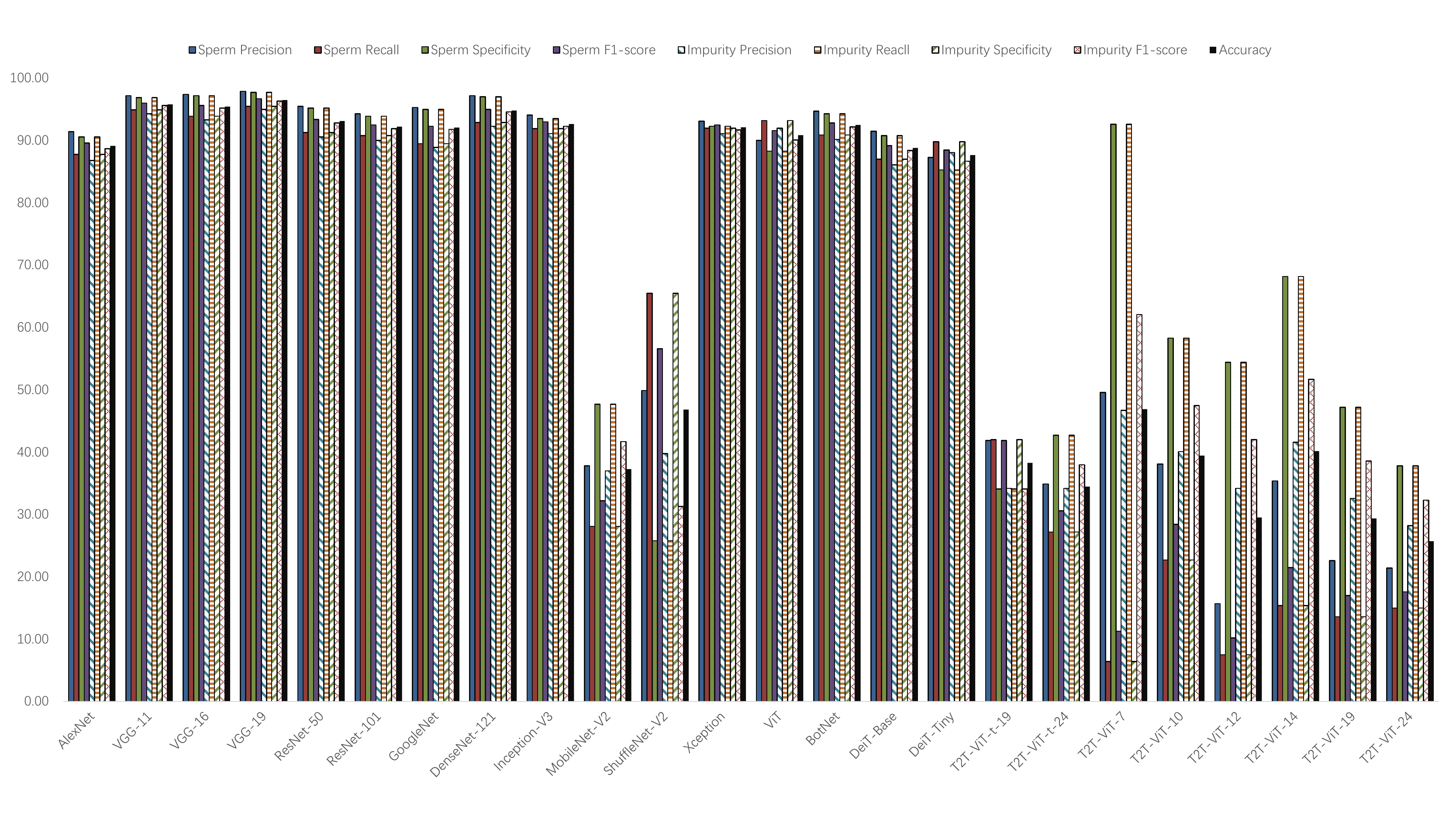}
}
\caption{The histograms of each evaluation metric of test set of CNN and VT models under FGSM. }
\label{FIG:}
\end{figure*} 
\addtocounter{figure}{-1}       
\begin{figure*} 
\flushleft
\addtocounter{subfigure}{2}      
\subfigure[FGSM: $\epsilon$ = $0.004$.]{
\includegraphics[scale=0.5]{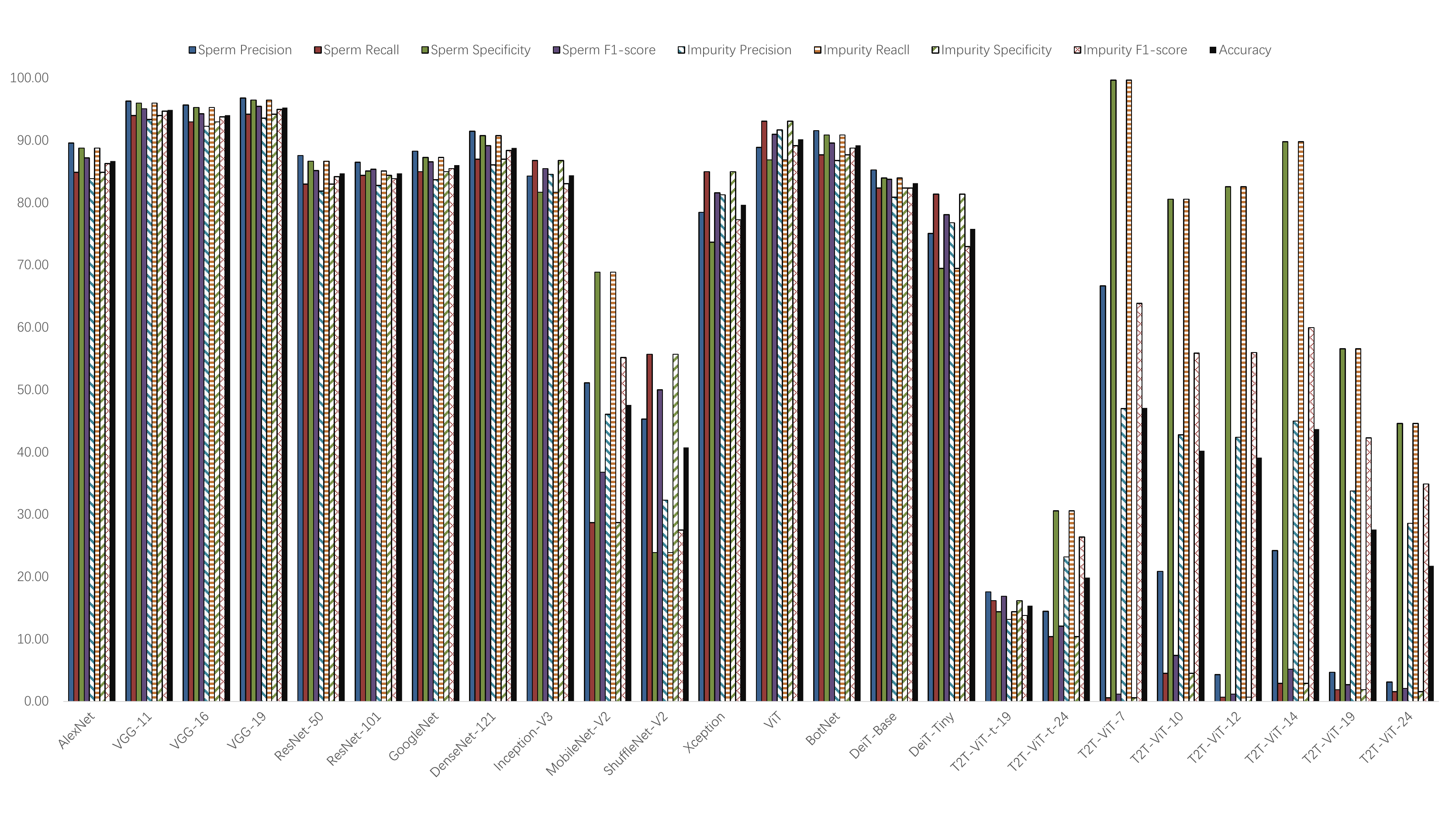}
}
\quad
\subfigure[FGSM: $\epsilon$ = $0.008$.]{
\includegraphics[scale=0.5]{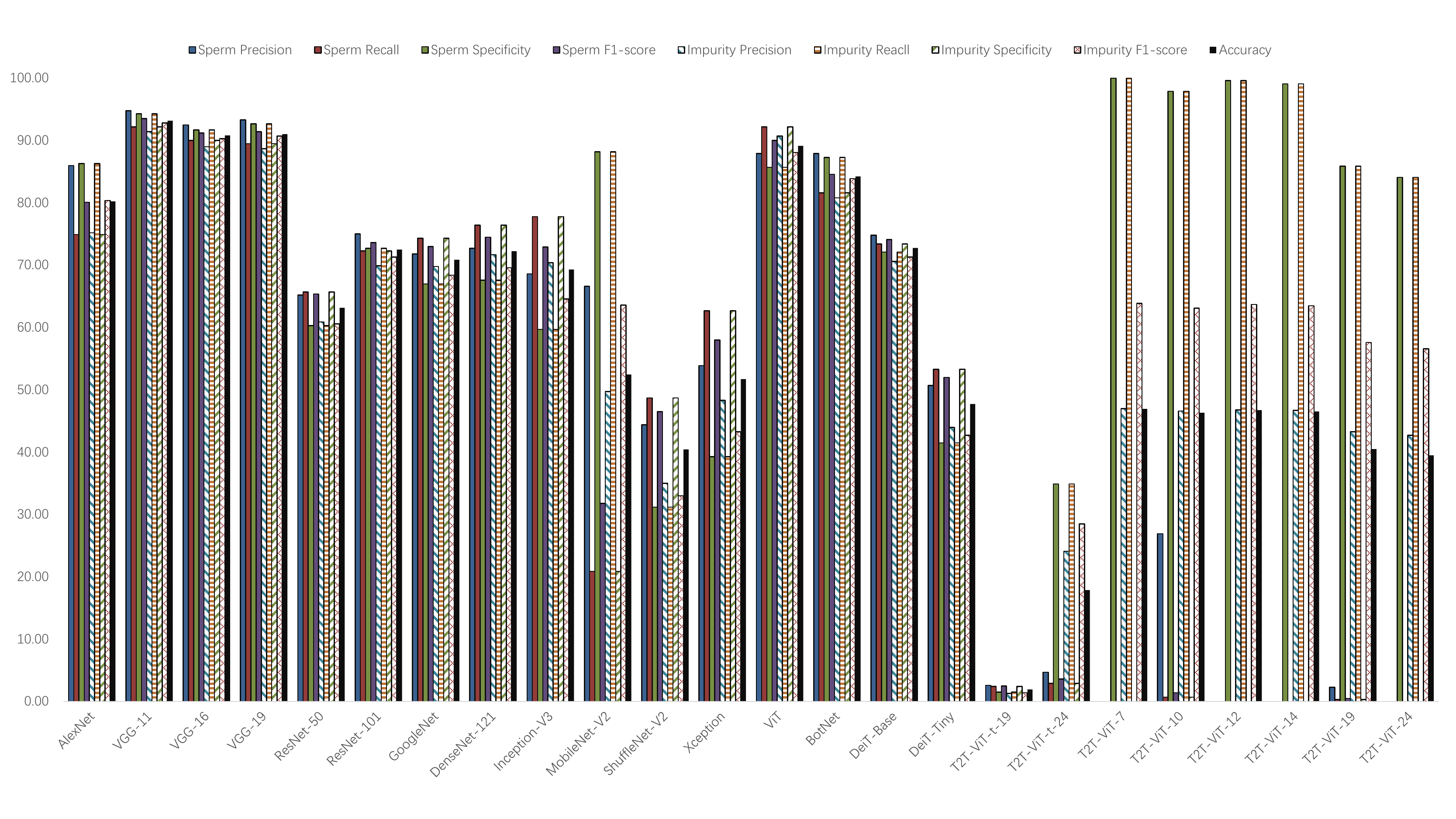}
}
\caption{The histograms of each evaluation metric of test set of CNN and VT models under FGSM. }
\label{FIG:}
\end{figure*}
\addtocounter{figure}{-1}       
\begin{figure*} 
\flushleft
\addtocounter{subfigure}{4}      
\subfigure[FGSM: $\epsilon$ = $0.016$.]{
\includegraphics[scale=0.5]{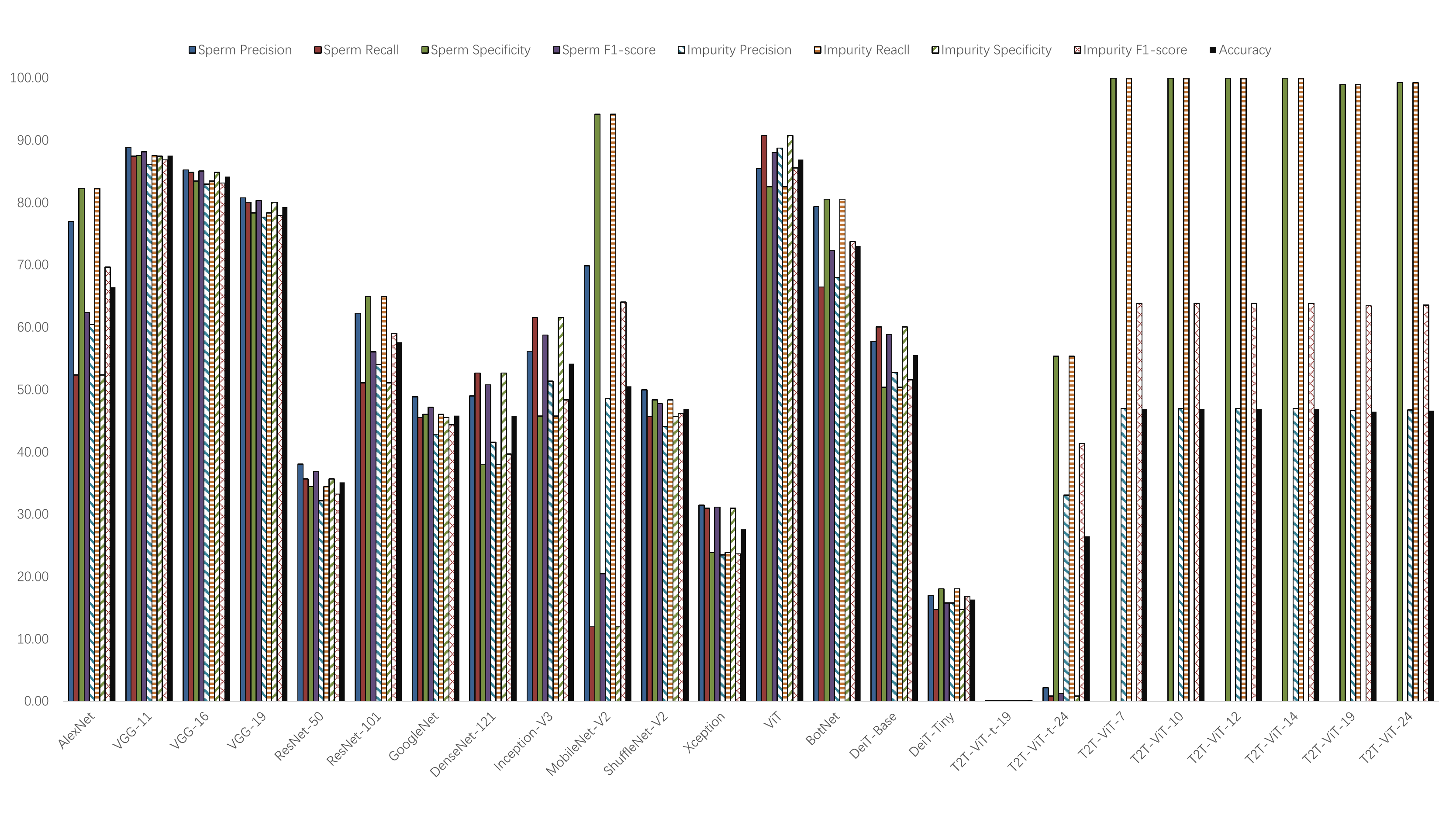}
}
\quad
\subfigure[FGSM: $\epsilon$ = $0.032$.]{
\includegraphics[scale=0.5]{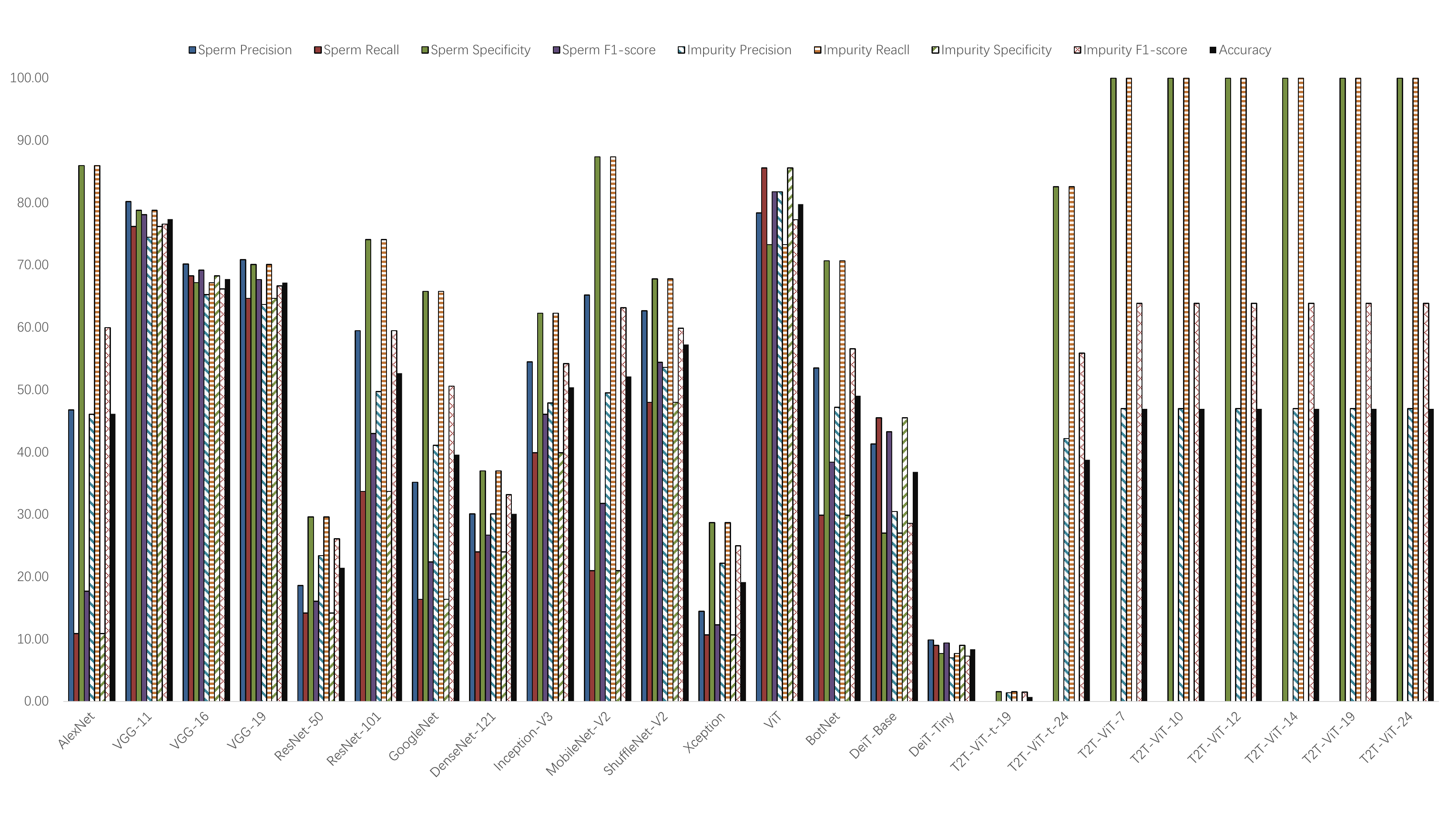}
}
\caption{The histograms of each evaluation metric of test set of CNN and VT models under FGSM. }
\label{FIG:}
\end{figure*}
\addtocounter{figure}{-1}       
\begin{figure*} 
\flushleft
\addtocounter{subfigure}{6}      
\subfigure[FGSM: $\epsilon$ = $0.64$.]{
\includegraphics[scale=0.5]{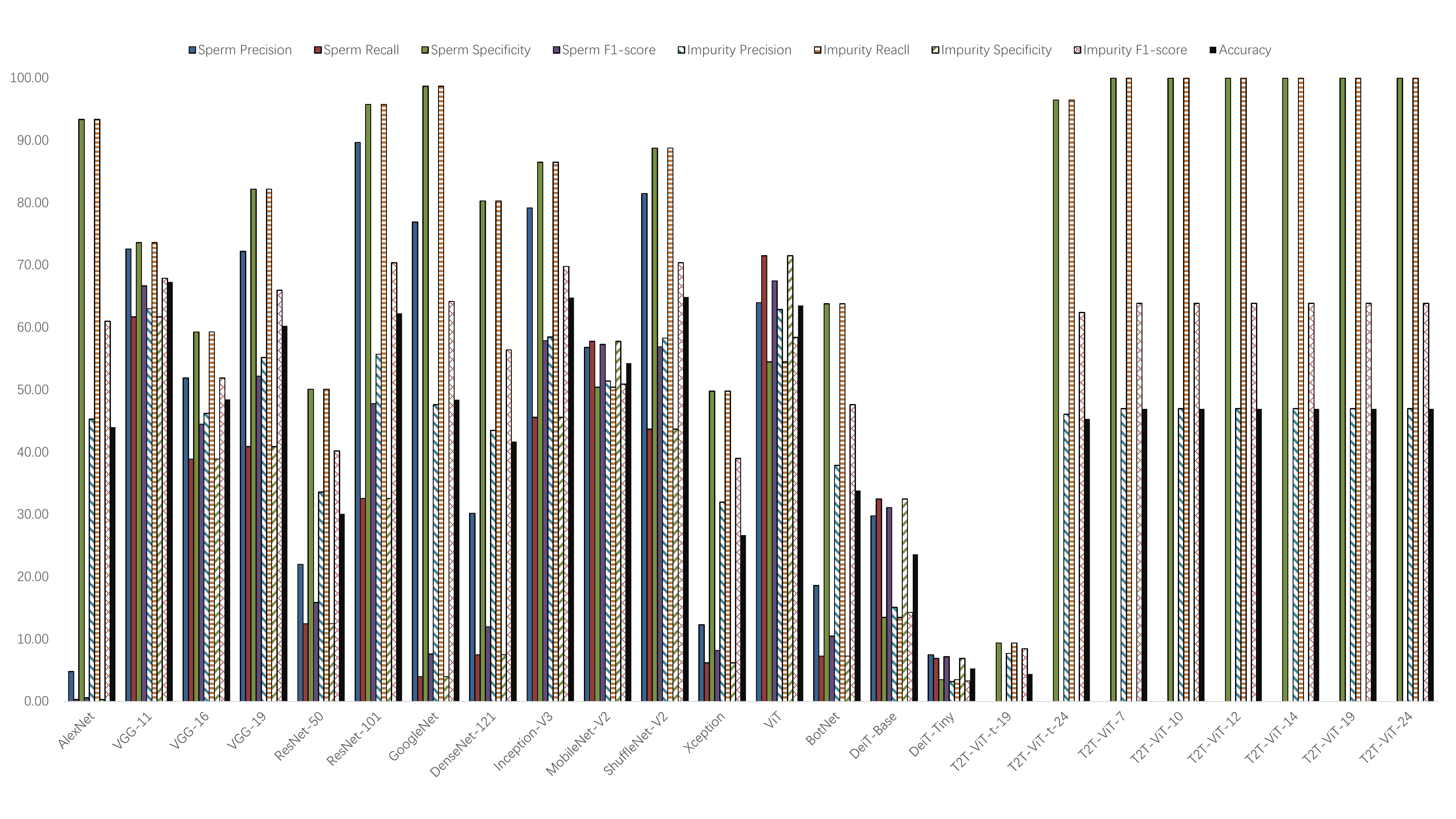}
}
\quad
\subfigure[FGSM: $\epsilon$ = $0.128$.]{
\includegraphics[scale=0.5]{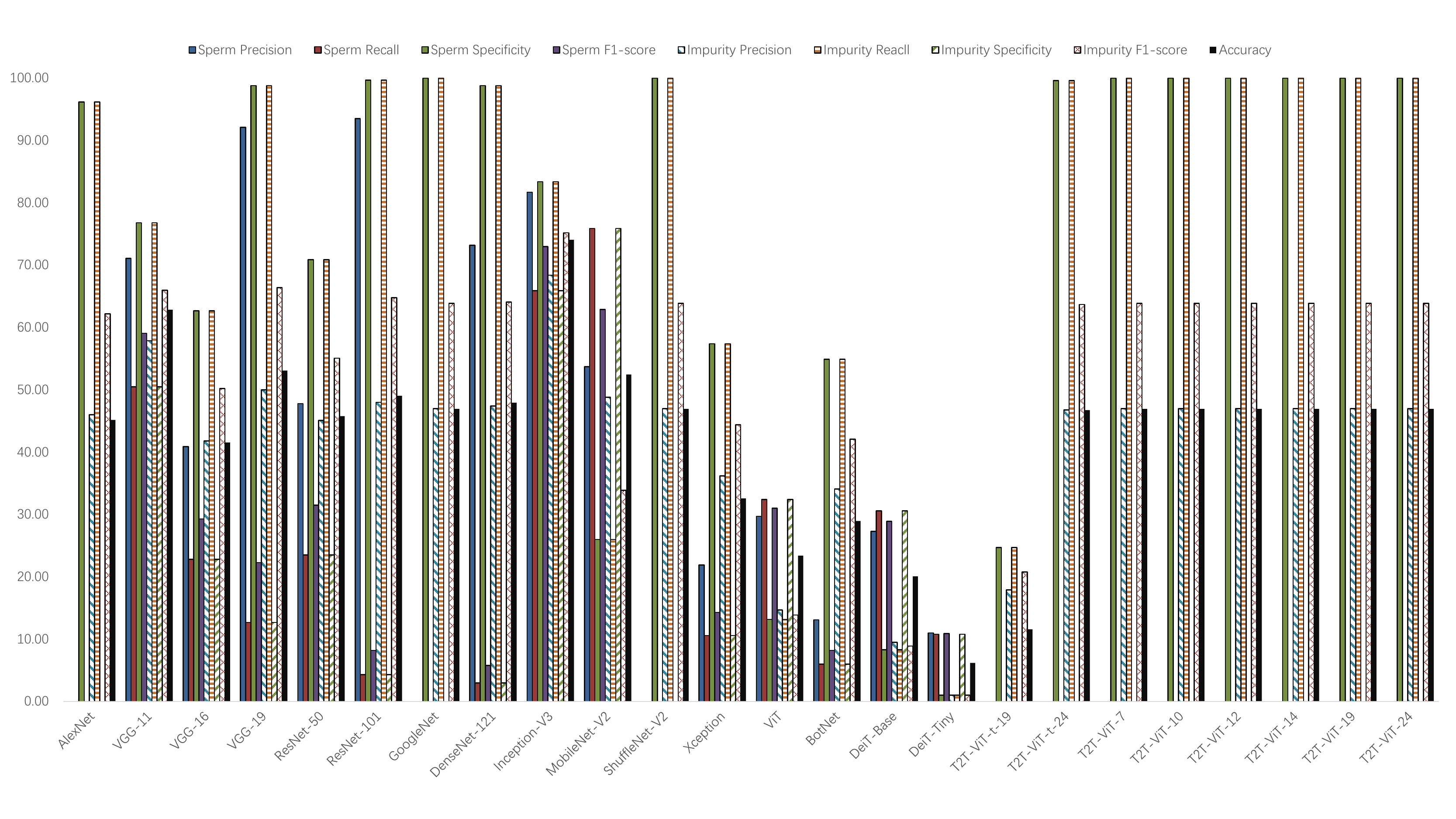}
}
\caption{The histograms of each evaluation metric of test set of CNN and VT models under FGSM. }
\label{FIG:}
\end{figure*}
\addtocounter{figure}{-1}       
\begin{figure*} 
\flushleft
\addtocounter{subfigure}{8}      
\subfigure[FGSM: $\epsilon$ = $0.256$.]{
\includegraphics[scale=0.5]{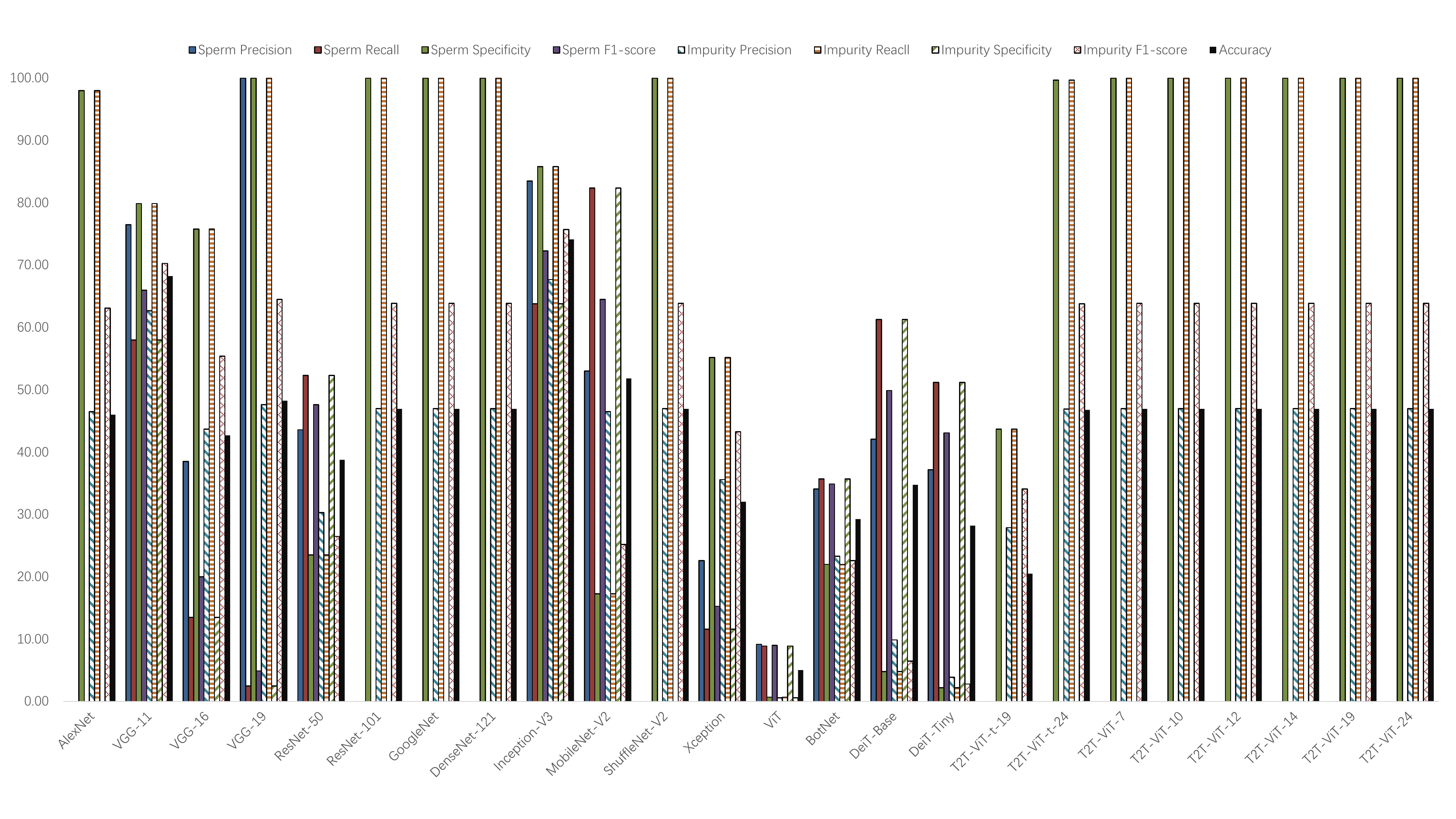}
}
\caption{The histograms of each evaluation metric of test set of CNN and VT models under FGSM. }
\label{FIG:18}
\end{figure*}

\begin{figure*}
\flushleft
\subfigure[I-FGSM: $\epsilon$ = $0.001$.]{
\includegraphics[scale=0.5]{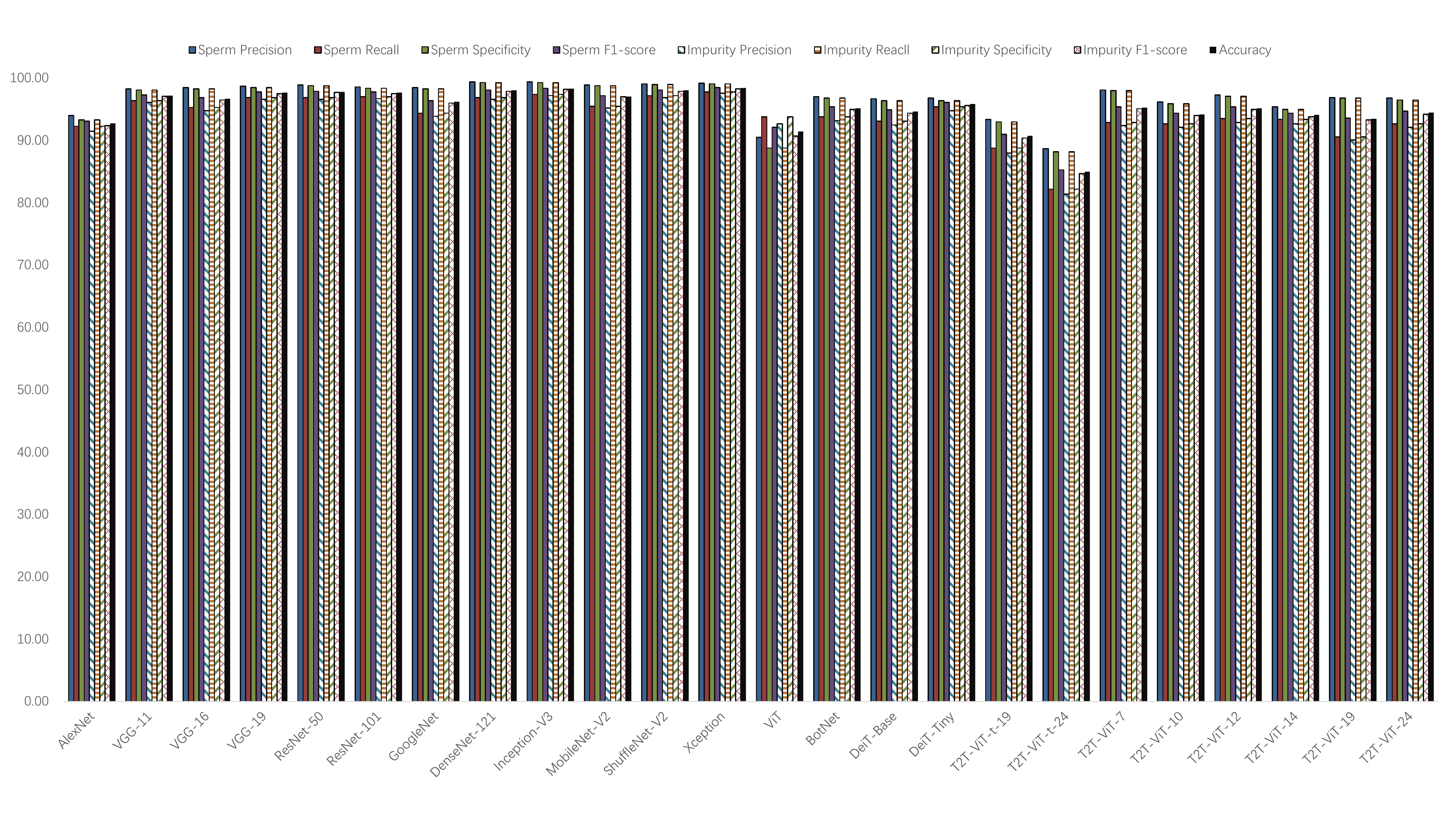}
}
\subfigure[I-FGSM: $\epsilon$ = $0.002$.]{
\includegraphics[scale=0.5]{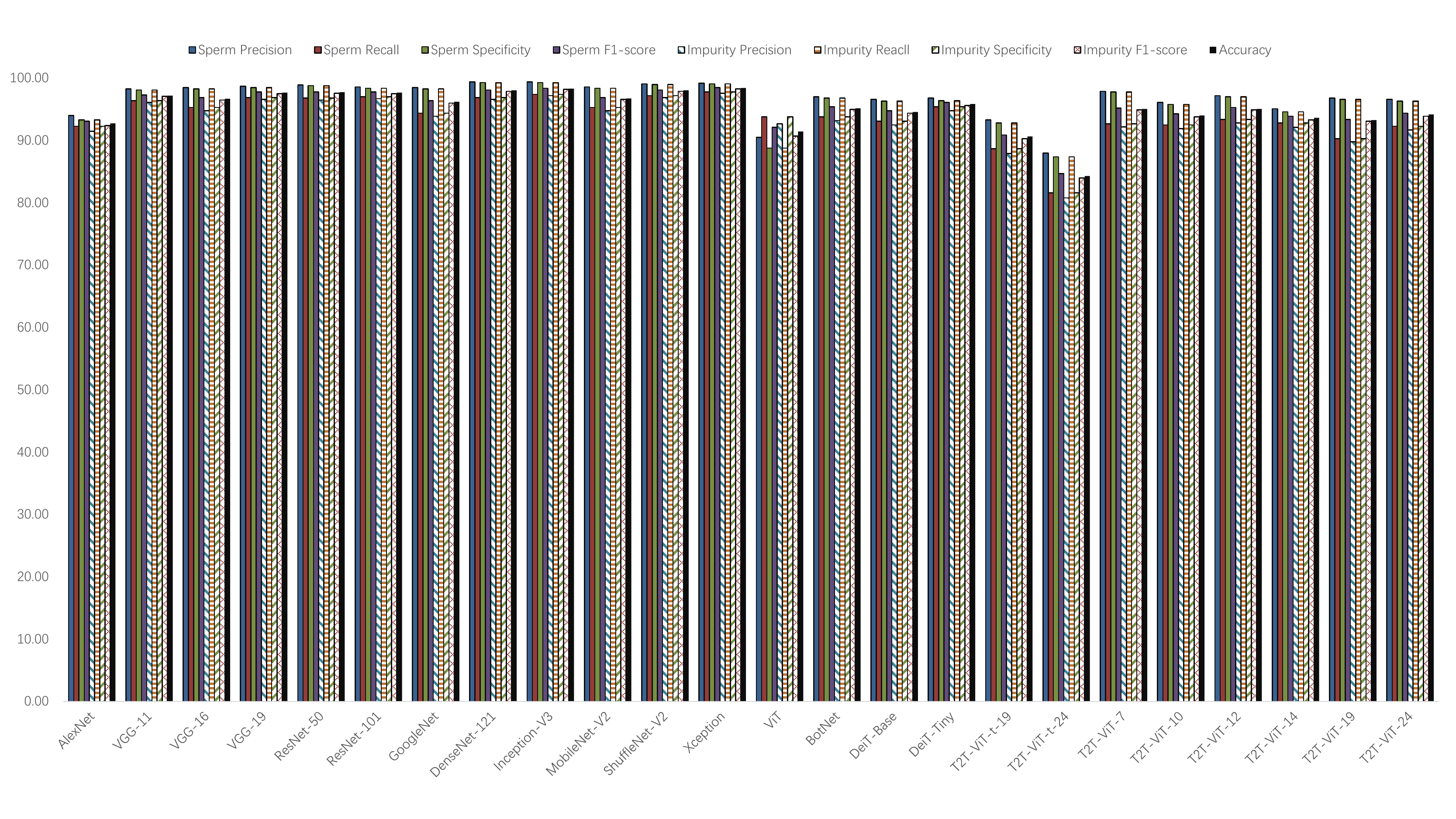}
}
\caption{The histograms of each evaluation metric of test set of CNN and VT models under I-FGSM. }
\label{FIG:}
\end{figure*} 
\addtocounter{figure}{-1}       
\begin{figure*} 
\flushleft
\addtocounter{subfigure}{2}      
\subfigure[I-FGSM: $\epsilon$ = $0.004$.]{
\includegraphics[scale=0.5]{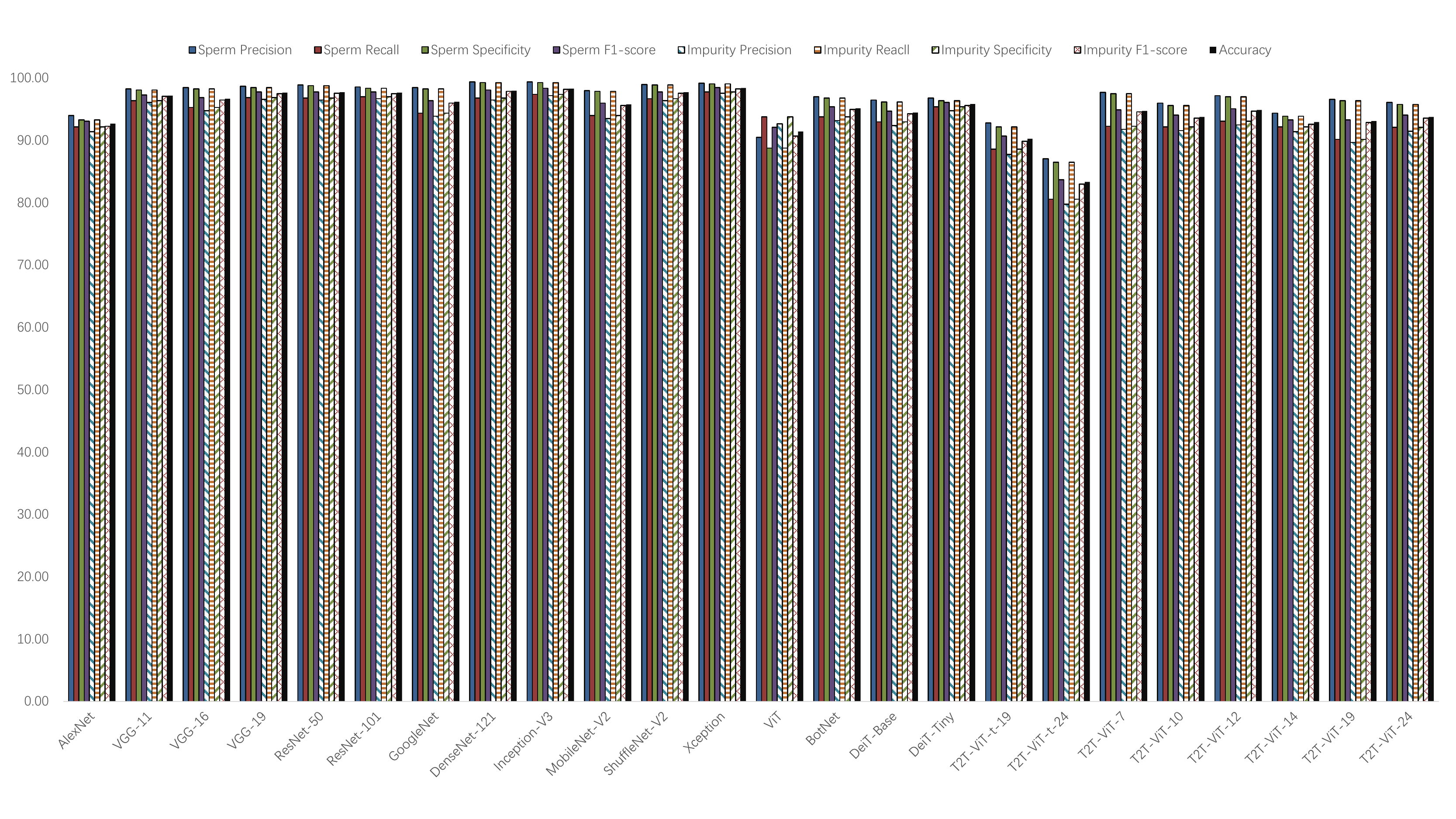}
}
\quad
\subfigure[I-FGSM: $\epsilon$ = $0.008$.]{
\includegraphics[scale=0.5]{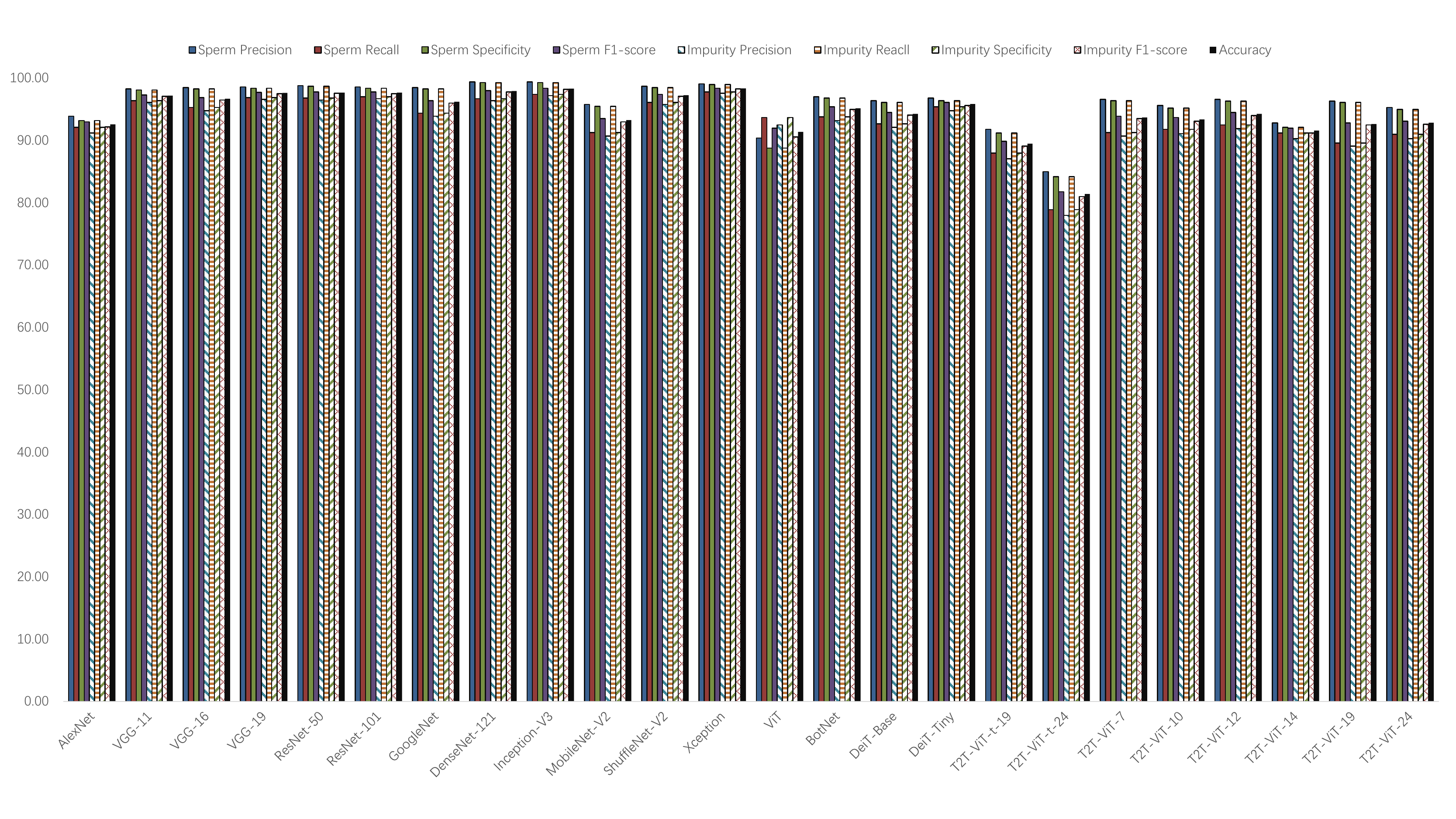}
}
\caption{The histograms of each evaluation metric of test set of CNN and VT models under I-FGSM. }
\label{FIG:}
\end{figure*}
\addtocounter{figure}{-1}       
\begin{figure*} 
\flushleft
\addtocounter{subfigure}{4}      
\subfigure[I-FGSM: $\epsilon$ = $0.016$.]{
\includegraphics[scale=0.5]{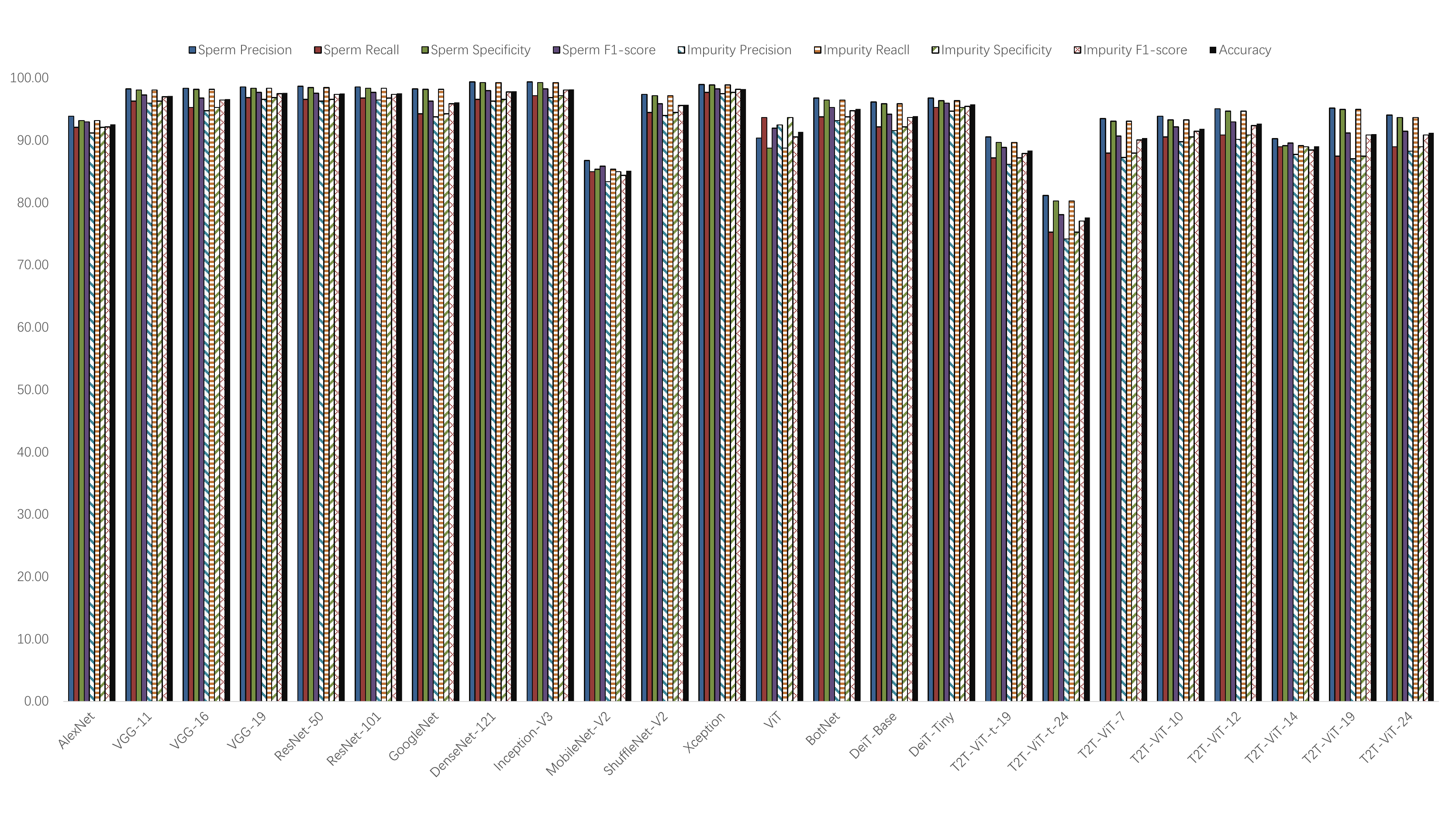}
}
\quad
\subfigure[I-FGSM: $\epsilon$ = $0.032$.]{
\includegraphics[scale=0.5]{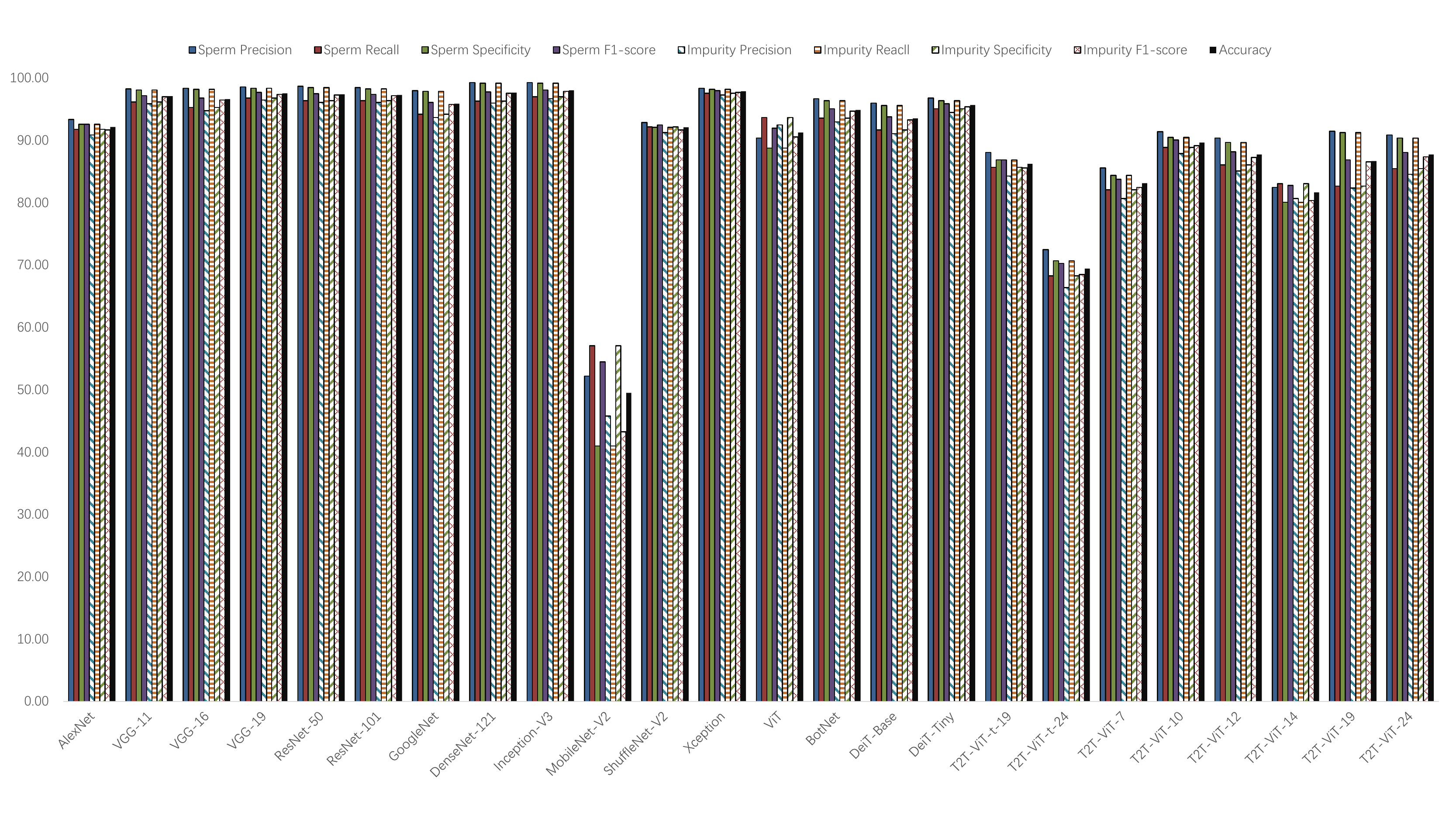}
}
\caption{The histograms of each evaluation metric of test set of CNN and VT models under I-FGSM. }
\label{FIG:}
\end{figure*}
\addtocounter{figure}{-1}       
\begin{figure*} 
\flushleft
\addtocounter{subfigure}{6}      
\subfigure[I-FGSM: $\epsilon$ = $0.64$.]{
\includegraphics[scale=0.5]{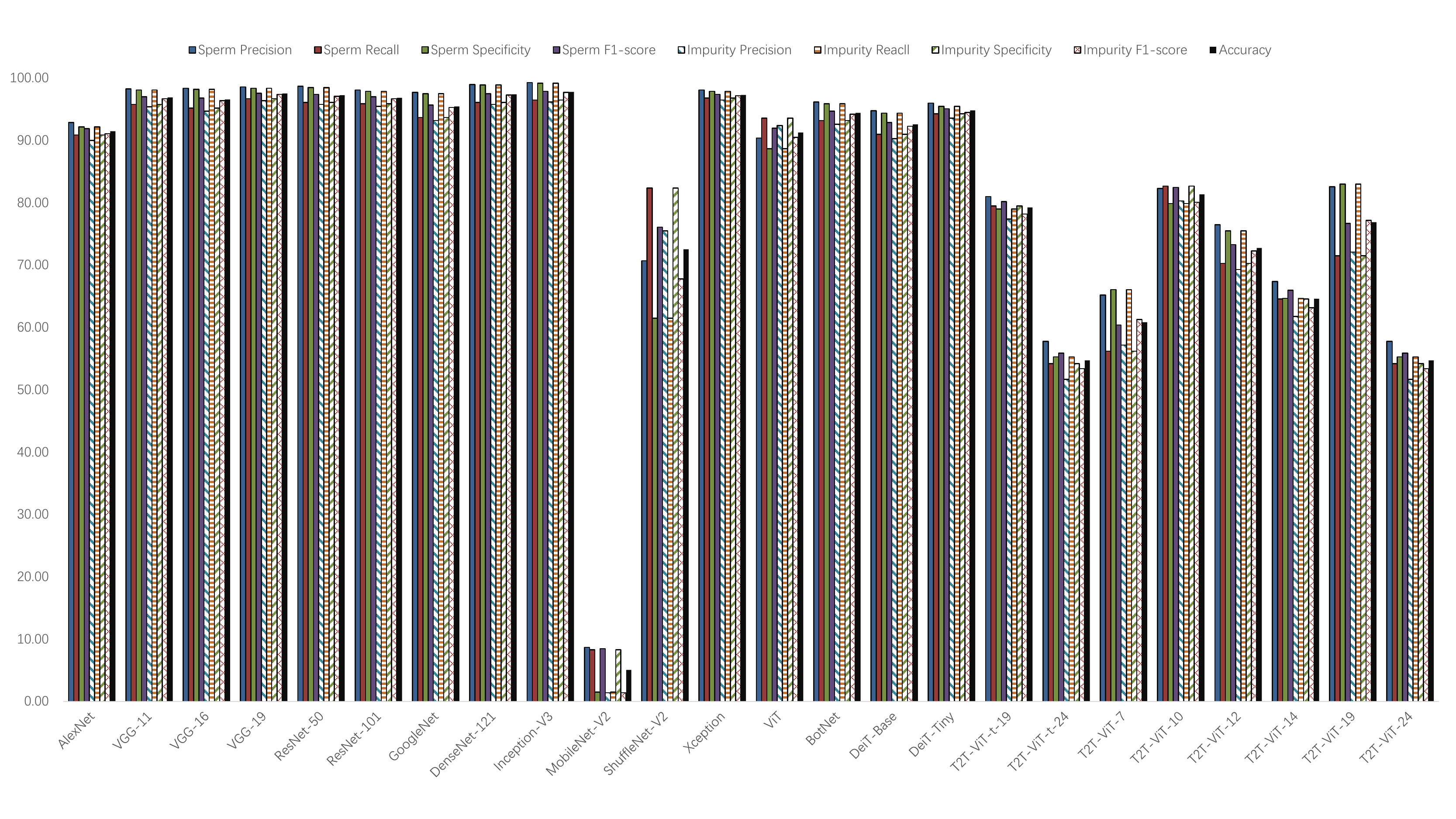}
}
\quad
\subfigure[I-FGSM: $\epsilon$ = $0.128$.]{
\includegraphics[scale=0.5]{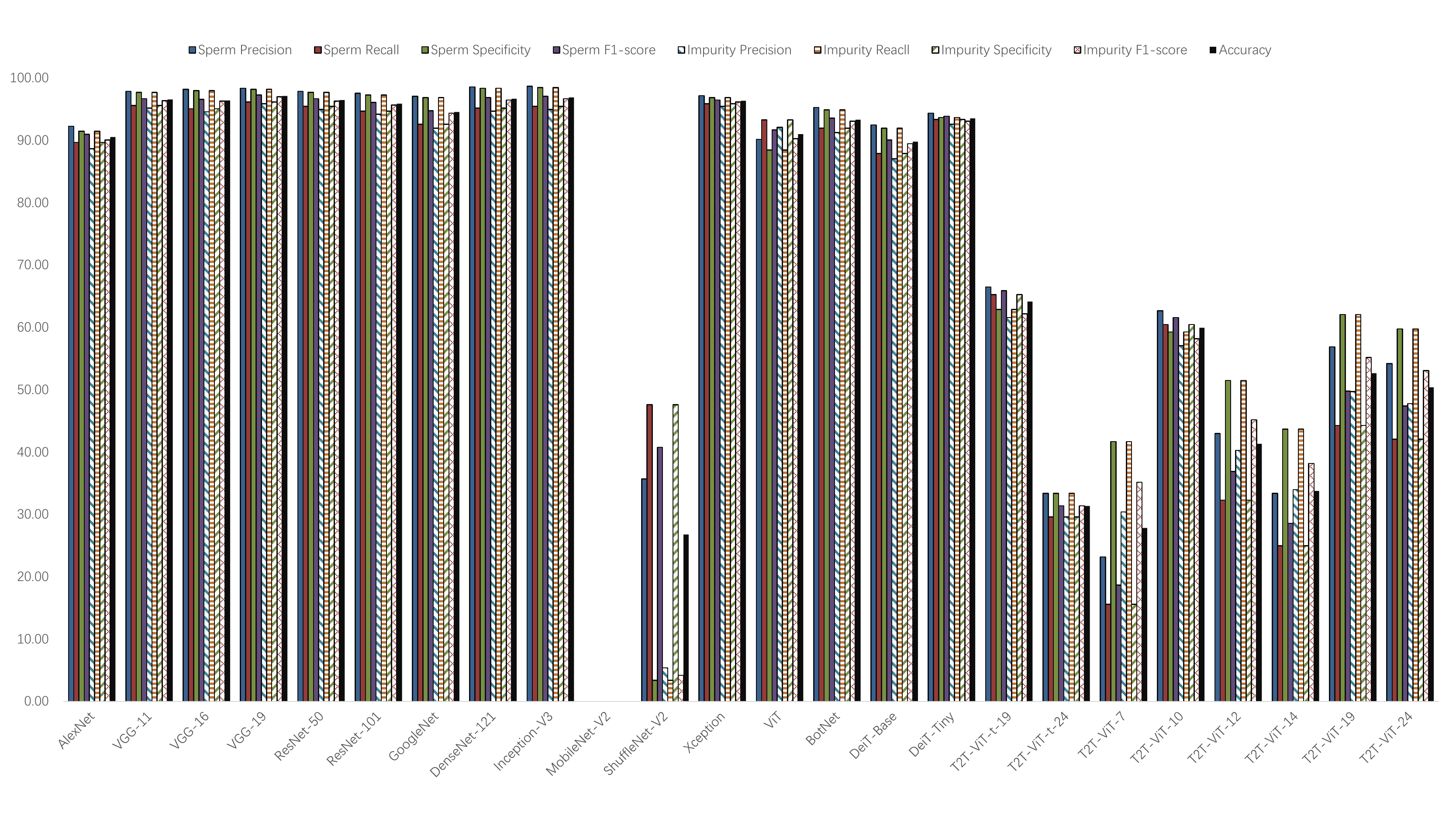}
}
\caption{The histograms of each evaluation metric of test set of CNN and VT models under I-FGSM. }
\label{FIG:}
\end{figure*}
\addtocounter{figure}{-1}       
\begin{figure*} 
\flushleft
\addtocounter{subfigure}{8}      
\subfigure[I-FGSM: $\epsilon$ = $0.256$.]{
\includegraphics[scale=0.5]{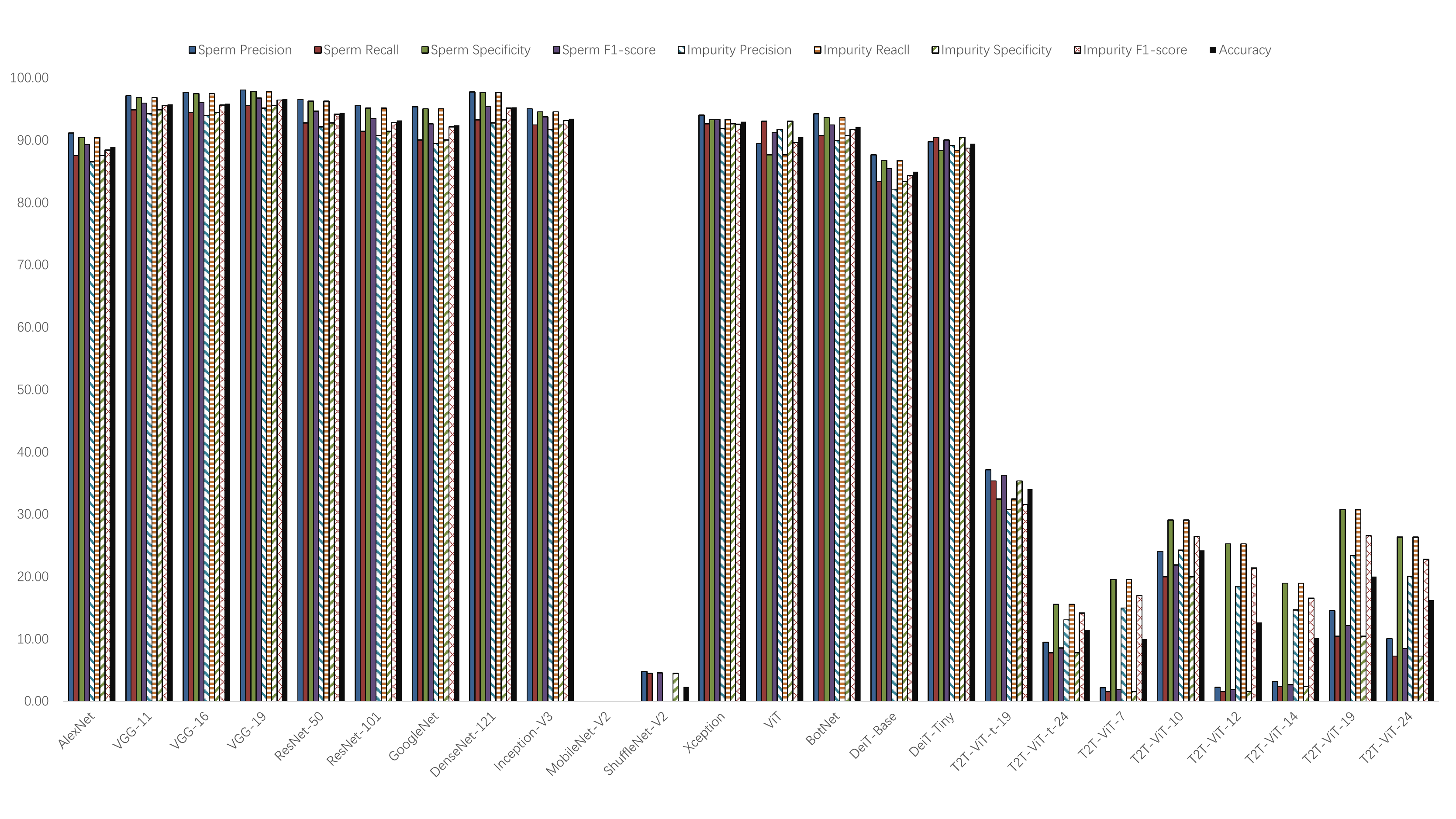}
}
\caption{The histograms of each evaluation metric of test set of CNN and VT models under I-FGSM. }
\label{FIG:19}
\end{figure*}
\end{appendix}

\end{document}